\documentclass{article}

\PassOptionsToPackage{numbers, compress}{natbib}
\usepackage[preprint]{neurips_2024}

\usepackage[utf8]{inputenc} % allow utf-8 input
\usepackage[T1]{fontenc}    % use 8-bit T1 fonts
\usepackage{hyperref}       % hyperlinks
\usepackage{url}            % simple URL typesetting
\usepackage{booktabs}       % professional-quality tables
\usepackage{amsfonts}       % blackboard math symbols
\usepackage{nicefrac}       % compact symbols for 1/2, etc.
\usepackage{microtype}      % microtypography
\usepackage{xcolor}         % colors

\usepackage{multirow}
\usepackage[inline]{enumitem}
\usepackage{subcaption}
\usepackage{amsmath}
\usepackage{amssymb}
\usepackage{mathtools}
\usepackage{amsthm}

\theoremstyle{plain}
\newtheorem{theorem}{Theorem}[section]

\theoremstyle{definition}

\theoremstyle{remark}

\usepackage[export]{adjustbox}

\usepackage[capitalize,noabbrev]{cleveref}

\usepackage{mdframed}   % for framing

\newmdtheoremenv[%
linecolor=gray,leftmargin=10,%
rightmargin=10,
backgroundcolor=gray!40,
nobreak=true,
]{myprop}{Template}[section]

\usepackage[toc]{appendix}
\usepackage{minitoc}

\newcommand{\expsec}[1]{\vspace{+0.2cm} \noindent \textcolor{violet}{\underline{\textbf{#1}}}} 

\title{REAL Sampling: Boosting Factuality and Diversity of Open-Ended Generation via Asymptotic Entropy}

\author{%
  Haw-Shiuan Chang \;\; Nanyun Peng \;\; Mohit Bansal \;\; Anil Ramakrishna \;\; Tagyoung Chung \\
  Amazon AGI Foundations, USA\\
  \texttt{ \{chawshiu,pengnany,mobansal,aniramak,tagyoung\}@amazon.com} \\
}

\begin{document}

\maketitle

%when LLMs are more likely to hallucinate, REAL sampling uses a lower $p$ threshold to increase the factuality. 

\begin{abstract}
Decoding methods for large language models (LLMs) usually struggle with the tradeoff between ensuring factuality and maintaining diversity. For example, a higher $p$ threshold in the nucleus (top-$p$) sampling increases the diversity but decreases the factuality, and vice versa~\citep{lee2022factuality}. In this paper, we propose REAL (\textbf{R}esidual \textbf{E}ntropy from \textbf{A}symptotic \textbf{L}ine) sampling, a decoding method that achieves improved factuality and diversity over nucleus sampling by predicting an adaptive threshold of $p$. Specifically, REAL sampling predicts the step-wise likelihood of an LLM  to hallucinate, and lowers the $p$ threshold when an LLM is likely to hallucinate. Otherwise, REAL sampling increases the $p$ threshold to boost the diversity. To predict the step-wise hallucination likelihood without supervision, we construct a \textbf{T}oken-level \textbf{H}allucination \textbf{F}orecasting (THF) model to predict the asymptotic entropy (i.e., inherent uncertainty) of the next token by extrapolating the next-token entropies from a series of LLMs with different sizes. If a LLM's entropy is higher than the asymptotic entropy (i.e., the LLM is more uncertain than it should be), the THF model predicts a high hallucination hazard, which leads to a lower $p$ threshold in REAL sampling. In the \textsc{FactualityPrompts} benchmark~\citep{lee2022factuality}, we demonstrate that REAL sampling based on a 70M THF model can substantially improve the factuality and diversity of 7B LLMs simultaneously, judged by both retrieval-based metrics and human evaluation. 
After combined with contrastive decoding, REAL sampling outperforms 9 sampling methods, and generates texts that are more factual than the greedy sampling and more diverse than the nucleus sampling with $p=0.5$. Furthermore, the predicted asymptotic entropy is also a useful unsupervised signal for hallucination detection tasks.

%the text from REAL sampling is 
%approximate models' inherent uncertainty by extrapolating the next-token entropies from a series of LLMs with different sizes and predicting the asymptotic entropy accordingly. 

%We further combine REAL sampling with contrastive decoding~\citep{li2022contrastive}, and generate texts that are more factual than the greedy sampling and more diverse than the nucleus sampling with $p=0.5$. We also show that REAL sampling could also be used to improve hallucination detection and creative writing tasks. \violet{the baseline seems to the a little weak. have you compared with typical sampling or/and eta sampling?}
\end{abstract}

\doparttoc % Tell to minitoc to generate a toc for the parts
\faketableofcontents  % Run a fake tableofcontents command for the partocs

%\part{} % Start the document part
%\parttoc % Insert the document TOC

\section{Introduction}
\label{intro}

\begin{figure}
\centering
\begin{minipage}{.55\textwidth}
  \centering
  \includegraphics[width=1\linewidth]{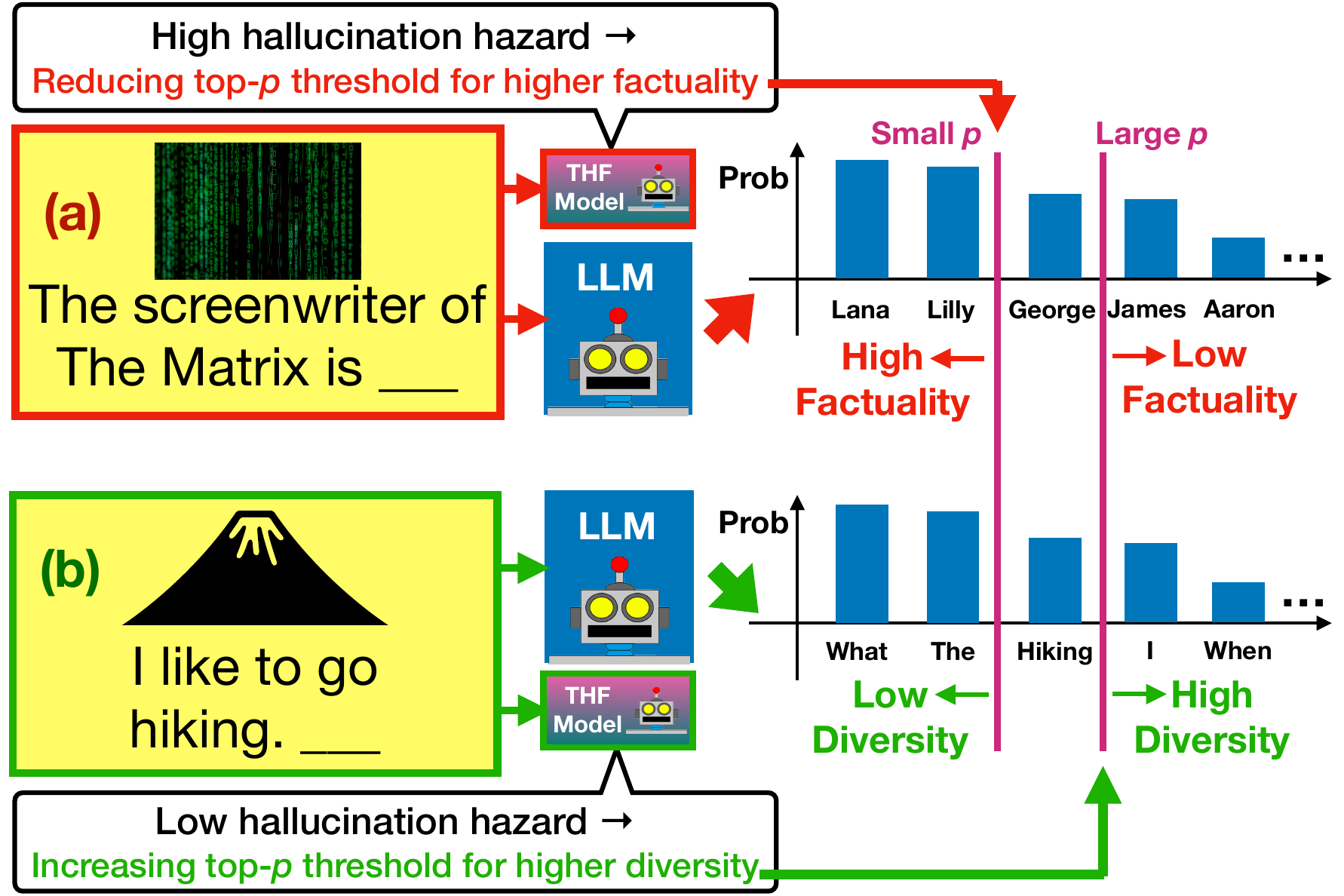}
  \captionof{figure}{(a) For the factual question, only a few next tokens are correct but the target LLM assigns high probabilities to many tokens, so our THF model predicts the next token from the LLM is likely to be incorrect if using a large $p$ threshold. (b) For the beginning of a sentence, many tokens could be used, so our THF model predicts that sampling from more tokens increases the diversity without hurting the factuality.}
  \label{fig:first_sampling}
\end{minipage} \; %
\begin{minipage}{.4\textwidth}
  \centering
  \includegraphics[width=1\linewidth]{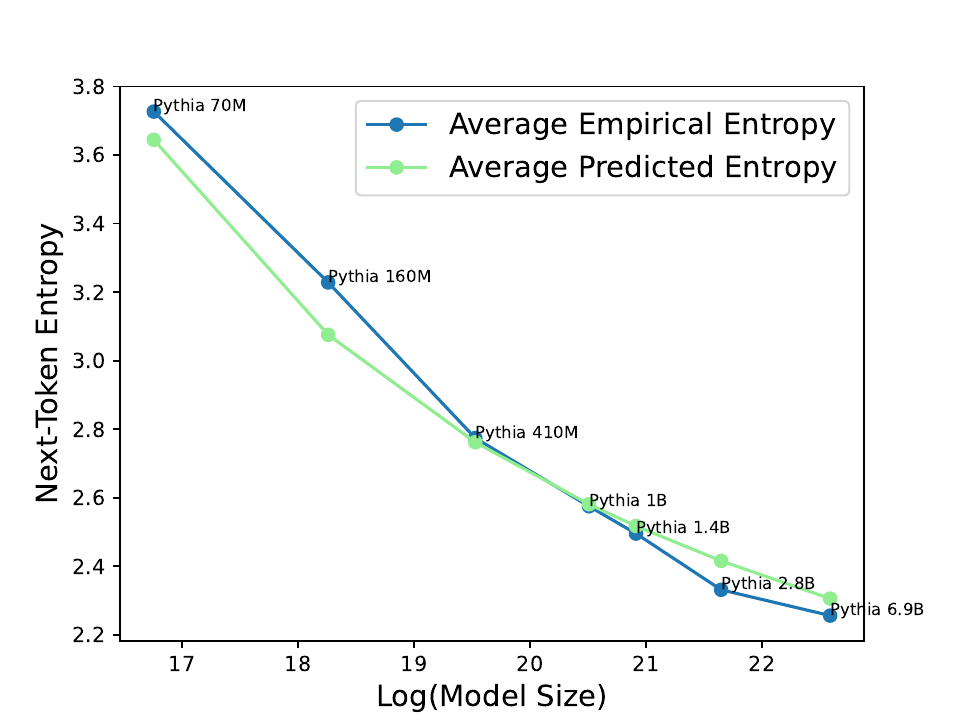}
  \captionof{figure}{The entropies of the Pythia’s distributions versus the model size in a logarithmic scale. The entropies are averaged across all tokens in a Wikipedia subset. The blue entropy decay curve plots actual entropies from Pythia LMs; the green curve is the entropies predicted by our THF model. }
  \label{fig:ent_decay_vis}
\end{minipage}
\end{figure}

% \begin{figure}[t!]
% \centering
% \includegraphics[width=0.5\linewidth]{figs/first_sampling_v2.pdf}
% \caption{We propose to dynamically change the $p$ threshold in top-$p$ sampling using a tiny model called THF (\textbf{T}oken-level \textbf{H}allucination \textbf{F}orecasting) model. (a) Only a few next tokens are correct but LLM assigns high probabilities to many tokens, so THF model predicts the next token from LLM is likely to be incorrect if using a large $p$ threshold. (b) Many next tokens are correct, so THF model predicts that sampling from more next tokens increases the diversity without hurting the factuality.}
% %\mbc{nice fig! but `real lm' is sounding like it's a real vs./ unreal LM? so we should change REAL to a slightly different acronym? also highlight the red and green arrows coming from the left to right a bit bolder/thicker}} 
% \label{fig:first_sampling}
% \end{figure}

Hallucination is a major problem that limits the applications of LLMs (large language models), especially in open-ended generation tasks~\citep{zheng2023does, huang2023survey,tonmoy2024comprehensive,sun2024trustllm}.  %\mbc{open-ended tasks or open-ended generation tasks? and add 1-2 more citations apart from our survey paper here?} 
Recent studies\footnote{\citet{burns2022discovering,li2023inference,azaria2023internal,slobodkin2023curious,ch2023androids} show that we can predict hallucination based on its internal states and \citet{agrawal2023language,guan2023language,manakul2023selfcheckgpt,zhang2023language,varshney2023stitch} show that a LLM can sometimes improve itself by editing or verifying its own answer. } show that a LLM often ``knows'' if it is hallucinating. The findings suggest that the decoding methods of LLMs are major sources of the hallucination.

Sampling is one of the most widely used decoding strategies in LLM due to its simplicity, efficiency, %low cost 
and high generation diversity~\citep{holtzman2019curious,hewitt2022truncation,meister2022typical}. 
%\mbc{add some basic sampling citations} 
Nevertheless, sampling often intensifies an LLM's hallucination problem. \Cref{fig:first_sampling} (a) illustrates a simple example. When an LLM is uncertain about who is the screenwriter of a movie, the next-token distribution usually has a high entropy, where some incorrect answers receive high probabilities. Recent studies show that hallucination often happens as the result of such high-entropy distribution and/or the lower probabilities of the sampled tokens \citep{van2022mutual,marfurt-etal-2022-corpus,manakul2023selfcheckgpt,rawte2023troubling,varshney2023stitch}.

%, which further motivates the studies of decoding strategies for LLMs

%Looking at \Cref{fig:first_sampling} (a) and (b) more closely, we observe that the next token of (b) is inherently uncertain while the next token of (a) should have a lower ``inherent uncertainty''. In \Cref{fig:ent_decay_vis}, we also observe that the entropy of a larger LM's distribution tends to be smaller. As LLM's model size becomes larger, the entropy of its distribution should be closer to the inherent uncertainty, so we should be able to extrapolate the entropy decay curve and estimate the inherent uncertainty by an inferred asymptotic entropy, which comes from an imaginary LLM with an infinite size.
%it is very challenging for the existing methods to effectively 

%try to adjust the 

%To address the issue, various thresholding methods have been proposed, and 
Nucleus (top-$p$) sampling~\citep{holtzman2019curious} is one of the representative methods\footnote{OpenAI provides top-$p$ sampling at \url{https://platform.openai.com/playground?mode=chat}.} proposed to alleviate the issue. By decreasing the constant global $p$ threshold, we can trade the generation diversity for higher factuality~\citep{dziri2021neural,lee2022factuality,aksitov2023characterizing}. For example, \Cref{fig:first_sampling} shows that a lower $p$ threshold could reduce the chance of sampling the incorrect writer names in (a), but it would also eliminate the legitimate starts of the possible next sentences in (b). This tradeoff limits nucleus sampling's ability to generate both high diversity and high factuality outputs. 
Some existing methods such as typical~\citep{meister2022typical} and eta~\citep{hewitt2022truncation} sampling are proposed to adjust the
threshold by characterizing the token-wise distributions of LLM. However, this distribution alone is often not enough to detect the hallucination. For example, both distributions in \Cref{fig:first_sampling} are similar but the high entropy of (a) arises due to the LLM's own limitation while that of (b) arises due to the task ``inherent uncertainty''.

\begin{figure*}[t!]
\centering
\includegraphics[width=1\linewidth]{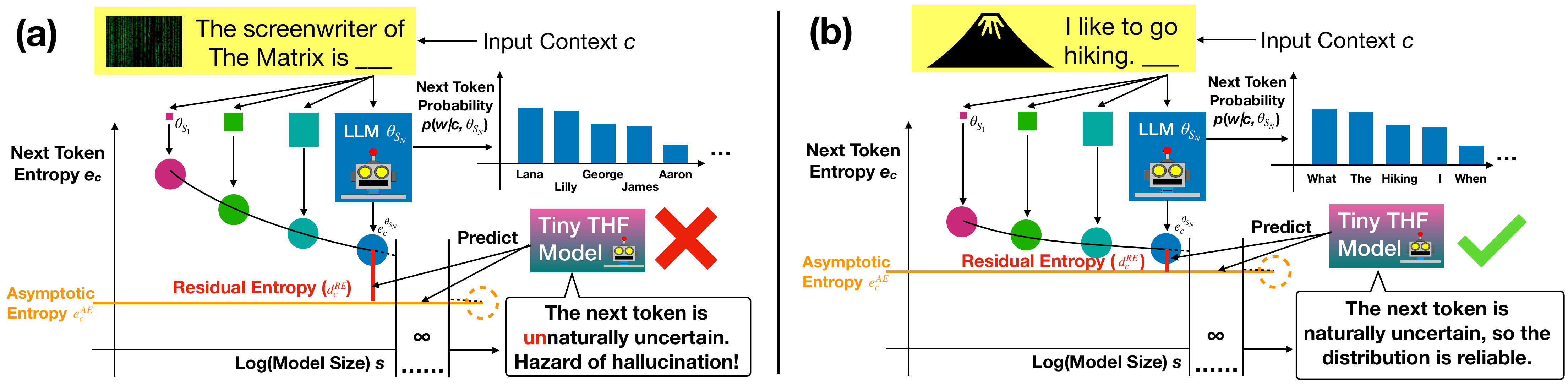}
\caption{Given the input context, the LLMs with different sizes generate the next-token distributions. By extrapolating the curve using a tiny THF model, we estimate the asymptotic entropy, the entropy from an imaginary LLM with an infinite size (i.e., the inherent uncertainty of the next token), and compute the residual entropy as a measurement of the hallucination hazard. (a) The LLM's entropy is much higher than the asymptotic entropy. This implies that the LLM is more uncertain than it should be and thus likely to hallucinate next. (b) LLM's high entropy is fine because the next token is inherently uncertain.}
\label{fig:asymptotic_entropy}
\end{figure*}

In this paper, we tackle this problem from a brand-new angle: estimating inherent uncertainty by extrapolating the entropy of LLMs with different sizes. We empirically observe that the entropy of a larger LM's distribution tends to be smaller as shown in \Cref{fig:ent_decay_vis}. 
As LLM's model size becomes larger, the entropy of its distribution should be closer to the inherent uncertainty. As a result, we can extrapolate the entropy decay curve to estimate the asymptotic entropy from an imaginary LLM with an infinite size, which approximates the inherent uncertainty. For example, for the questions discussed in \Cref{fig:asymptotic_entropy} (a), the LLM tends to be more certain about the answer as the size of LLM increases, so we can expect the asymptotic entropy to be low. In contrast, the entropies from different model sizes for the example in \Cref{fig:asymptotic_entropy} (b) should be similar, so we can infer that the next token distribution should have a high asymptotic entropy.

Based on this insight, we propose a tiny unsupervised method to predict the hazard of generating a nonfactual next token, called THF (\textbf{T}oken-level \textbf{H}allucination \textbf{F}orecasting) model. As shown in \Cref{fig:asymptotic_entropy}, we parameterize the decay curves of next-token entropies for LLMs and use the THF model to predict the curve parameters. Next, the THF model estimates the LLM's hallucination hazard by computing the difference between the asymptotic entropy and the LLM's entropy, which we call the residual entropy (RE). If the LLM is much more uncertain than it should be (i.e., the LLM's entropy is much larger than the asymptotic entropy), the THF model would forecast a high RE and hence a high hallucination hazard.
%LLM's distribution is not trustworthy and the predicted hallucination hazard should be high.

Relying on the residual entropy predicted by our THF model, we propose a novel context-dependent decoding method for open-ended text generation, which we call `REAL (\textbf{R}esidual \textbf{E}ntropy from \textbf{A}symptotic \textbf{L}ine) sampling'. REAL sampling %enjoys both high diversity and high factuality by dynamically 
adjusts the $p$ threshold in the top-$p$ (nucleus) sampling based on the forecasted hallucination hazard. For example, in \Cref{fig:first_sampling} (a),  the THF model learns that a movie usually does not have many credited screenwriters but the LLM's distribution entropy is high, so REAL sampling should use a lower threshold to mitigate the hallucination. On the other hand, in \Cref{fig:first_sampling} (b), THF model learns that the given prompt can be completed 
%a sentence could be started 
in many different ways, so REAL sampling should increase the threshold to boost the generation diversity.

%Although the ideal distribution is unknown in reality, we can estimate the asymptotic entropy by extrapolating the entropies of LLMs with different sizes. 

%\mbc{say a bit about how we prove} 
%that the threshold from REAL sampling would be smaller than the ideal threshold (raised to a constant power determined by the hyperparameter of REAL sampling) by bounding the entropies using the ideal threshold, if LLM's distribution of the top tokens is ideal and the residual entropy is estimated accurately.

To the best of our knowledge, REAL sampling is %not only based on clear intuitions but also 
the first sampling method that is tightly bounded by the ideal threshold that separates all the factual and nonfactual next tokens without making assumptions on the distribution of the nonfactual next tokens. 
%\violet{what does this `ideal threshold' mean?} 
%without the assumptions on the noise pattern of LLM’s distribution. \violet{what does `the assumptions on the noise pattern of LLM’s distribution' mean?}
Besides enjoying the theory guarantee, REAL sampling achieves large and robust empirical improvements in various tasks.
In our main experiment, we follow the evaluation protocol in \textsc{FactualityPrompts}~\citep{lee2022factuality} and find that %First,  %and use the relevant Wikipedia pages to verify its factuality. 
%the continuation of a sentence prompt 
sentences generated by Pythia 6.9B LLM~\citep{biderman2023pythia} with our REAL sampling contains less hallucination and less duplicated n-grams in both in-domain and out-of-domain settings. %(e.g., using THF model trained on Pythia family to improve OPT-6.7b~\citep{zhang2022opt}). 
%Our method reaches the high factuality of greedy decoding (i.e., $p=0$) and the high diversity of top-$p$ sampling with $p=0.4$ at the same time.
%After combined with contrastive decoding~\citep{li2022contrastive}, our factuality improvement could even be doubled. 
Our human evaluation indicates that REAL sampling not only improves the factuality but also informativeness, fluency, and overall quality.  
%compared to top-$p$ sampling. We can also combine REAL sampling with other decoding methods to achieve state-of-the-art performances. 
%Moreover, we show that the THF LM trained on Pythia family could also improve both diversity and factuality of OPT-6.7b~\citep{zhang2022opt} and OpenLLaMA2-7b~\citep{openlm2023openllama}.
%improves the factuality and diversity of top-$p$ sampling simultaneously. 
We further evaluate the generality of the THF model and show improved performance on several hallucination detection tasks. We plan to publicly release our code. 

%We theoretically prove that the threshold from REAL sampling would be smaller than the ideal threshold if LLM's distribution of the top tokens is ideal and the residual entropy is estimated accurately.

%use its predictions as features for detecting hallucination in open-ended text generation, and show improvement in three hallucination detection datasets. 
%\todo{Mohit: add some code as supplementary too (anonymized)}

Overall, our main contributions include
\begin{itemize}[leftmargin=.1in,topsep=0pt]
\setlength\itemsep{-0.1em}
%\vspace{-0.1em}
    \item We propose REAL sampling, a context-dependent sampling method that relies on a THF model to predict the asymptotic entropy of an infinitely large LLM to help decide the sampling threshold.
    \item We theoretically prove that the threshold from our REAL sampling is upperbounded by the ideal value if the top predicted tokens are ideal and the residual entropy is estimated accurately.
    %under reasonable assumptions.
    \item We demonstrate that the tradeoffs between factuality and diversity exist in the 9 state-of-the-art unsupervised sampling methods such as typical sampling~\citep{meister2022typical} and our REAL sampling can consistently boost their factuality given the same diversity, and vice versa. Furthermore, we conduct comprehensive analyses on the THF model and REAL sampling, including evaluating our design choices and their generality using hallucination detection tasks.
    %\item We demonstrate that the output of THF model is also an effective unsupervised signal for hallucination detection.
    %threshold from our REAL sampling is smaller than the ideal threshold under reasonable assumptions.
\end{itemize}

\section{Preliminary and Motivation}

Given a context $c$ and a next token candidate $w$ in a vocabulary $V$, a LLM ($\theta$) outputs the next token probability $p(w|c, \theta)$. Assuming $w^c_i$ is the $i$th token with the highest probability given the context $c$, top-$p$ (nucleus) sampling first determines the number of tokens $J$ by
\begin{equation}
J = \max_I( \{ I | \sum_{i=1}^I p(w^c_i|c, \theta) \le t^p \} ).
\label{eq:top_p}
\end{equation}
Then, it sets the probabilities from $w_{J+1}$ to $w_{|V|}$ to 0 and re-normalizes the distribution of the top $J$ tokens. In top-$p$ sampling, $t^p$ is a fixed global hyperparameter. 

As illustrated in \Cref{fig:first_sampling}, lower $t^p$ would lead to a better factuality but worse diversity. In practice, many users would like to select from diverse responses.
Furthermore, diverse and factual responses could also improve LLM's performance in reasoning tasks~\citep{li2022competition,wang2022self,bertsch2023s,yao2023tree,naik2023diversity}. %When using these techniques, LLM's generation diversity and factuality are both very important. 
%researchers recently discovered that we can improve LLM's reasoning performance by sampling multiple responses
If we can estimate the hallucination possibility of the next token, we can have a better context-dependent $t^p$. 

It is notoriously challenging to estimate the hallucination likelihood of each token in general open-ended text generation tasks. One common strategy is to annotate if each generated token is factual and learn a classifier through supervised learning~\citep{zhou2021detecting}. However, this approach has several drawbacks. First, human annotators often need to take a very long time to check if the generated text is factual, especially in an open-ended generation task, and provide token-level annotation. Second, due to the expense of getting the labels, the classifier is often trained using a few domain-specific examples that are generated by a specific LLM. Therefore, the classifier might not generalize well in other domains, other languages, or other LLMs. This motivates us to develop an unsupervised hallucination forecasting model that only needs the LLMs with different sizes. Then, we can apply our method to any domain, any language, and any LLM without the expensive human annotations.
%train our model using any domains, languages, and LLMs to ensure its output quality in the applications of interest. 

%supervised detection method
%So many domains

%every LLM is different 
%Given the same context, some LLMs are unlikely to output the incorrect answers while some LLMs are prone to have hallucination.

%The method can easily work well on all the language, including the ones where the entities are hard to detect
%Some language would put the object earlier
%we can have almost infinite amount of training data

%In some applications such as code generation, math QA, or story generation, factuality and diversity are both very important. 

%if we want to change the p, we need to estimate the hallucination probability

\section{Method}
\label{method}

As the LLMs get larger, their performances increase at the cost of higher inference expense, so an institute often trains LLMs (e.g., GPT-4 family~\citep{achiam2023gpt}) with different sizes using the same training data to let the users balance the cost and quality. We denote the parameters of a LLM family as $\{\theta_{s_1}, \theta_{s_2}, ... \theta_{s_N} \}$, where $s_n$ is the size of $n$th model in a logarithmic scale. 
In this paper, we focus on improving the generation of the largest LLM ($\theta_{s_N}$) in its family that can fit into our GPU memory. 
%In this paper, we focus on jointly improving both factuality and generation diversity of the LLM; we experiment with medium sized LLMs but note that our approach is applicable to the largest LLM in the model family. 

In this section, we leverage the LLM family to train a THF model, which aims at predicting the entropy of the ideal (ground-truth) distribution without actually knowing the ideal distribution. In \Cref{sec:extrapolating}, we first parameterize the entropy decay curve of each next token prediction to predict asymptotic entropy (AE). In \Cref{sec:REAL_LM}, we introduce the architecture of the THF model and how it learns to predict the residual entropy (RE). Finally, we describe REAL sampling, our context-dependent token truncation method based on the THF model in \Cref{sec:REAL_sampling}.

\subsection{Parameterization and Extrapolation of the Entropy Decay Curve}
\label{sec:extrapolating}

As we see in \Cref{fig:asymptotic_entropy}, the asymptotic entropy (AE) $e_c^{AE}$ is the entropy of the next-token distribution from an infinitely-large LLM ($\lim_{s \to \infty } \theta_s$). Formally, we define $e_c^{AE}$ as
\begin{equation}
\lim_{s \to \infty } e_c^{\theta_s} = \lim_{s \to \infty } \sum_w p(w|c, \theta_s) \log( p(w|c, \theta_s) ).
\label{eq:symptotic_entropy}
\end{equation}
To simplify our discussion, we assume an ideal distribution exists and the LLM's output approaches the 
%the infinitely large LLM is powerful enough to output the 
ideal distribution as its size increases, so AE measures the next-token inherent uncertainty.

We cannot get the ideal distribution ($\lim_{s \to \infty } p(w|c, \theta_s)$) when training the LM to predict the next token, which is a crucial challenge of text generation~\citep{zhang2023mixce}. Consequently, we cannot compute $e_c^{AE}$ using \Cref{eq:symptotic_entropy}. Nevertheless, we can use the LLM family to get the pairs of the LLM size and its corresponding entropy $(s_i, e_c^{\theta_{s_i}})$ given each context $c$. Then, we can model the entropy decay by formulating it as an one-dimensional regression problem and estimate $e_c^{AE}$ by extrapolation. 

We propose to parameterize the entropy decay trend using a fractional polynomial~\citep{chang2020using}\footnote{In \Cref{sec:decay_func}, we discuss the influences of different parameterizations and $K$ values.}: 
\begin{equation}
     e_c(s) = z_c + b_c ( \frac{a_{c,0.5}}{ x_c(s)^{0.5} } + \sum_{k=1}^{K} \frac{a_{c,k}}{ x_c(s)^{k} }  ), 
\label{eq:frac_poly}
\end{equation}
where $s$ is the model size in a logarithmic scale, $x_c(s) = \max(1, q_c (s - g_c))$ is a normalized model size, $e_c(s)$ is our entropy prediction, and $a_{c,0.5}, a_{c,k}, b_c, q_c, g_c$, and $z_c$ are the parameters of the curve. All the parameters are non-negative to ensure the non-increasing property of $e_c(s)$. Since $\lim_{s \to \infty } e_c(s) = z_c$, we can estimate $e_c^{AE}$ using $z_c$.

Given a context $c$, one approach is to estimate all the $K + 5$ parameters by fitting the $(s_i, e_c^{\theta_{s_i}})$ on the fly. However, this approach has several problems. First, it is time-consuming to run all the LLMs in the family and fit the curve. Second, we often cannot get many $(s_i, e_c^{\theta_{s_i}})$ pairs and the entropy signal of LLMs could be noisy, so the parameter estimation is unstable especially if we want to use a large degree of fractional polynomial $K$. To address the problems, we propose to use a tiny LM to predict the parameters in the next subsection.

\begin{figure*}[t!]
\centering
\includegraphics[width=1\linewidth]{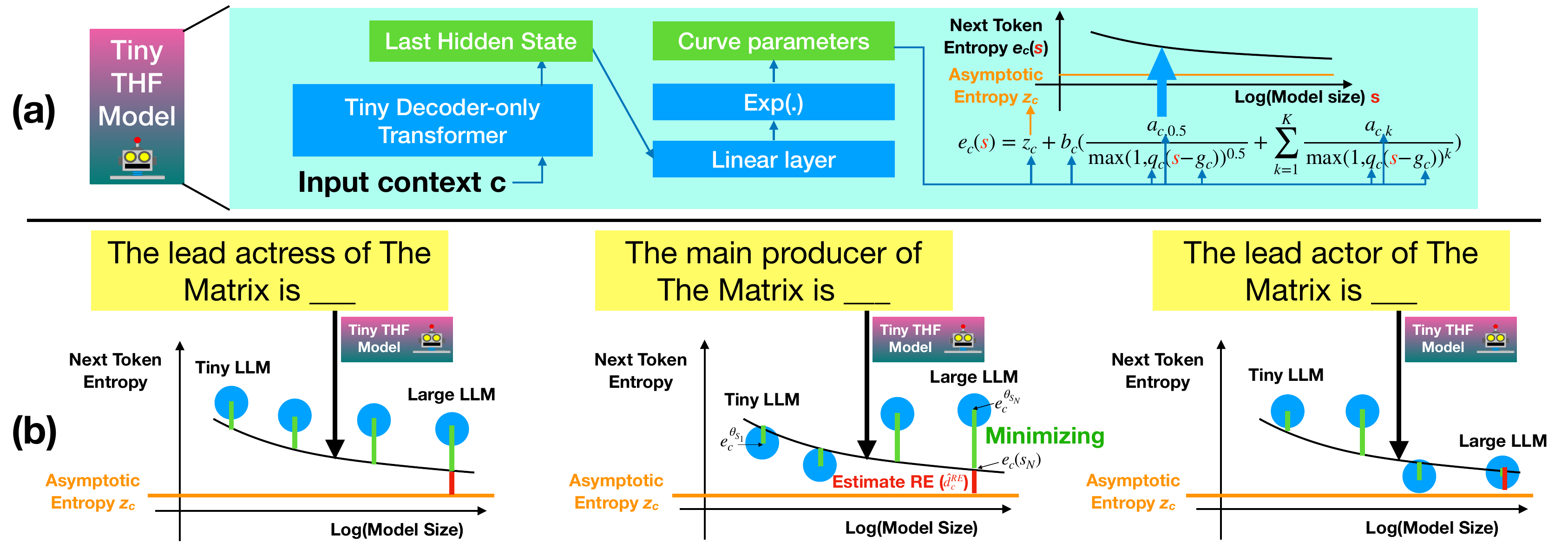}
\caption{The architecture and the training of the THF model. We use the THF model to predict the parameters of the entropy decay curves and we train the THF model by minimizing the distances between the predicted entropy curves and the empirical entropies from the LLM family.} 
\label{fig:curve_prediction}
\end{figure*}

\subsection{Residual Entropy Prediction using the THF Model}
\label{sec:REAL_LM}

The proposed THF (\textbf{T}oken-level \textbf{H}allucination \textbf{F}orecasting) model takes the input context and outputs the parameters of the entropy decay curve. As illustrated in \Cref{fig:curve_prediction} (a), the THF model projects the last hidden state of a pretrained tiny LM decoder\footnote{We choose a decoder-only transformer architecture because of its training efficiency. Our experiment uses the smallest LM, $\theta_{s_1}$, to initialize its weights. } to a vector with $K+5$ variables, which are passed through an exponential layer to ensure the positivity of the output parameter predictions.

We train the THF model by minimizing the square root of the mean square error between the predicted entropy and the actual entropy from LLMs. Specifically, our loss could be written as
\begin{equation}
L = \sqrt{ \frac{1}{|B|N} \sum_{c \in B} \sum_{i=1}^N (e_c^{\theta_{s_i}} - e_c(s_i) )^2 },
\label{eq:REAL_loss}
\end{equation}
where $B$ is a training batch. 

The entropy signal could be noisy\footnote{Compared to entropy, perplexity is even more noisy due to its dependency on the actual next token, so we choose to model the curves of entropy decay rather than perplexity decay.} even though all the LLMs are trained on the same corpus. For example, \Cref{fig:curve_prediction} (b), LLM's entropy of similar contexts are very different, and LLMs with a larger size sometimes have a larger empirical entropy. 

Using a tiny model to predict the entropy decay can not only reduce the inference time but also stabilize the parameter estimation. As a model gets smaller, it cannot memorize the small differences between similar input contexts \citep{biderman2023emergent}, so its predictions tend to be similar given the similar input. For example, when the tiny model receives three similar input contexts in \Cref{fig:curve_prediction} (b), if its hidden states and output parameters for the entropy decay curves are all identical, the gradient descent would encourage the predicted curves to be close to all the empirical entropy measurements of similar context inputs, which effectively increases the number of $(s_i, e_c^{\theta_{s_i}})$ pairs and reduces the influence of the noise in the entropy measurements.

As shown in \Cref{fig:asymptotic_entropy}, we use the THF model to predict residual entropy (RE) during inference as a measurement of the hallucination hazard: 
\begin{equation}
d_c^{RE} = e_c^{\theta_{s_N}} - e_c^{AE} \approx \hat{d}_c^{RE} = e_c(s_N) - z_c.
\label{eq:RE}
\end{equation}
Notice that although the entropy of LLMs, $e_c^{\theta_{s_N}}$, is measurable during the inference, we use the predicted entropy $e_c(s_N)$ to estimate the residual entropy $\hat{d}_c^{RE}$. This reduces the possible inconsistency between the LLM and the THF model and allows us to estimate the RE without actually running the LLM, which makes our method efficient in hallucination detection applications.

It is worth mentioning that we cannot expect a tiny model to very accurately estimate the inherent uncertainty at every position, which requires the knowledge that even the generation LLM cannot memorize (e.g., how many screenwriters every movie has). Nevertheless, the tiny THF model could still learn that the entropy should be higher at the beginning of a clause but lower if the next token should be something very specific such as an entity. In our experiment, we found that such a rough estimation is sufficient to improve the state-of-the-art decoding methods.

\subsection{REAL Sampling}
\label{sec:REAL_sampling}

We convert the residual entropy (RE) to the threshold between 0 and 1 for the cumulative probability in \Cref{eq:top_p} using
%To convert the residual entropy (RE) to the threshold between 0 and 1 for the cumulative probability in \Cref{eq:top_p}, we set the context-dependent threshold 
\begin{equation}
    \hat{t}_c^p = \exp(\frac{-\hat{d}_c^{RE}}{T}) = \exp(\frac{-\left(e_c(s_N) - z_c\right)}{T}),
\label{eq:REAL_sampling}
\end{equation}
where $T$ is our temperature hyperparameter used to control the tradeoff between factuality and diversity. When the $T$ is high, the $\hat{t}_c^p$ would be closer to $1$, so the generation diversity increases at the cost of the lower factuality.

Let's assume the top tokens from the LLM are factual and its top token distribution is correct (i.e., the same as the distribution of an infinitely large LLM after normalization). Then, there is an ideal threshold $g^p_c$ for the LLM, which sums the probabilities of all the top factual tokens (e.g., the lower $p$ in \Cref{fig:first_sampling} (a)), and we can derive an elegant relation between the ideal threshold and the threshold of REAL sampling from an ideal THF model.

%LLM predicts the distribution of all factual tokens correctly but ranks all the factual tokens on top of the non-factual tokens
%such that all the tokens with probabilities larger than this ideal threshold are factual and all the tokens below this threshold are nonfactual.  %(i.e., the probabilities of factual tokens are always larger than the probabilities of non-factual tokens).

\begin{theorem}
\label{thm:main_theorem}
If the residual entropy is estimated accurately (i.e., $\hat{d}_c^{RE}=d_c^{RE}$), and there is an ideal threshold $g_c^p$ such that the distribution of the top tokens above the threshold is ideal, then 
\begin{equation}
    t_c^p = \exp(\frac{-d_c^{RE}}{T}) \le (g_c^p) ^ \frac{1}{T}.
\label{eq:REAL_threshold_th}
\end{equation}
\end{theorem}
Please see our proof in \Cref{sec:proof}. That is, when the ideal threshold exists and our RE is accurate, our threshold $t_c^p$ is not larger than the ideal threshold raised to power $\frac{1}{T}$.

\section{Experiments}
\label{exp}

\begin{figure*}[t!]
\centering
\begin{subfigure}{.45\textwidth}
  \centering
  \includegraphics[width=1\linewidth]{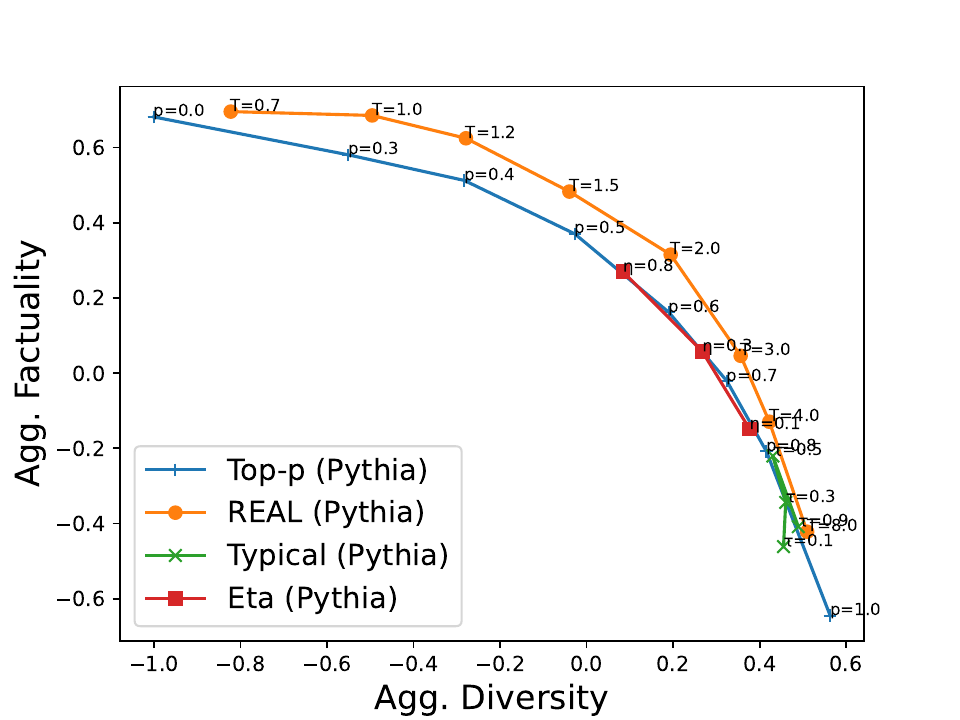}
  \caption{REAL vs Thresholding Methods}
  \label{fig:topp_factual}
\end{subfigure} \;\;%
\begin{subfigure}{.45\textwidth}
  \centering
  \includegraphics[width=1\linewidth]{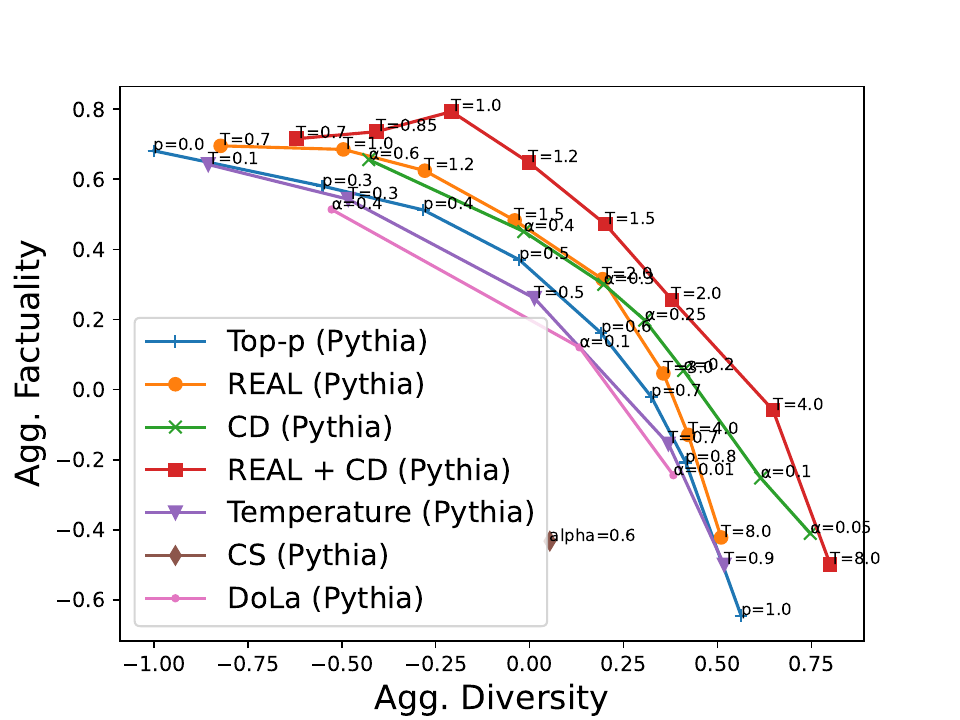}
  \caption{REAL vs Distribution Modifications}
  \label{fig:cd_factual}
\end{subfigure}
\begin{subfigure}{.45\textwidth}
  \centering
  \includegraphics[width=1\linewidth]{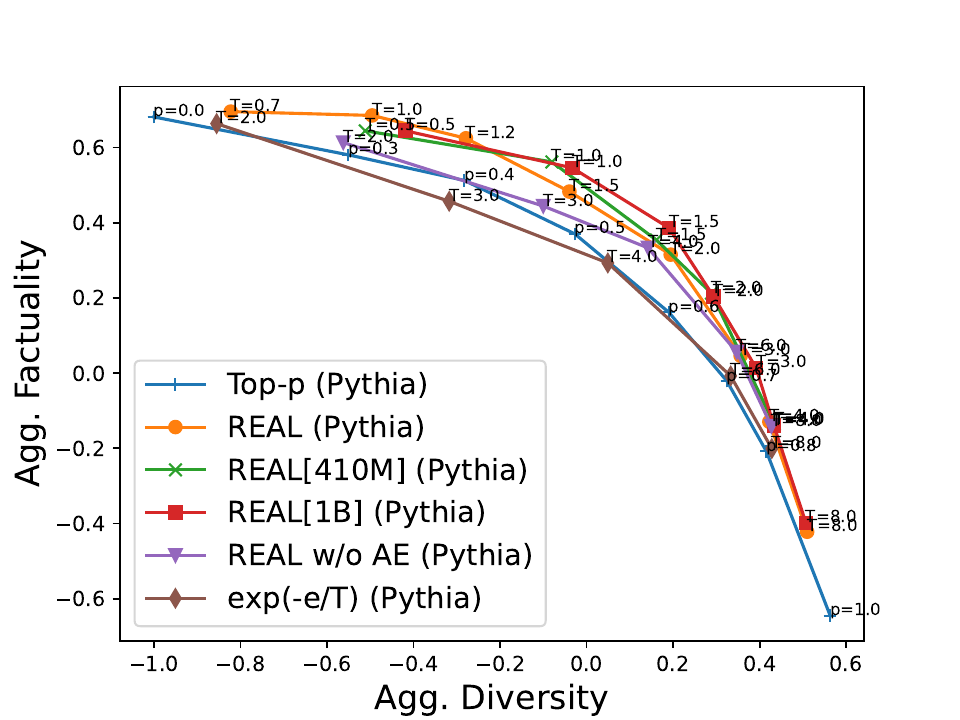}
  \caption{Comparison of the THF model sizes and ablation on our dynamic thresholding formula}
  \label{fig:ablation_factual}
\end{subfigure} \;\;%
\begin{subfigure}{.45\textwidth}
  \centering
  \includegraphics[width=1\linewidth]{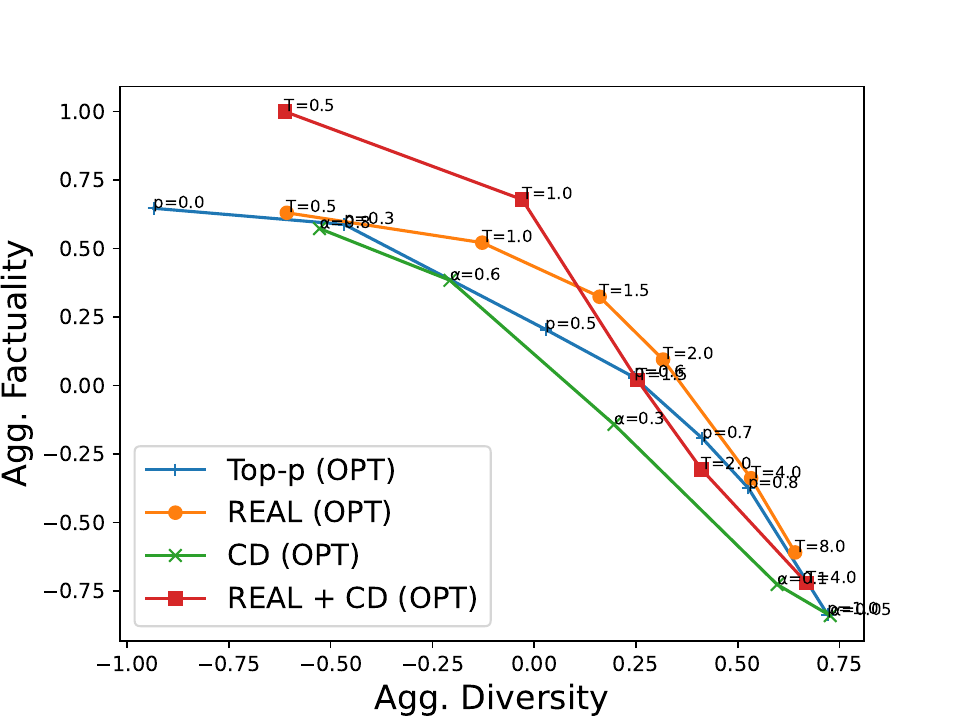}
  \caption{Using OPT-6.7b as the generation LLM and the THF model trained on Pythia. }
  \label{fig:opt_factual}
\end{subfigure}
\caption{Open-ended text generation performance comparison between REAL sampling and state-of-the-art unsupervised thresholding methods, including top-$p$~\citep{holtzman2019curious}, eta~\citep{hewitt2022truncation}, and typical~\citep{meister2022typical} sampling. The factuality and diversity are evaluated using the \textsc{FactualityPrompts} benchmark from~\citet{lee2022factuality}. We also conduct an ablation study and compare REAL sampling with the distribution modification methods including temperature sampling~\citep{ficler2017controlling}, contrastive search~\citep{su2022contrastive} (CS), contrastive decoding (CD)~\citep{li2022contrastive} and DoLa~\citep{chuang2023dola}. See more comparisons at \Cref{fig:K_abaltion}, \Cref{fig:decay_function}, and \Cref{fig:comp_gen_more}.}%In \ref{fig:opt_factual}, we show that THF model trained on Pythia could also improve the generation of OPT. %\violet{there are some inconsistency in the figure. E.g., figure (a),(b),(d) didn't mention the base model while only figure (c) did. Figure (d) just says `ablation', which is too vague. Can you comment on what you want to show in this ablation? Maybe you can split this figure into 2 figures, first how results on pythia and focus on comparisons with other sampling (you compared many different sampling approach, why do you only say top-p?) and CD, then move to other models and other ablations(?) Or, maybe swap (c) and (d) to first finish discuss all results on pythia model, then move to OPT. Also suggest update all `ours' in figure into `REAL'.}}
\label{fig:comp_gen}
\end{figure*}

We evaluate our REAL sampling using the \textsc{FactualityPrompts} benchmark~\citep{lee2022factuality} and using human evaluation in this section. In \Cref{sec:detection_eval} and \Cref{sec:decay_func}, we evaluate our methods in hallucination detection and entropy curve regression tasks, respectively. 

We use the de-duplicated variant of Pythia LLM series~\citep{biderman2023pythia} to train our THF model\footnote{Theoretically, our method could be applied to any LLM family with at least 3 different model sizes that are trained and tuned usng the same data. We choose Pythia because it has a high transparency and our computational resources do not allow us to run the LLMs with larger model sizes. }. By default, we use Pythia 6.9B as our LLM generation model ($\theta_{s_N}$) and the THF model is based on the transformer from Pythia 70M.  

\subsection{Retrieved-based Evaluation using \textsc{FactualityPrompts}}
\label{sec:lee_eval}

\expsec{Setup:}
%It is difficult to evaluate the factuality of the generated open-ended text even for humans. 
\citet{lee2022factuality} propose an evaluation benchmark, \textsc{FactualityPrompts}, for open-ended text generation that first lets different LLMs generate continuations of each prompt sentence and retrieves the relevant Wikipedia pages~\citep{hanselowski2018ukp} to evaluate the factuality of the generation. %The prompt sentences come from FEVER~\citep{thorne2018fever} and could be either factual or non-factual.

%We follow their evaluation protocol. 
\textsc{FactualityPrompts} provides 8k factual sentences and 8k nonfactual sentences from FEVER~\citep{thorne2018fever} as the prompts. We use the first 1k (non)factual prompts as our validation set to select the THF models and the rest 7k prompts as our test set. For each decoding method, the LLM generates 4 continuations for each prompt.

\expsec{Metrics:}
\textsc{FactualityPrompts} uses Entail$_R$ and NE$_{ER}$ to evaluate the factuality. Entail$_R$ is the ratio of the generated sentences entailed by the sentences in the relevant Wikipedia pages, while NE$_{ER}$ is the ratio of the entities that are not in the pages. Both metrics are shown to have high correlations ($\sim$0.8) with the hallucination labels from an expert~\citep{lee2022factuality}. \citet{lee2022factuality} use distinct n-grams (Dist-n)~\citep{li2016diversity} to measure the diversity across generations and repetition ratio (Rep)~\citep{holtzman2019curious} to measure the diversity within a generation. A good method should get high Entail$_R$ and Dist-n, but low NE$_{ER}$ and Rep. 

To compare the performances of methods in one figure, we first normalize all metrics from a generation LLM using max-min normalization and average the scores from both factual prompts and nonfactual prompts as Entail$_{Rn}$, NE$_{ERn}$, Dist-2$_n$, and Rep$_n$. Next, we define the aggregated metrics 
\begin{equation}
     \text{Agg. Factuality} = \text{Entail}_{Rn} - \text{NE}_{ERn}, \;\;\;\; \text{and} \;\;\;\; \text{Agg. Diversity} = \text{Dist-2}_n - \text{Rep}_{n}.
\label{eq:overall_factual}
\end{equation}
The scores of the original 4 metrics will be reported in \Cref{sec:org_metrics}.

\expsec{Methods:}
We compare multiple variants of our (ablated) methods described in detail below, with several state-of-the-art decoding methods such as top-$p$~\citep{holtzman2019curious}, eta~\citep{hewitt2022truncation}, typical~\citep{meister2022typical} sampling, temperature sampling~\citep{ficler2017controlling}, contrastive search~\citep{su2022contrastive} (CS), contrastive decoding (CD)~\citep{li2022contrastive} and DoLa~\citep{chuang2023dola}. 
%By default, we use Pythia 6.9B as our generation LLM, so we leave out the \textbf{(Pythia)} in each method name.
\begin{itemize}[leftmargin=.1in,topsep=0pt]
\setlength\itemsep{-0.3em}
%\vspace{-0.1em}
    \item \textbf{REAL (Pythia)}: REAL sampling using 70M THF model. 
    \item \textbf{REAL + CD (Pythia)}: Combining our methods with contrastive decoding (\textbf{CD})~\citep{li2022contrastive}. We first truncate the tokens using the threshold $\hat{t}_c^p$ in REAL sampling and apply the contrastive decoding (i.e., computing the probabilities of the top tokens using the logit differences between $\theta_{s_N}$ and $\theta_{s_1}$).
    \item \textbf{REAL[410M]} or \textbf{REAL[1B] (Pythia)}: REAL sampling using 410M or 1B THF model. 
    \item \textbf{REAL w/o AE (Pythia)}: Our method after removing the asymptotic entropy (AE) estimation as $\hat{t}_c^p = \exp(\frac{- e_c(s_N) }{T})$.
    \item \textbf{exp(-e/T) (Pythia)}: Instead of using THF model to predict the entropy of LLM, we estimate the entropy from LLM and set $\hat{t}_c^p = \exp(\frac{- e_c^{\theta_{s_N}}}{T})$. The method simply reduces the $p$ threshold whenever encountering a flat distribution (e.g., distributions in both (a) and (b) of \Cref{fig:first_sampling}).
    \item \textbf{* (OPT)}: In the methods, we replace the Pythia 6.9B with OPT-6.7b~\citep{zhang2022opt} as the generation LLM, respectively. Notice that the THF model is still trained using the Pythia family.
    %\item \textbf{* (OPT)} or \textbf{* (LLaMA)}: In the methods, we replace the Pythia 6.9B with OPT-6.7b~\citep{zhang2022opt} and OpenLLaMA2-7b~\citep{touvron2023llama,together2023redpajama,openlm2023openllama} as the generation LLM, respectively. Notice that the THF model is still trained using the Pythia family.
\end{itemize}

\begin{figure*}[t!]
\centering
\includegraphics[width=1\linewidth]{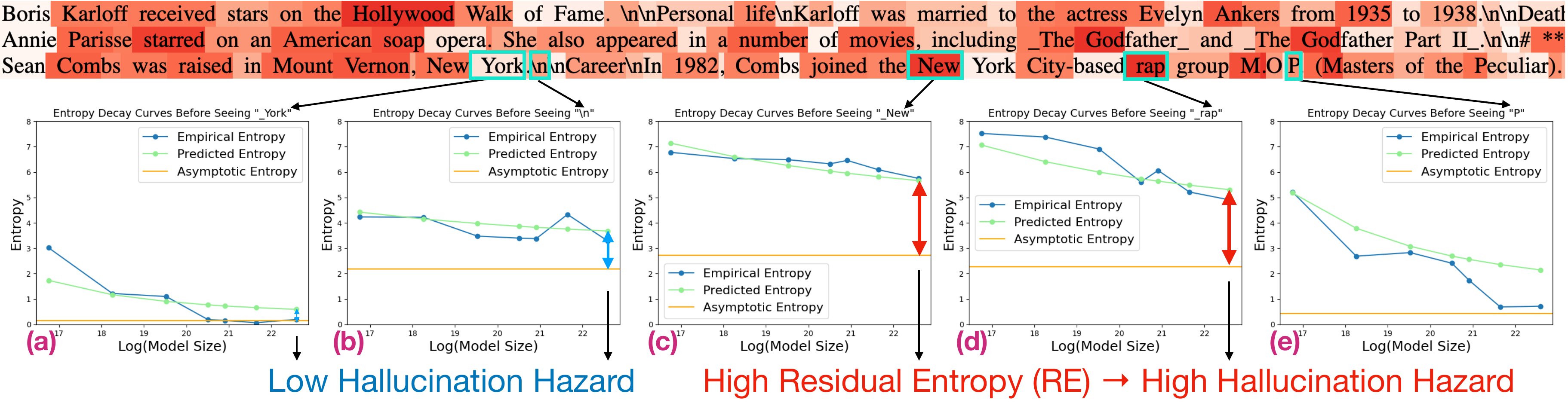}
\caption{The visualization of the estimated residual entropy ($\hat{d}_c^{RE}$) and entropy decay curves. The top three lines come from the first three testing factual prompts in \textsc{FactualityPrompts} and continuations generated by Pythia 6.9B. A darker red highlights a larger hallucination hazard ($\hat{d}_c^{RE}$) forecasted by our THF model based on the context before the position, and thus it leads to a smaller $p$ threshold in REAL sampling. The bottom figures (a)-(e) visualize the empirical entropy decay curves from five tokens in the third example, along with the corresponding curves predicted by our THF model and asymptotic entropies.}
\label{fig:RE_vis}
\end{figure*}

\expsec{Results:}
In \Cref{fig:topp_factual}, \textbf{REAL} sampling consistently outperforms \textbf{top-$p$}, \textbf{eta}, and \textbf{typical} sampling across the whole spectrum. Overall, we often improve the factuality more when the temperature $T$ is low (i.e., diversity is relatively low) probably because lower $T$ emphasizes the effect of $\hat{d}_c^{RE}$ in \Cref{eq:REAL_sampling}. Notice that some diversities actually come from hallucination, so it is hard to increase the diversity and the factuality at the same time, especially by only adjusting the truncation threshold without changing the distribution of LLM like our methods. In \Cref{fig:cd_factual}, \textbf{REAL + CD} is prominently better than using contrastive decoding \textbf{CD} alone, which shows that REAL sampling is complimentary with other distribution modification methods.

%Contrastive decoding (\textbf{CD}) modifies the logits rather than simply tuning the truncation threshold,  \Cref{fig:cd_factual} demonstrates that REAL sampling is always better than these baselines in some diversity intervals. Moreover, we can easily combine our method with these baselines and achieve even better performances. The improvement of \textbf{Ours + CD} over \textbf{CD} is especially prominent.
%\Cref{sec:more_baselines}

In \Cref{fig:ablation_factual}, the worse performance of \textbf{REAL w/o AE} (especially with low diversity) verifies the effectiveness of predicting asymptotic entropy (AE). The 70M THF model (\textbf{REAL}) performs similarly compared to the larger THF models (\textbf{REAL (410M)} and \textbf{REAL (1B)}), and using the LLM entropy predicted by THF model (\textbf{REAL w/o AE}) is much better than using the empirical LLM entropy (\textbf{exp(-e/T)}). These two results in our ablation study suggest that a tiny model indeed stabilizes the entropy decay curve prediction (see \Cref{sec:size_analysis} and \Cref{sec:ent_baseline} for more details). 

To evaluate our generalization capability, we use the THF model trained on the Pythia family to improve OPT. \Cref{fig:opt_factual} indicates that REAL sampling can still improve the factuality, and the improvement is especially prominent if CD is used. To explain the strong generalization ability across the serious misalignment between the training and testing objectives, we visualize the residual entropy (RE) from our THF model in \Cref{fig:RE_vis}. We observe that the residual entropy tends to be larger at the positions where the LLMs generally are more likely to hallucinate. For example, in (c), the THF model forecasts a high hallucination hazard for \textit{New}, which is the first token in an entity name. Nevertheless, we also observe that the THF model cannot always predict LLM's entropy accurately due to its small size, and (e) is an example. Finally, \Cref{sec:more_baselines} shows that REAL sampling can also improve OpenLLaMA2-7b~\citep{openlm2023openllama}, factual sampling~\citep{lee2022factuality}, and
top-$k$~\citep{fan2018hierarchical}.

%the next token is very likely to be the first token of an entity as we illustrated in \Cref{fig:asymptotic_entropy}. For example, predicting \textit{New} in (c) received a very high RE in the sequence. 

%The visualization in \Cref{fig:RE_vis} sugge that 
%regardless if \textbf{CD} is used. 
%Predicting the entity name is generally a major source of hallucination in LLMs, which explains why our method can generalize well even being trained and tested in different LLMs.

% \Cref{fig:RE_vis}

%Unlike Pythia, we found that \textbf{CD} performs significantly worse compared to \textbf{top-$p$} sampling when the generation model is OPT. Nevertheless, our method consistently improves the factuality regardless if \textbf{CD} is used.

%when the diversity is not very low
%while tends to be small near the beginning of the sentence

\begin{table}[t!]
\caption{Human evaluation for the open-ended generation. We highlight the better number between win and loss. $\dagger$ the win is significantly more than lose under Fisher's exact test~\citep{fisher1922interpretation} with $p=0.5$.}
\scalebox{0.85}{
\begin{tabular}{@{ }c|ccc|ccc|ccc|ccc@{ }}
\hline
              Model Comparision     & \multicolumn{3}{c|}{Overall} & \multicolumn{3}{c}{Factuality} & \multicolumn{3}{c|}{Informativeness}                 & \multicolumn{3}{c}{Fluency}                \\ \hline
                   & win           & tie  & loss & win            & tie   & loss  & win           & tie           & loss        & win           & tie           & loss          \\ 
REAL \; vs \; Top-$p$      & \textbf{29.5}$\dagger$ & 53.5 & 17   & \textbf{26}    & 53.5  & 20.5  & \textbf{26}   & 51            & 23 & \textbf{24.5}$\dagger$ & 58.5 & 17            \\
REAL+CD \; vs \; CD    & \textbf{27}$\dagger$   & 53   & 20   & \textbf{23.5}  & 53.5  & 23    & \textbf{26.5} & 52   & 21.5        & \textbf{19.5} & 64.5 & 16   \\
CD \; vs \; Top-$p$        & \textbf{34.5}$\dagger$ & 46   & 19.5 & \textbf{31}$\dagger$    & 49.5  & 19.5  & \textbf{33.5}$\dagger$ & 41.5 & 25 & \textbf{25.5} & 53.5          & 21            \\
REAL+CD \; vs \; Top-$p$ & \textbf{38}$\dagger$   & 44   & 18   & \textbf{35.5}$\dagger$  & 43    & 21.5 & \textbf{30.5}$\dagger$ & 45.5 & 24 & \textbf{27}$\dagger$   & 54.5 & 18.5 \\ \hline
 
\end{tabular}
}
\label{tb:human_exp}
\end{table}

\subsection{Human Evaluation}

To verify that our methods are still better from the humans' perspective, we ask the workers from Amazon Mechanical Turk (MTurk) to evaluate the factuality and the quality of the generated continuations using the Internet. Given 100 factual prompts in \textsc{FactualityPrompts}, we generate the next sentences using \textbf{Top-$p$} ($p=0.6$), \textbf{REAL} ($T=2.0$), \textbf{CD} ($\alpha=0.3$), and \textbf{REAL + CD} ($T=1.5$)  because the methods have similar diversity. In each task, the workers are asked to judge their factuality, fluency, informativeness, and overall quality. In the meanwhile, each worker needs to provide the URL(s), the statement(s) in the URL(s), and/or reason(s) that can justify their factuality annotations. Given a metric and a decoding method in a task, the worker provides a $1$ to $5$ score and we compare the scores to get the pairwise comparison results. Every task is answered by 2 workers.

\expsec{Results:} In \Cref{tb:human_exp}, our methods constantly outperform the corresponding baselines (i.e., \textbf{REAL} wins \textbf{Top-$p$} more and \textbf{REAL + CD} wins \textbf{CD} more) and the improvement of \textbf{REAL + CD} vs \textbf{Top-$p$} is larger than \textbf{CD} vs \textbf{Top-$p$}. The factuality evaluation results verify the effectiveness of the retrieved-based evaluation. Furthermore, our methods also achieve better informativeness and fluency. Consequently, we get the largest improvement in the overall metric.

\section{Related Work}
\label{related_work}

Due to the importance of LLM's hallucination problems, various mitigation approaches are proposed. For a comprehensive discussion, please see the recent surveys from \citet{huang2023survey,tonmoy2024comprehensive}. Nevertheless, as far as we know, no existing sampling methods that can improve both factuality and diversity in open-ended text generation without annotations or domain-specific heuristics/assumptions.

Some methods can improve the factuality by relying on some domain-specific assumptions. For example, \citet{lee2022factuality} assume the hallucination is more likely to appear at the latter part of a sentence. \citet{burns2022discovering} assume there is a set of statements that are either true or false. Several studies~\citep{van2022mutual,marfurt-etal-2022-corpus,chang2023kl,shi2023trusting,chen2023fidelity} assume that the generated text should be relevant to a source document. These methods might not be applicable to other domains (e.g., other languages or open-ended text generation tasks) and could (potentially) be combined with our method to achieve better performance in the specific domain (e.g., see \Cref{fig:lee_factual} and \Cref{sec:future}). 

In terms of high-level methodology, our method is related to some recent extrapolation-based methods in other applications. For example, \citet{das2024entropy} use a linear regressor to extrapolate the distribution of a deeper LM, \citet{lu2023open} extrapolate the probability distribution to get negative examples for text quality assessment, and \citet{zheng2024weak} extrapolate the weights of a LM after training on more preference data. However, none of them studies the threshold for sampling the next-token distribution.
%studies the thresholding problem of text generation.
%Recently, there are Extrapolation is 
%Methodologically,

%\todo{not open-ended text generation}
%assuming the existence of the source
%summarization
%\citep{van2022mutual,marfurt-etal-2022-corpus,li2022contrastive,chang2023kl}

%consider to add not decoding method (need some fine-tuning)

%reducing the diversity
%method similarity
%contrastive decoding
%\citep{chuang2023dola,li2022contrastive}
%increases the quality/factuality at the cost of higher reptition and lower diversity
%using the figure one as an example
%we don't change the logits, so our method is compatible with other contrastive decoding method (e.g., in RAG) or logit postprocessing such as sensitive word removal.

%propose factual sampling that simply uses high $p$ threshold in top-$p$ sampling near the beginning of sentence. 
%heuristics

%The problem of heuristics

%change the hidden state to improve the factuality
%\citep{burns2022discovering,li2023inference}
%we still need some training data or heuristics to identify the patterns of hidden states during hallucination.

%\citep{lu2023open}

%various studies only focus on reducing the hallucination without paying attention to diversity

%recent survey

%li2022contrastive,

\section{Conclusion}
\Cref{fig:first_sampling} suggests that it is difficult or sometimes even impossible in open-ended text generation tasks to predict the hallucination likelihood of the next token only based on the LLM's distribution without considering the inherent uncertainty of the task. In this paper, we demonstrate the feasibility of training a tiny model to forecast the hallucination hazard of LLM without supervision and domain-specific heuristics. Based on this finding, we propose REAL sampling along with its theoretical guarantee. Our comprehensive experiments indicate that most existing sampling methods cannot consistently outperform top-$p$ sampling in \textsc{FactualityPrompts}. In contrast, our proposed REAL sampling not only outperforms top-$p$ sampling but also can be combined with other decoding methods (e.g., contrastive decoding) to further reduce hallucination. We also demonstrate a THF model trained on one LLM family could be used to forecast/detect the hallucination from the LLM from another family, which highlight the strong out-of-domain generalization ability of our THF model. 
%To address the challenge, we propose REAL sampling based on a novel unsupervised hallucination forecasting model.
%We discover the difficulty of forecasting   
%We propose THF model, an unsupervised token-level hallucination forecasting model. 
%We leverage its prediction to improve the capability of an LLM in open-ended text generation and hallucination detection. 
%top-$p$ sampling is still a state-of-the-art method and 
%In the generation experiments, 
 %(contrastive decoding is an exception for Pythia but not for OPT).
%there is a tradeoff between diversity and factuality in all the decoding methods we tested and it is difficult to improve both metrics simultaneously by tuning the truncation threshold (e.g., eta sampling and typical sampling perform very similarly compared to top-$p$ sampling). 

%In the hallucination detection experiments, the result suggests that the predictions from THF model are an effective unsupervised feature across three datasets for detecting hallucination in open-ended text generation and the signal cannot be replaced by simple heuristics. 

%\begin{ack}

%\end{ack}

\bibliographystyle{abbrvnat}
\bibliography{example_paper}

%%%%%%%%%%%%%%%%%%%%%%%%%%%%%%%%%%%%%%%%%%%%%%%%%%%%%%%%%%%%

\newpage
\appendix
%\onecolumn

\addcontentsline{toc}{section}{Appendix} % Add the appendix text to the document TOC
\part{Appendix} % Start the appendix part
\parttoc % Insert the appendix TOC

\newpage

%\tableofcontents

\section{Overview}
%The $\mathtt{\backslash onecolumn}$ command above can be kept in place if you prefer a one-column appendix, or can be removed if you prefer a two-column appendix.  Apart from this possible change, the style (font size, spacing, margins, page numbering, etc.) should be kept the same as the main body.

In the appendix, we prove \Cref{thm:main_theorem} in \Cref{sec:proof}, 
%visualize the residual entropy from our THF model in \Cref{sec:visualization}, 
use the residual entropy to detect hallucination in \Cref{sec:detection_eval}, 
explore different parameterizations of the entropy decay function and different THF model sizes in \Cref{sec:decay_func}, 
provide more experiment results in \Cref{sec:more_results}, 
provide more explanation to the empirical observations in \Cref{sec:ent_baseline},
test REAL sampling for a creative writing task in \Cref{sec:creative}, provide our impact statements in \Cref{sec:impact}, 
describe our limitation and future work in \Cref{sec:future}, 
discuss why the entropy decreases as the model size increases in \Cref{sec:decrease_reason}, 
report more implementation details of our method in \Cref{sec:method_details}, 
report more experimental details in \Cref{sec:exp_details}.
%, and state the possible social impact of the work in \Cref{sec:impact}. 

\begin{figure}[t!]
\centering
\includegraphics[width=0.8\linewidth]{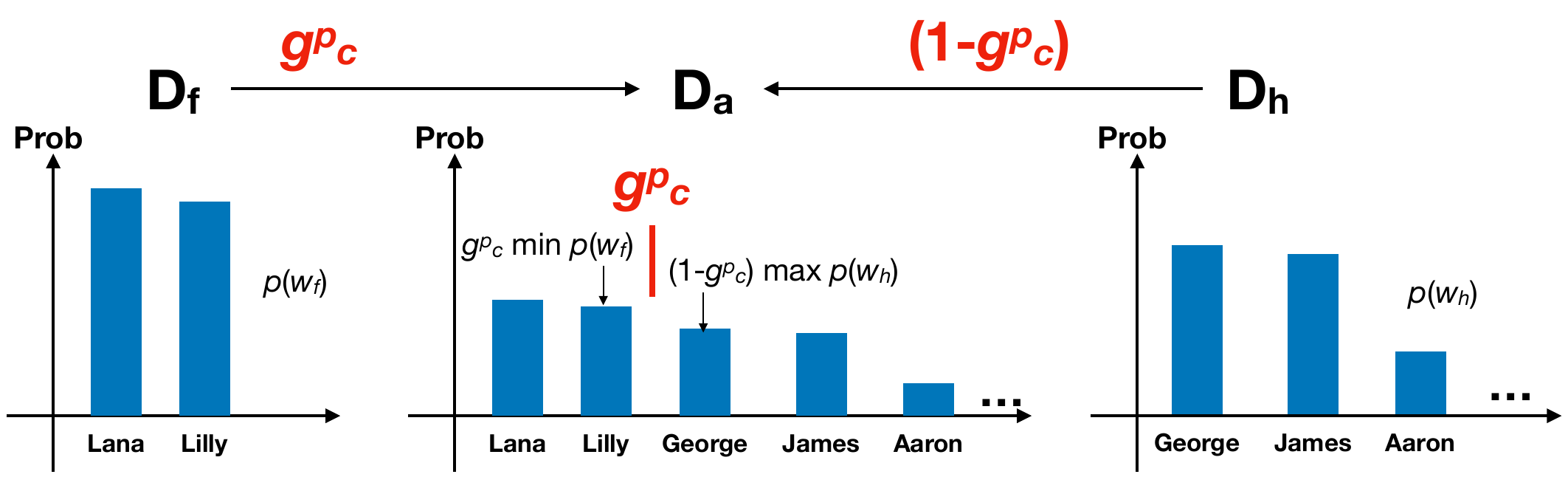}
\caption{Illustration of the notations used in \Cref{sec:proof}. All the next tokens are sorted based on its probabilities. } 
\label{fig:theory}
\end{figure}

\section{Proof of \Cref{thm:main_theorem}}
\label{sec:proof}

\begin{proof}
%\label{proof:thm}

To simplify our notations, we write the conditional probability $p(w|c)$ as $p(w)$ in the following derivation and \Cref{fig:theory} since every probability is conditioned on $c$.

In \Cref{fig:theory}, we illustrate our notations. One condition of \Cref{thm:main_theorem} is that the top token distribution is ideal, so we can decompose the next-token distribution $D_a$ into factual/ideal distribution $D_f$ (i.e., the distribution from an infinitely large LLM) and hallucination distribution $D_h$ that we want to truncate. We denote the factual token as $w_f$ and the hallucinated tokens as $w_h$. The ideal $p$ threshold that separates two distributions is $g^p_c$, so the probabilities of each factual token and hallucinated token in $D_a$ are $g^p_c \cdot p(w_f)$ and $(1-g^p_c) \cdot p(w_h)$, respectively. From \Cref{fig:theory}, we can see that
\begin{equation}
g^p_c \min_{w_f}( p(w_f) ) \ge ( 1 - g^p_c ) \max_{w_h}( p(w_h) ).
\label{eq:proof_cond}
\end{equation}

The condition of \Cref{thm:main_theorem} states that $\hat{d}_c^{RE}=d_c^{RE}$, so we know 
\begin{equation}
t_c^p = \exp(\frac{-d_c^{RE}}{T}) = \exp(\frac{Ent(D_f) - Ent(D_a)}{T}), 
\label{eq:proof_tp}
\end{equation}
where $Ent(D)$ is the entropy of the distribution $D$.

Based on the above two conditions, we can get
\footnotesize
\begin{align*} 
&- T \cdot \log(t_c^p) = Ent(D_a) - Ent(D_f) \\
& = - \sum g^p_c \cdot p(w_f)\log( g^p_c \cdot p(w_f)) - \sum (1-g^p_c) \cdot p(w_h)\log( (1-g^p_c) \cdot p(w_h)) - Ent(D_f) \\
& = g^p_c \cdot Ent(D_f) - g^p_c \log( g^p_c ) + (1-g^p_c) \cdot Ent(D_h) - (1 - g^p_c)  \log( 1 - g^p_c ) - Ent(D_f) \\
& = - (1 - g^p_c) \cdot Ent(D_f) + (1-g^p_c) \cdot Ent(D_h) + (1 - g^p_c) \log( g^p_c ) - (1 - g^p_c)  \log( 1 - g^p_c ) - \log( g^p_c ) \\
& = (1 - g^p_c) \left( Ent(D_h) - Ent(D_f) + \log( g^p_c ) - \log( 1 - g^p_c ) \right) - \log( g^p_c ) \\
& = (1 - g^p_c) \left( - \sum p(w_h)\log( p(w_h)) + \sum p(w_f)\log( p(w_f)) + \log( \frac{g^p_c}{1 - g^p_c} )  \right) - \log( g^p_c ) \\
& \ge (1 - g^p_c) \left( - \sum p(w_h)\log( \max_{w_h} p(w_h)) + \sum p(w_f)\log( \min_{w_f} p(w_f)) + \log( \frac{\max\limits_{w_h} p(w_h)}{\min\limits_{w_f} p(w_f)} ) \right) - \log( g^p_c ) \\
& = - \log( g^p_c )
\end{align*}
\normalsize

Therefore, 
\begin{equation}
    t_c^p = \exp(\frac{-d_c^{RE}}{T}) \le (g_c^p) ^ \frac{1}{T}.
\label{eq:REAL_threshold_th_proof}
\end{equation}

\end{proof}

\begin{table*}[t!]
\centering
\caption{Hallucination detection in open-ended text generation. A random forest classifier predicts the hallucination using the features from Pythia 6.9B LLM, Pythia 70M LM, and THF model. 1 Feature (6.9B per) refers to only using the perplexity of Pythia 6.9B to detect hallucination~\citep{muhlgay2023generating, varshney2023stitch}. The better performances in each section are highlighted. }
\scalebox{0.65}{
\begin{tabular}{c|cccccc|cc|cc|c}
                            \multicolumn{1}{r|}{Dataset $\rightarrow$}                  & \multicolumn{6}{c|}{Factor}                                                                            & \multicolumn{2}{c|}{TF ext}        & \multicolumn{2}{c|}{HaDes}       & \multirow{3}{*}{Avg} \\
                            \multicolumn{1}{r|}{Creation Method $\rightarrow$}          & \multicolumn{6}{c|}{Revising a Factual Sentence using ChatGPT}                                                                    & \multicolumn{2}{c|}{Template + Table} & \multicolumn{2}{c|}{BERT Infill} &                      \\
                            \multicolumn{1}{r|}{Subset / Size $\rightarrow$}            & \multicolumn{2}{c}{Wiki / 47025} & \multicolumn{2}{c}{News / 7663} & \multicolumn{2}{c|}{Expert / 355} & \multicolumn{2}{c|}{All / 9830}    & \multicolumn{2}{c|}{All / 1000}  &                      \\ \hline
                             Feature Subsets $\downarrow$ Metrics $\rightarrow$                  & 1-4 ACC         &  AUC       & 1-4 ACC        & AUC       & 1-4 ACC         &  AUC    & ACC   &  AUC                & ACC     & AUC                   &                      \\ \hline
1 Feature ($6.9B\_per$)    & 0.374           & 0.315          & 0.367          & 0.312          & 0.347           & 0.290          & 0.619  & 0.691           &  \textbf{0.528} & \textbf{0.599} &  0.444                \\
2 Features ($6.9B\_per$ + $heur\_ent$)    & \textbf{0.424}  & \textbf{0.322} & 0.359          & \textbf{0.313} & 0.347           & 0.300         & 0.624 & 0.700             & 0.503 & 0.581 &  0.447                \\
2 Features ($6.9B\_per$ + $RE$)        & 0.393           & 0.319          & \textbf{0.390} & 0.303          & \textbf{0.364}  & \textbf{0.320}  & \textbf{0.635} &  \textbf{0.711}  & 0.521 & 0.580  & \textbf{0.454}       \\ \hline
6 Freatures (6.9B and 70M)             & 0.490           & \textbf{0.341} & 0.432          & \textbf{0.326} & 0.534           & \textbf{0.356}& 0.654 & 0.754                     & \textbf{0.578} & 0.646           & 0.511                \\
All (6.9B, 70M, $RE$, and $AE$) & \textbf{0.498}  & \textbf{0.341} & \textbf{0.465} & \textbf{0.326} & \textbf{0.619}  & 0.346       & \textbf{0.671}   & \textbf{0.769}   & 0.565 & \textbf{0.669}           & \textbf{0.527}      
\end{tabular}
}
\label{tb:hallucination_detection}
\end{table*}

\section{Hallucination Detection for Open-ended Text Generation}
\label{sec:detection_eval}

Perplexity and entropy are widely used to detect the hallucination~\citep{van2022mutual,marfurt-etal-2022-corpus,muhlgay2023generating,manakul2023selfcheckgpt,rawte2023troubling,varshney2023stitch}. However, high perplexity or entropy could mean multiple correct answers instead of hallucination as in \Cref{fig:first_sampling} (b), so we test if the residual entropy (RE) and asymptotic entropy (AE) could be useful unsupervised signals for the hallucination detection tasks.

\expsec{Setup:} We test the features using three hallucination detection datasets: Factor~\citep{muhlgay2023generating}\footnote{\url{https://github.com/AI21Labs/factor} MIT license}, extended True-False dataset (TF ext)~\citep{azaria2023internal}\footnote{\url{https://github.com/balevinstein/Probes/} MIT license}, and HaDes~\citep{liu2022token}\footnote{\url{https://github.com/microsoft/HaDes} MIT license}. We train a random forest classifier to combine these unsupervised features from the input context and phrase/sentence.
%has hallucination labels on phrases and the other two datasets have sentence-level labels. 
%We first compute the unsupervised features from the input context and phrase/sentences and train a random forest to combine these features. 

%random forest.
%The results might not be better than fine-tuning a LM.
%Use less features -> generalize better

\expsec{Metrics:} 
The goal of all three datasets is to classify the text into either factual or nonfactual, so we can use the area under the precision recall curve (AUC) to measure the performances of classifiers. In the Factor dataset, one of the four sentence continuations is factual.
%each context has four sentence continuations and only  them is factual. 
Thus, we follow \citet{muhlgay2023generating} to measure the accuracy of detecting the factual sentence (1-4 ACC). In TF ext and HaDes, we also report the accuracy of the classifiers. To summarize the performance of each method, we report the average of all the scores. 

%AP@all

\expsec{Methods:} We consider the following features: 
%\begin{enumerate*}
\begin{itemize}[leftmargin=.1in,topsep=0pt]
\setlength\itemsep{-0.3em}
  \item Perplexity of Pythia 6.9B ($6.9B\_per$),
  \item Entropy of Pythia 6.9B ($6.9B\_ent$), 
  \item Perplexity of Pythia 70M ($70M\_per$), 
  \item Entropy of Pythia 70M ($70M\_ent$), 
  \item $\sqrt{6.9B\_per \cdot \max(0, 70M\_per - 6.9B\_per)} $ ($heur\_per$)
  \item $\sqrt{6.9B\_ent \cdot \max(0, 70M\_ent - 6.9B\_ent)} $ ($heur\_ent$)
  \item $\hat{d}_c^{RE}$ in \Cref{eq:RE} ($RE$)
  \item $z_c$ in \Cref{eq:frac_poly} ($AE$),
\end{itemize}
%\end{enumerate*}
where all features are averaged across the tokens in the input phrase/sentence, and $heur\_ent$ is a simple hallucination detection heuristic that approximates $RE$. If the next token is a hallucination, it should have a high LLM entropy and large entropy difference between the LLM and the smallest LM in the family as in \Cref{fig:asymptotic_entropy} (a).

Given a subset of the above features, we conduct an exhaustive feature selection to boost/stabilize the performance and train a random forest classifier with 100 estimators.

%using the code in HaDes~\citep{liu2022token} 

%+  (based on the testing scores)
%RF
%max depth=5, n estimators=100

%the heuristics using entropy difference
%All list of features
%compute average perplexity and residual entropy
%random forest %and logistic regression

%all other features
%the top three result rows of 

\expsec{Results:}
In \Cref{tb:hallucination_detection}, \textbf{2 Features ($6.9B\_per$ + $RE$)} usually outperforms \textbf{2 Features ($6.9B\_per$ + $heur\_ent$)} and \textbf{1 Feature ($6.9B\_per$)}, which indicates that adding the $RE$ features can improve the widely-used perplexity measurement of LLM~\citep{muhlgay2023generating, varshney2023stitch} and the improvement cannot be achieved by the heuristics that try to leverage the same signal. Similarly, compared to \textbf{6 Features (6.9B and 70M)}, the better performance of \textbf{All (6.9B, 70M, $RE$, and $AE$)} demonstrates that even letting the random forest combine all the features from the Pythia 6.9B and 70M, residual entropy (RE) and asymptotic entropy (AE) from our THF model still provide extra information for hallucination detection. 

\begin{figure*}[t!]
\centering
\begin{subfigure}{.33\textwidth}
  \centering
  \includegraphics[width=1\linewidth]{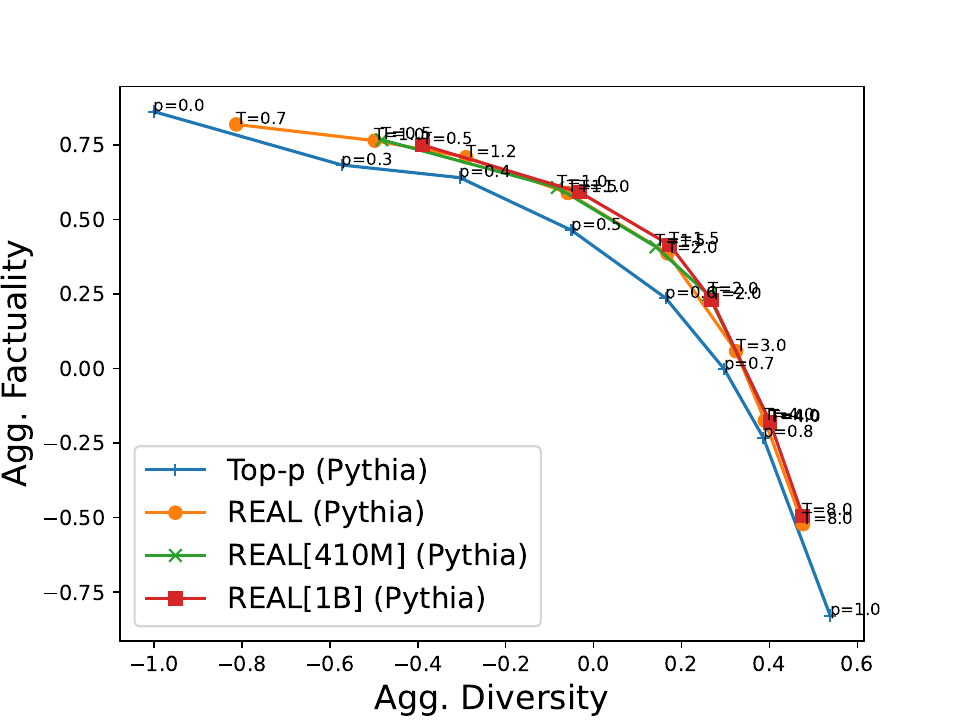}
  \caption{Factual + Nonfactual}
  \label{fig:size_all_per}
\end{subfigure}%
\begin{subfigure}{.33\textwidth}
  \centering
  \includegraphics[width=1\linewidth]{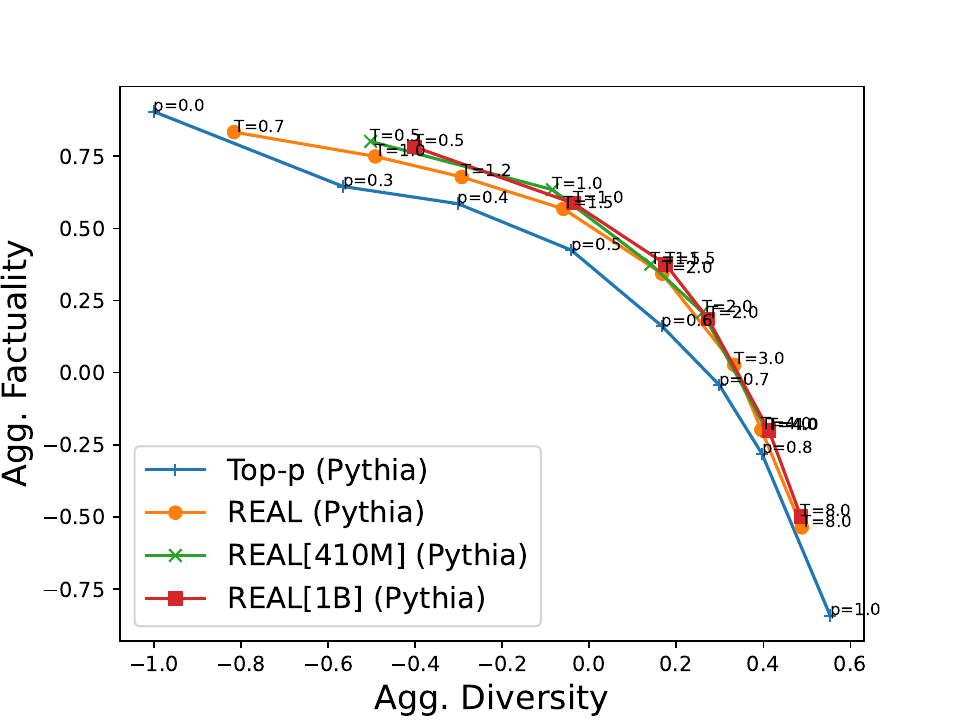}
  \caption{Factual}
  \label{fig:size_factual_per}
\end{subfigure}
\begin{subfigure}{.33\textwidth}
  \centering
  \includegraphics[width=1\linewidth]{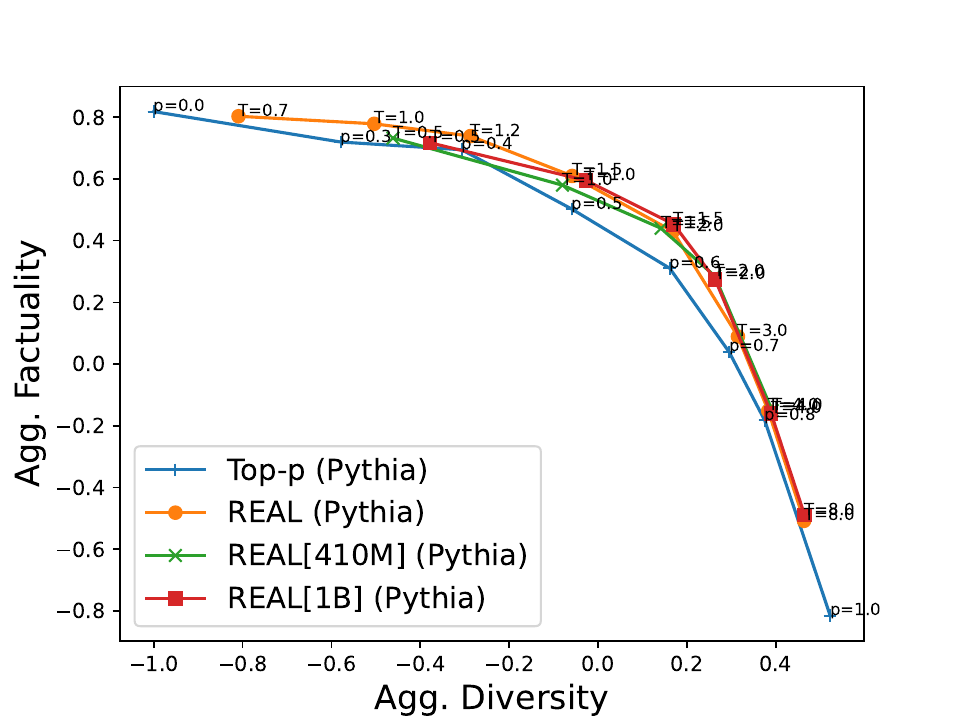}
  \caption{Nonfactual}
  \label{fig:size_nonfactual_per}
\end{subfigure}
\caption{Comparing different sizes of THF models. REAL means REAL[70M]. We evaluate 3 sentences after the prompt.  }
\label{fig:size_abaltion}
\end{figure*}

\section{Analyses of Entropy Decay Modeling}
\label{sec:decay_func}
We conduct a serious experiments to examine the influence of different design choices on our THF model, including its model size in \Cref{sec:size_analysis}, its $K$ value (i.e., highest order of fractional polynomial) in \Cref{sec:K_analysis}, and its fractional polynomial parameterization function in \Cref{sec:fun_analysis}. We also compare regresion error of predicting the LLM's entropy using different parameterization functions in \Cref{sec:regression}.

%Entropy Decay Functions

\subsection{Choice of the THF Model Size}
\label{sec:size_analysis}

We find that the performance is not sensitive to the choice of the THF model size. \Cref{fig:size_all_per} shows that the REAL sampling based on different sizes of THF models performs similarly in \textsc{FactualPrompts}. 
Nevertheless, we find that larger models perform slightly better given factual prompts in \Cref{fig:size_factual_per}, while the 70M model performs slightly better given nonfactual prompts in \Cref{fig:size_nonfactual_per}. Since THF is trained only on factual text, the results suggest that a larger model could perform better in an in-domain setting due to its more accurate entropy decay modeling. On the other hand, a smaller THF model is better at handling the noise both in the input context and in the entropy decay curves as illustrated in \Cref{fig:curve_prediction}, and hence has a strong out-of-domain generalization capability. 

%the smoothing effect illustrated 
%generalization ability

%To explore the performance of THF models that are larger than 410M, we train a THF model with 1B size. From \Cref{fig:K_abaltion}, we can see that all models perform similarly in FactualPrompts data. %The factuality of a THF model larger than 70M slightly increases on T=1.0 and T=1.5 while slightly decreases when T=0.5. Notice that 70M is the smallest size in the Pythia family. Since our THF model weights are initialized using the weights of pretrained Pythia LM, it is difficult for us to fairly compare the THF models smaller than 70M with our current THF models.

\begin{figure*}[t!]
\centering
\begin{subfigure}{.33\textwidth}
  \centering
  \includegraphics[width=1\linewidth]{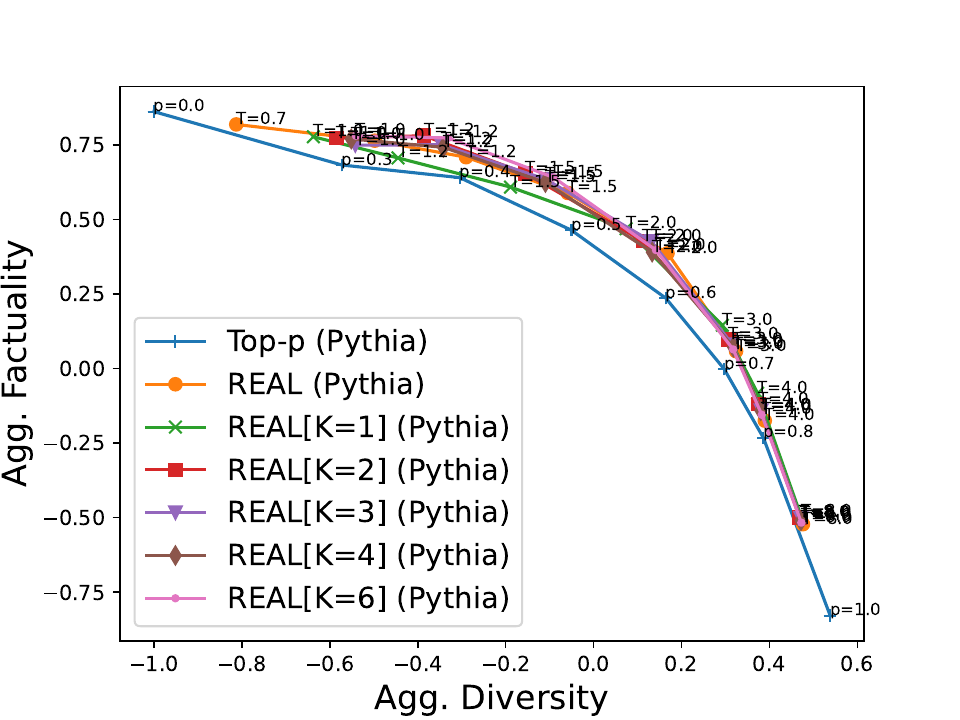}
  \caption{Factual + Nonfactual}
  \label{fig:K_all_per}
\end{subfigure}%
\begin{subfigure}{.33\textwidth}
  \centering
  \includegraphics[width=1\linewidth]{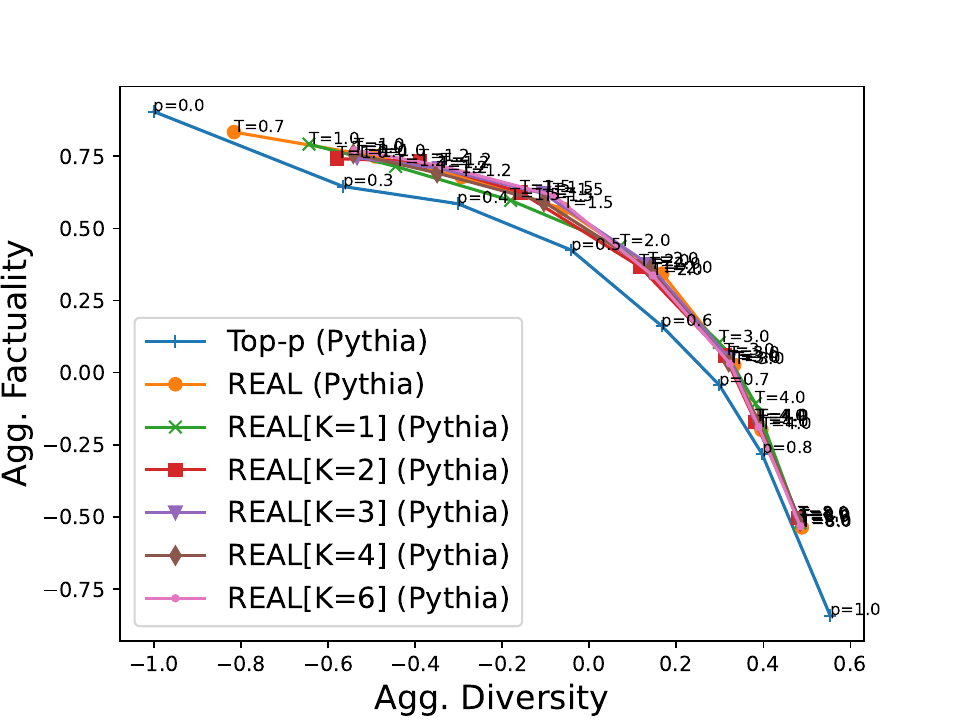}
  \caption{Factual}
  \label{fig:K_factual_per}
\end{subfigure}
\begin{subfigure}{.33\textwidth}
  \centering
  \includegraphics[width=1\linewidth]{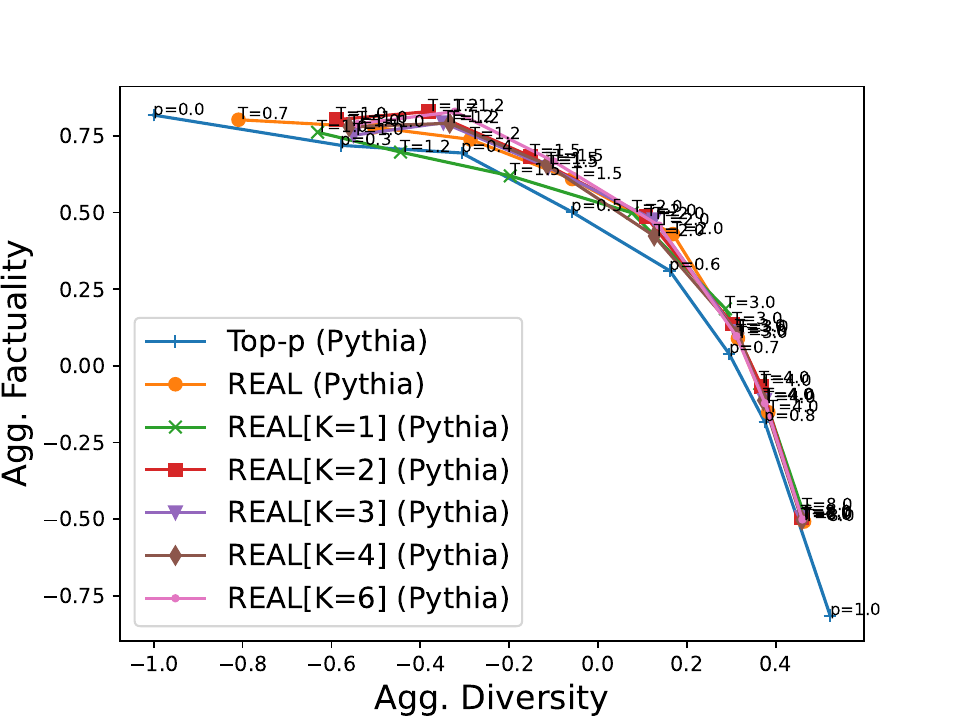}
  \caption{Nonfactual}
  \label{fig:K_nonfactual_per}
\end{subfigure}
\caption{Comparison of Pythia generation performance in FactualPrompt benchmark given different K (highest degrees of fractional polynomial). REAL means REAL[K=10]. We evaluate 3 sentences after the prompt. }
\label{fig:K_abaltion}
\end{figure*}

\begin{figure*}[t!]
\centering
\begin{subfigure}{.33\textwidth}
  \centering
  \includegraphics[width=1\linewidth]{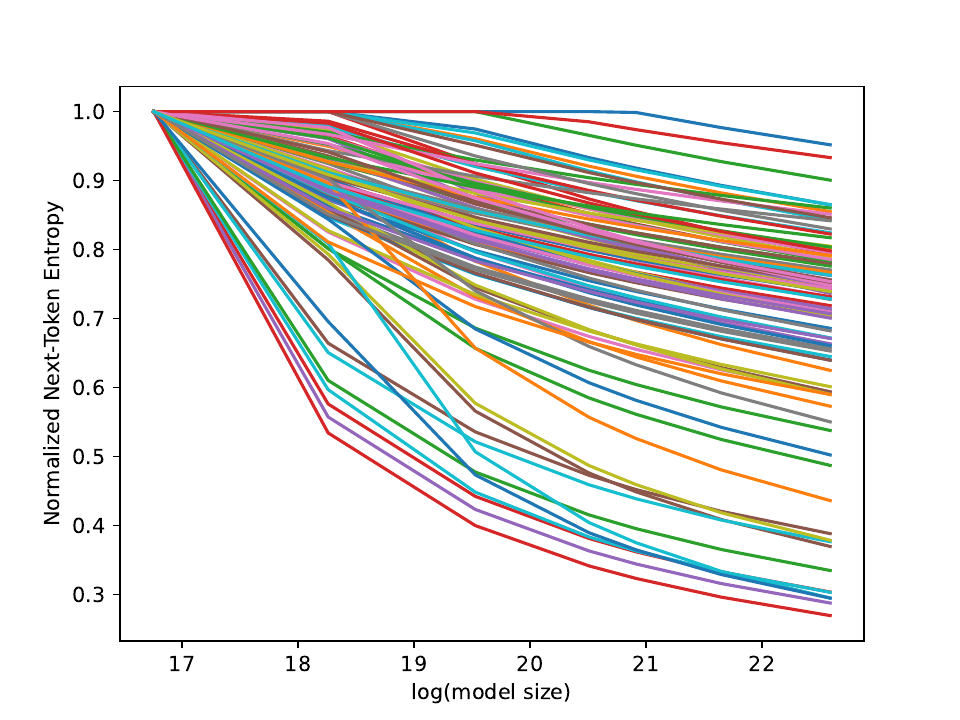}
  \caption{K=1}
  \label{fig:K1_vis}
\end{subfigure}%
\begin{subfigure}{.33\textwidth}
  \centering
  \includegraphics[width=1\linewidth]{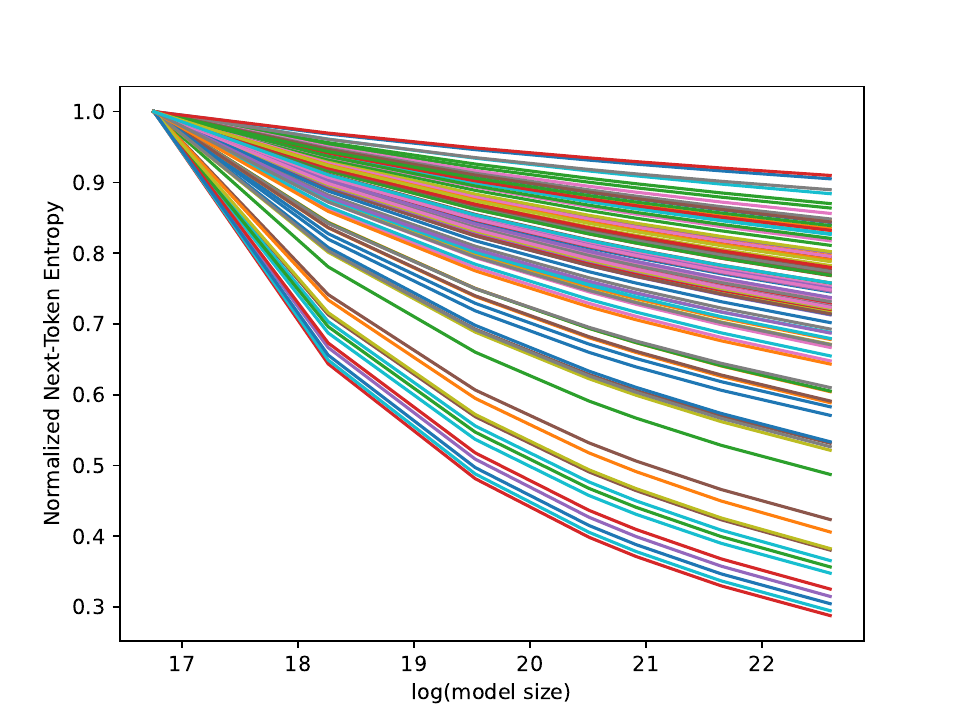}
  \caption{K=2}
  \label{fig:K2_vis}
\end{subfigure}%
\begin{subfigure}{.33\textwidth}
  \centering
  \includegraphics[width=1\linewidth]{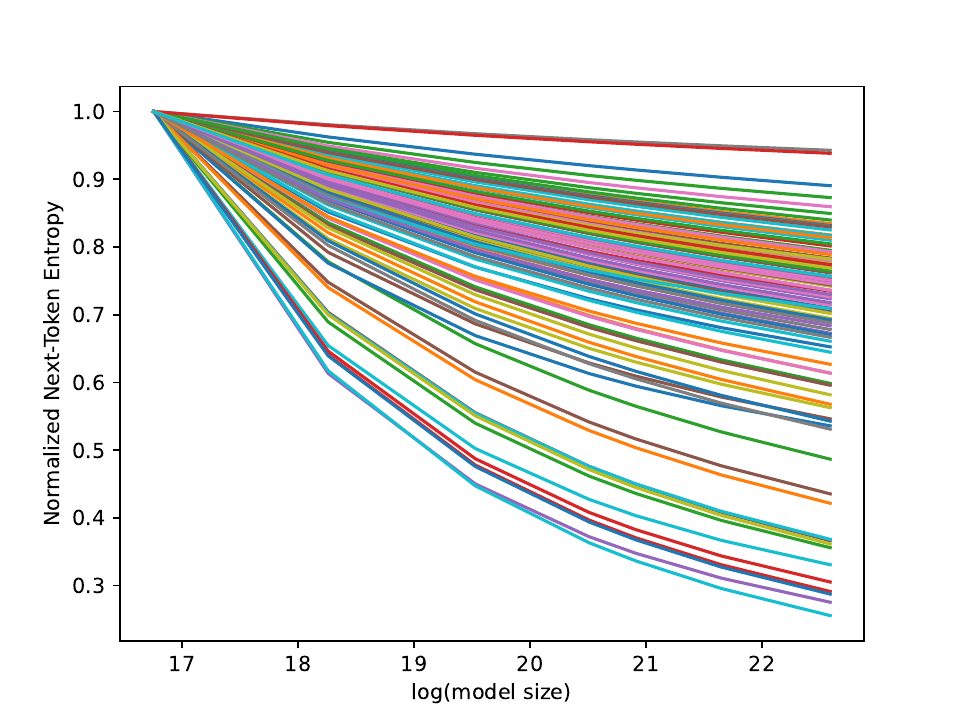}
  \caption{K=10}
  \label{fig:K10_vis}
\end{subfigure}
\caption{Normalized entropy decays predicted by the THF models with different $K$ (i.e., highest degree of the fractional polynomial). We normalize the predicted entropies by the predicted entropy of 70M Pythia (to make all the curves start from 1 given the model size of the smallest LM) and plot the curves for the first 100 tokens in our validation set.}
\label{fig:K_vis}
\end{figure*}

\subsection{Choice of K}
\label{sec:K_analysis}

We set the highest order of fractional polynomial ($K$) as $10$ because we find that the performance is not sensitive to the choice of $K$ once $K$ is larger than a number. In \Cref{fig:K_all_per}, the performance of REAL sampling stays almost the same when $K>=2$, but the performances of $K=1$ drop significantly. 

To explain the observation, we compare the normalized entropy decays of the models with $K=1$, $K=2$, and $K=10$ using the first 100 tokens in our validation set. \Cref{fig:K_vis} shows that the curves from $K=2$ and $K=10$ are very similar while the curves from $K=1$ are different. This indicates that increasing the $K$ to $10$ for the THF model does not have a serious overfitting problem because it could learn to assign $0$ weight to the unnecessary terms in fractional polynomials. Nevertheless, similar to the findings in the last section, we observe that a more complex THF model (i.e., a higher $K$ or a larger model size) seems to perform slightly better given factual prompts in \Cref{fig:K_factual_per} due to its prediction power but performing slightly worse given nonfactual prompts in \Cref{fig:K_nonfactual_per}.

\begin{table}[t!]
\centering
\caption{Comparing LLM's entropy predictions $e_c(s_N)$ from different THF models using regression metrics. FP means fractional polynomials we used in \Cref{eq:frac_poly}.}
\label{tb:reg_metrics}
\begin{tabular}{l|cccc}
         & \textbf{Pearson r} & \textbf{R2}   & \textbf{MSE ($\downarrow$)} & \textbf{Mean L1 ($\downarrow$)} \\ \hline
Exp      & 0.843              & 0.708          & 0.786            & 0.64                 \\
Logistic & 0.842              & 0.707          & 0.788            & 0.641                \\
FP[K=10] (REAL)    & 0.843              & 0.71           & 0.78             & 0.639                \\
FP[K=6]     & 0.843              & 0.71           & 0.781            & 0.639                \\
FP[K=4]      & \textbf{0.844}     & \textbf{0.712} & \textbf{0.776}   & \textbf{0.636}       \\
FP[K=3]      & 0.843              & 0.709          & 0.782            & 0.64                 \\
FP[K=2]      & \textbf{0.844}     & 0.711          & 0.778            & 0.638                \\
FP[K=1]      & \textbf{0.844}     & 0.711          & 0.777            & 0.641                \\         
\end{tabular}
\end{table}

\begin{figure*}[t!]
\centering
\begin{subfigure}{.5\textwidth}
  \centering
  \includegraphics[width=1\linewidth]{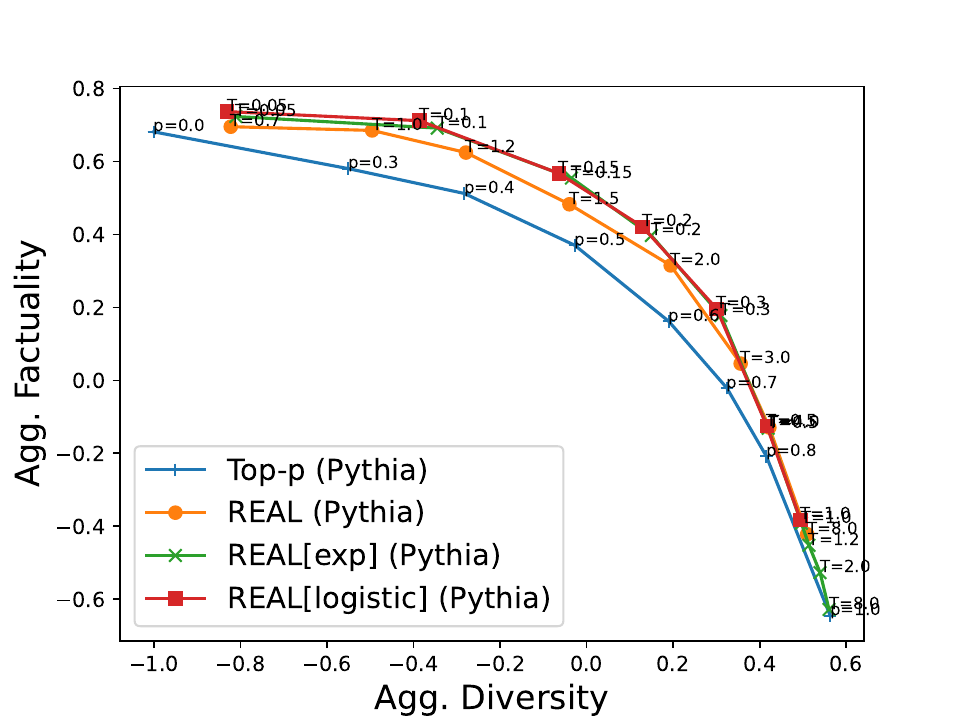}
  %\caption{REAL with Different Parameterizations}
  \label{fig:function_abaltion}
\end{subfigure}%
\begin{subfigure}{.5\textwidth}
  \centering
  \includegraphics[width=1\linewidth]{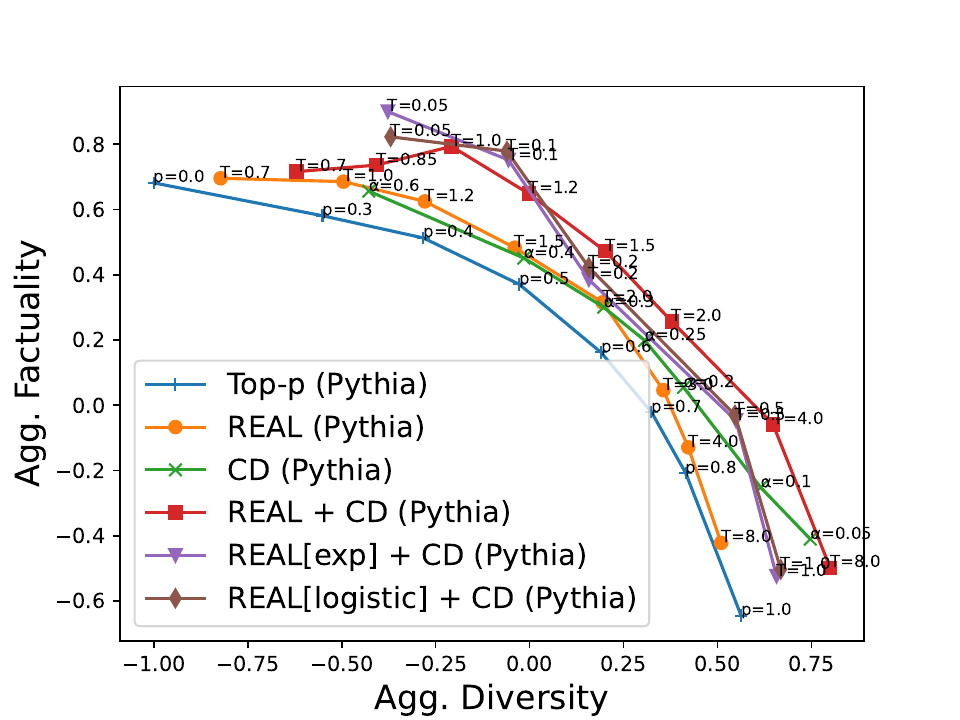}
  %\caption{REAL + CD with Different Parameterizations}
  \label{fig:fuction_abaltion_CD}
\end{subfigure}
% \begin{subfigure}{.5\textwidth}
%   \centering
%   \includegraphics[width=1\linewidth]{figs/sum_plot__both_Ablation2.pdf}
%   \caption{Comparison of Pythia generation performance in FactualPrompt benchmark given different K (highest degrees of fractional polynomial). REAL means REAL (K=10).}
%   \label{fig:K_abaltion}
% \end{subfigure}
\caption{Comparison of Pythia generation performance in \textsc{FactualPrompt} using different functions to model the entropy decay. REAL uses fractional polynomials in \Cref{eq:frac_poly}; REAL[exp] uses exponential decay functions $e_c(s) = z_c + b_c \exp(-\max(0, q_c (s-g_c) ) )$; REAL[logistic] uses logistic decay function ($e_c(s) = z_c + \frac{b_c}{1+\exp(- \max(0, q_c (s-g_c) ) )} $).}
\label{fig:decay_function}
\end{figure*}

\subsection{Choice of the Entropy Decay Functions}
\label{sec:fun_analysis}

In our experiments, we use fractional polynomials (FPs) to parameterize the entropy decay function in \Cref{eq:frac_poly}. When practitioners deploy the REAL sampling, more non-increasing functions could be tried. We recommend using FPs by default because of its flexibility. Given a new LLM family, we are not sure how fast their entropy decays would be, so it is good to let the THF model learn the weight of each term in the FP. To further justify the choice, we also try an exponential (exp) decay function ($e_c(s) = z_c + b_c \exp( - \max(0, q_c (s - g_c) ) )$) and a logistic decay function ($e_c(s) = z_c + \frac{b_c} {1+\exp( \max(0, q_c (s - g_c) ) )}$). 

In \Cref{fig:decay_function}, all the parameterization functions provide significant improvement for Pythia 6.9B LLM and the FP is slightly worse than the other ways. Nevertheless, after being combined with contrastive decoding (CD), the FP performs better for high temperature/diversity while other functions perform better for low temperature/diversity. \Cref{fig:func_cd_factual_entail} indicates that the exponential and logistic functions have a higher repetition rate after being combined with CD, especially when the temperature/diversity is high. This drawback makes FP's improvement more consistent across all settings.

%\Cref{fig:comp_gen_entail}
%We choose to use the FP
%exp and logistic seem to have some repetition problems

\subsection{Regression Metrics}
\label{sec:regression}

The goal of our THF model is to predict the difference between the entropy of generation LLM ($e_c^{\theta_{s_N}}$) and entropy of the infinitely-large LLM ($e_c^{AE}$). Since we cannot access infinitely-large LLM, we validate our prediction of $e_c^{\theta_{s_N}}$ (i.e., $e_c(s_N)$). %using \Cref{tb:reg_metrics}.

We compare different parameterization functions, including exponential (exp), logistic, and fractional polynomial function, and also different $K$ values using Pearson correlation coefficient (r), mean squared error (MSE), average L1 norm (Mean L1), and coefficient of determination (R2)\citep{draper1998applied}.

\Cref{tb:reg_metrics} indicates that all methods perform similarly well. Even though our THF model only has 70M parameters, it can predict the entropy of 6.9G LLM accurately by achieving around $0.84$ Pearson correlation coefficient (r). The fractional polynomials are slightly better than other parameterizations, including exponential and logistic functions and the prediction performance is not sensitive to the maximal degree of our fractional polynomial ($K$).

%fractional polynomials are not the only way to parameterize the entropy decay function.

\section{More Results for \textsc{FactualityPrompts}}
\label{sec:more_results}

%\todo{add some explanation}
%\subsection{Comparison with factual sampling}
%\label{sec:factual_sampling}

%\citet{lee2022factuality} assumes that the hallucination usually happened near the end of a sentence, so it decays the $p$ threshold exponentially of each sentence. 
%\citet{lee2022factuality} require some prior knowledge. The assumption might not be true in some datasets or applications.
%\citet{lee2022factuality} is a heuristic designed for this Wikipedia continuation

%visualize the estimated residual entropy in \Cref{sec:visualization}, 

In this section, we evaluate longer continuations in \Cref{sec:longer_responses},
combine REAL sampling with top-$k$ and factual (F) sampling and test REAL sampling on OpenLLaMA-7b~\citep{openlm2023openllama} in \Cref{sec:more_baselines}, 
analyze the scores of every metric provided by \textsc{FactualityPrompts} in \Cref{sec:org_metrics}, 
report the speed of our implementation in \Cref{sec:speed}, and
report additional statistics in our human experiment for content writing in \Cref{sec:human_stats}. 
%and analyze the possible reasons that \textbf{REAL w/o AE} is much better than \textbf{exp(-e/T)} in \Cref{sec:ent_baseline}.

%and report our human experiment for creative writing in \Cref{sec:creative}.

%The incoherency might increases the interestingness.

%The result indicates that \textbf{REAL} might 
%In story generation, the LLM would output an incoherent st

%we do not include enough short stories in  for the THF model.
%When we train 

%Although we focus on the factuality in this paper, we also test our paper in creative writing. 

%Two possible reasons
%Too different from our training data

%When $t^p=0$, top-$p$ sampling becomes greedy decoding. 

%all our results in the main paper. 

\begin{figure*}[t!]
\centering
\begin{subfigure}{.32\textwidth}
  \centering
  \includegraphics[width=1\linewidth]{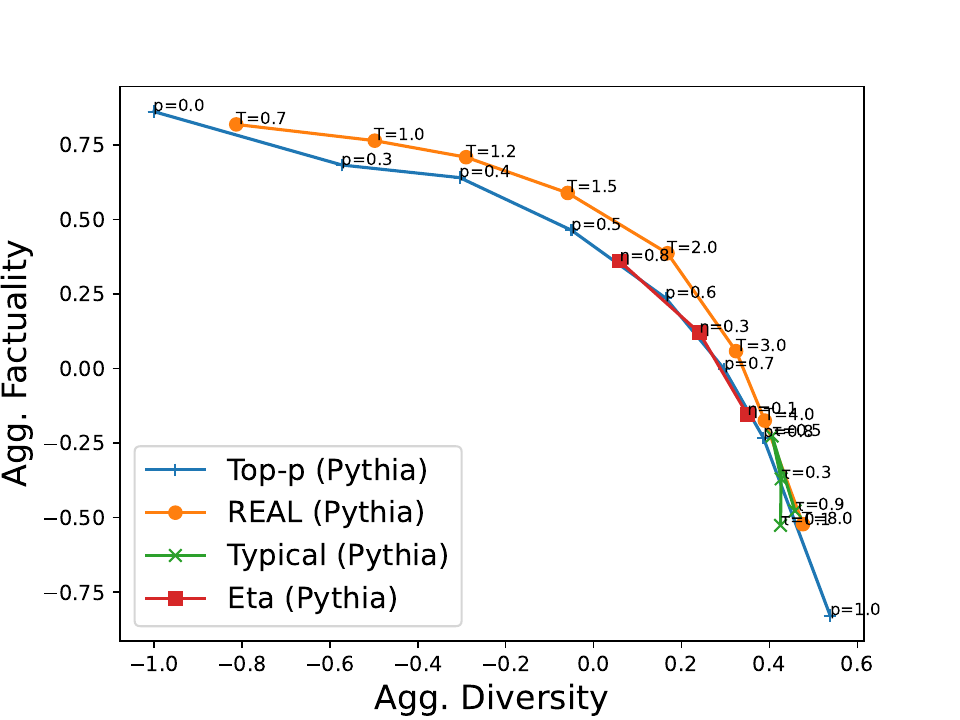}
  \caption{REAL vs Thresholding Methods}
  \label{fig:topp_factual_s3}
\end{subfigure} \;%
\begin{subfigure}{.32\textwidth}
  \centering
  \includegraphics[width=1\linewidth]{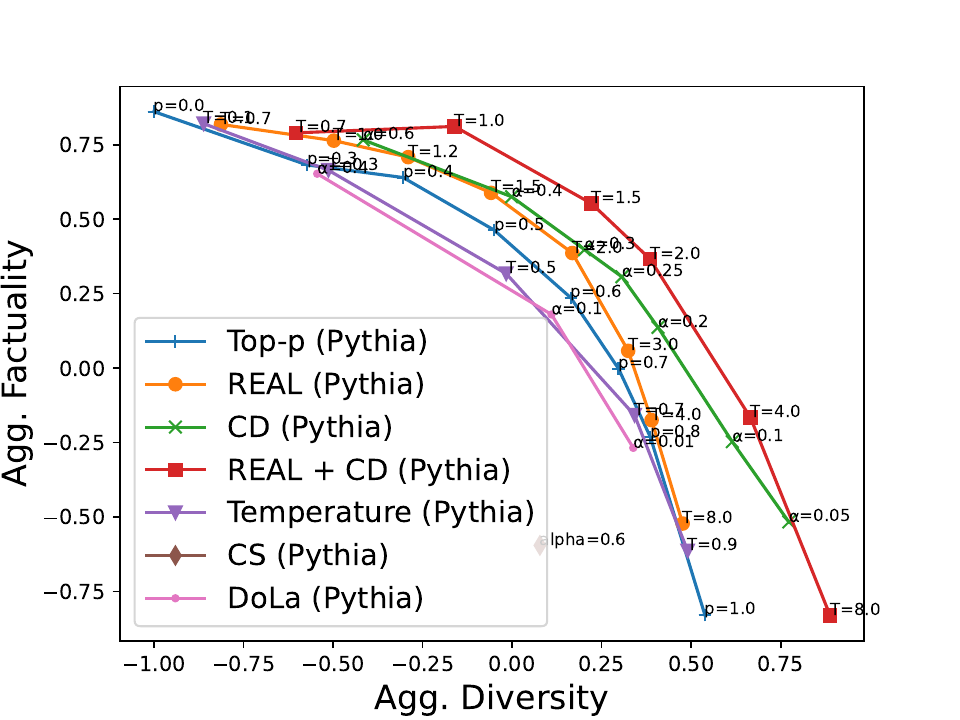}
  \captionsetup{justification=centering}
  \caption{REAL vs Distribution \\ Modifications}
  \label{fig:cd_factual_s3}
\end{subfigure}%
\begin{subfigure}{.32\textwidth}
  \centering
  \includegraphics[width=1\linewidth]{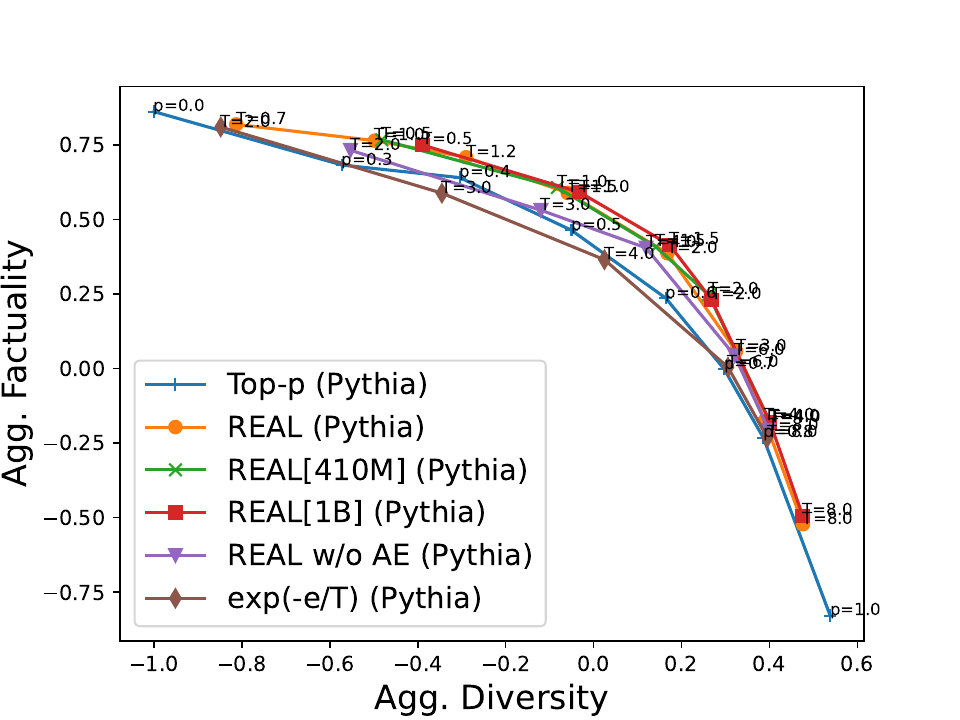}
  \caption{Ablation}
  \label{fig:ablation_factual_s3}
\end{subfigure}
\begin{subfigure}{.45\textwidth}
  \centering
  \includegraphics[width=1\linewidth]{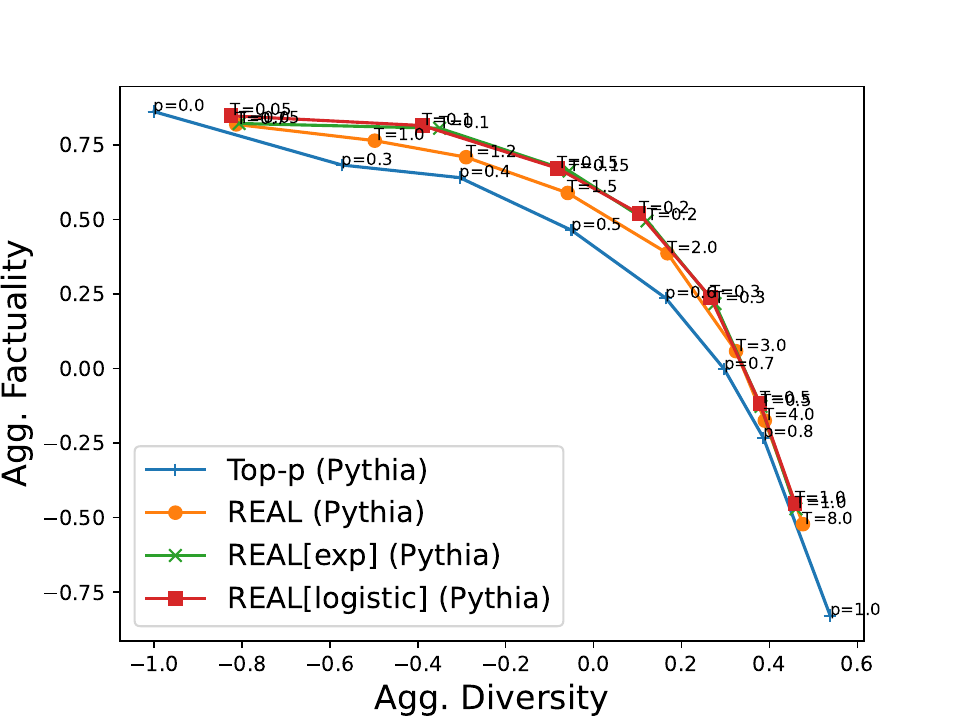}
  \caption{Different Parameterizations}
  \label{fig:ablation3_factual_s3}
\end{subfigure}%
\begin{subfigure}{.45\textwidth}
  \centering
  \includegraphics[width=1\linewidth]{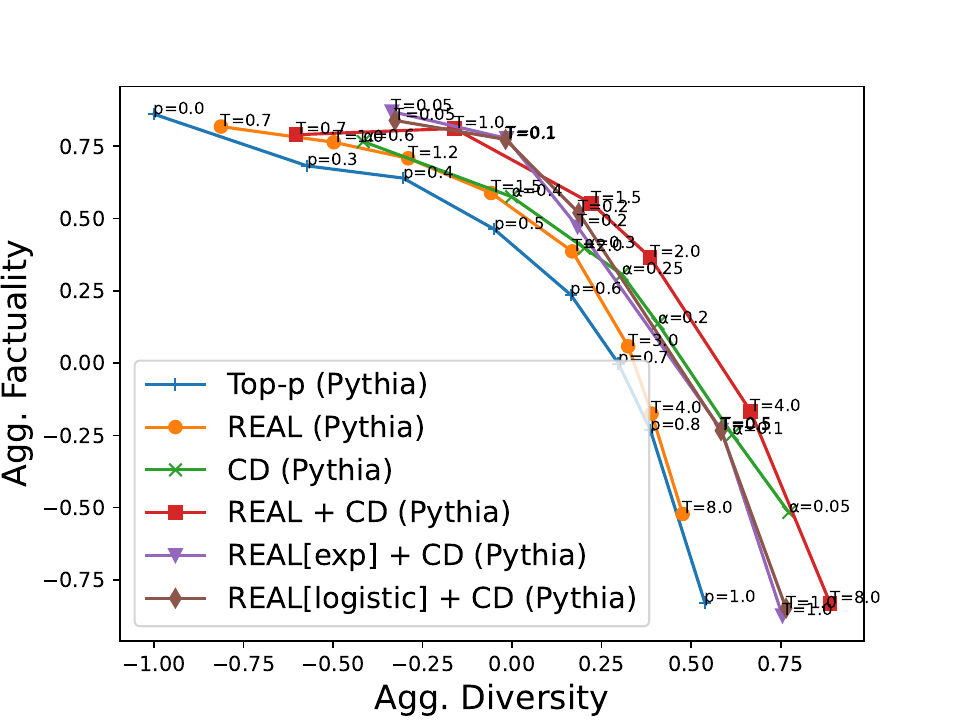}
  \caption{Different Parameterizations}
  \label{fig:cd2_factual_s3}
\end{subfigure}
% \begin{subfigure}{.33\textwidth}
%   \centering
%   \includegraphics[width=1\linewidth]{figs/sum_plot_s3__both_Ablation2.pdf}
%   \caption{Different Polynomial Degrees}
%   \label{fig:ablation2_factual_s3}
% \end{subfigure}
\begin{subfigure}{.45\textwidth}
  \centering
  \includegraphics[width=1\linewidth]{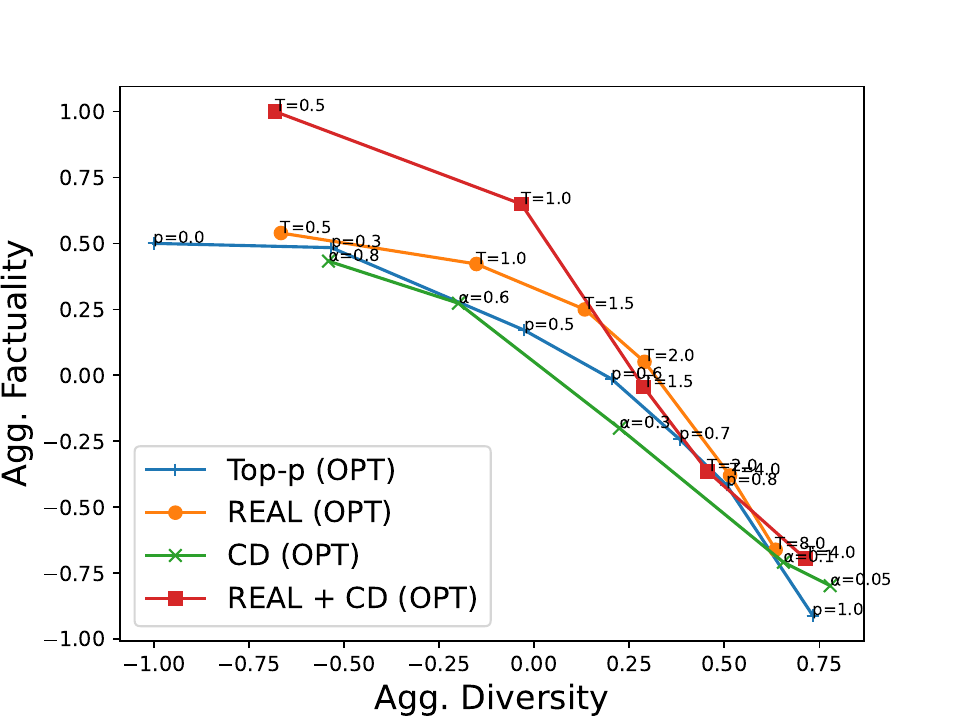}
  \caption{OPT-6.7b}
  \label{fig:opt_factual_s3}
\end{subfigure}%
\begin{subfigure}{.45\textwidth}
  \centering
  \includegraphics[width=1\linewidth]{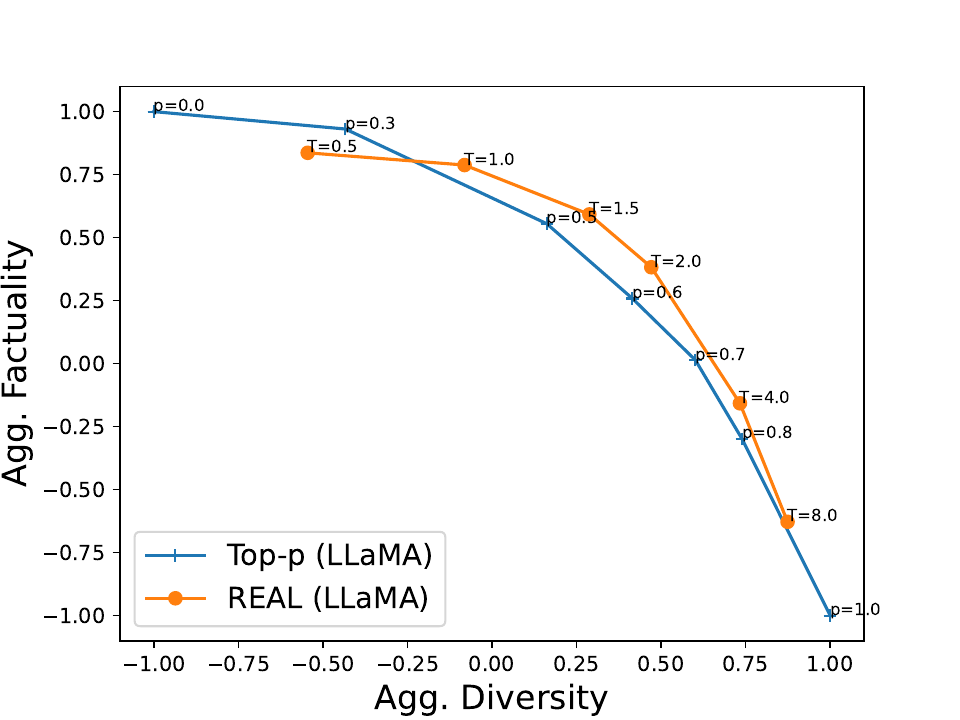}
  \caption{OpenLLaMA-7b}
  \label{fig:llama_factual_s3}
\end{subfigure}
\begin{subfigure}{.45\textwidth}
  \centering
  \includegraphics[width=1\linewidth]{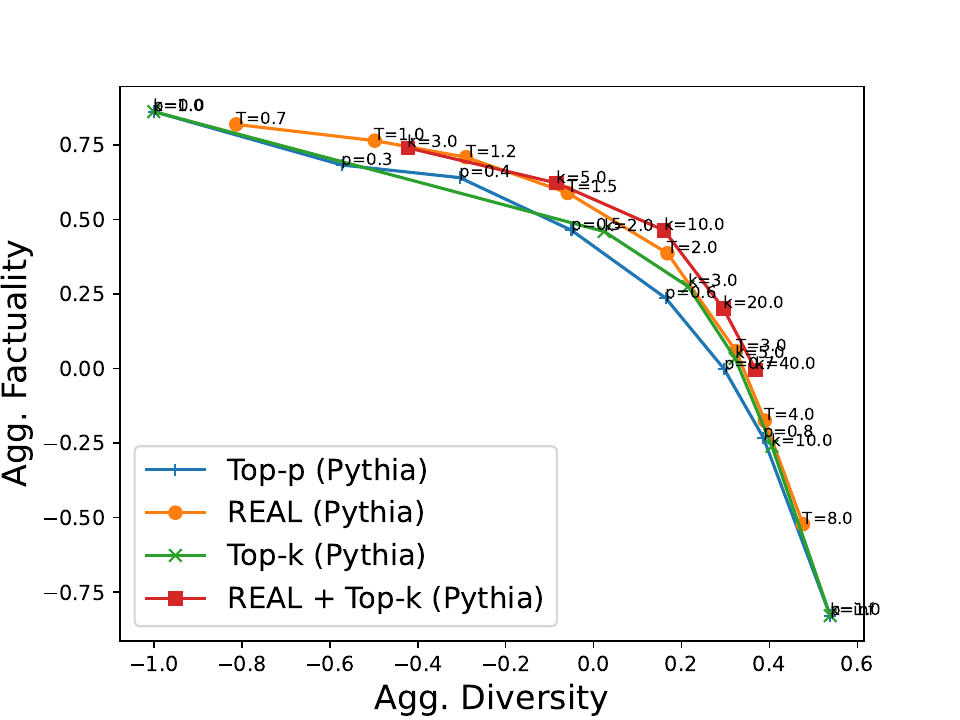}
  \caption{REAL vs Top-$k$ Sampling}
  \label{fig:topk_factual_s3}
\end{subfigure}%
\begin{subfigure}{.45\textwidth}
  \centering
  \includegraphics[width=1\linewidth]{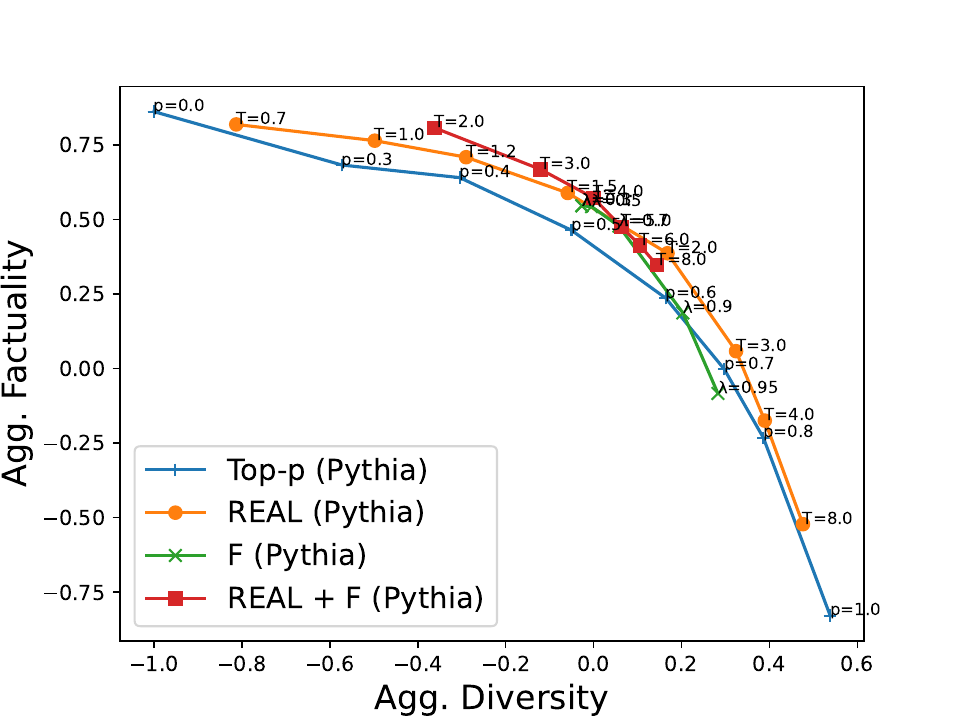}
  \caption{REAL vs Factual (F) Sampling}
  \label{fig:f_factual_s3}
\end{subfigure}
\caption{Open-ended text generation performance comparison using \textsc{FactualityPrompts}. The same experiment setup of \Cref{fig:comp_gen} except that the factuality metrics and dist-2 are applied to 3 generated sentences rather than 1 generated sentence.
 }
\label{fig:comp_gen_s3}
\end{figure*}

\subsection{Evaluation on Longer Continuations}
\label{sec:longer_responses}

\begin{figure*}[t!]
\centering
\begin{subfigure}{.5\textwidth}
  \centering
  \includegraphics[width=1\linewidth]{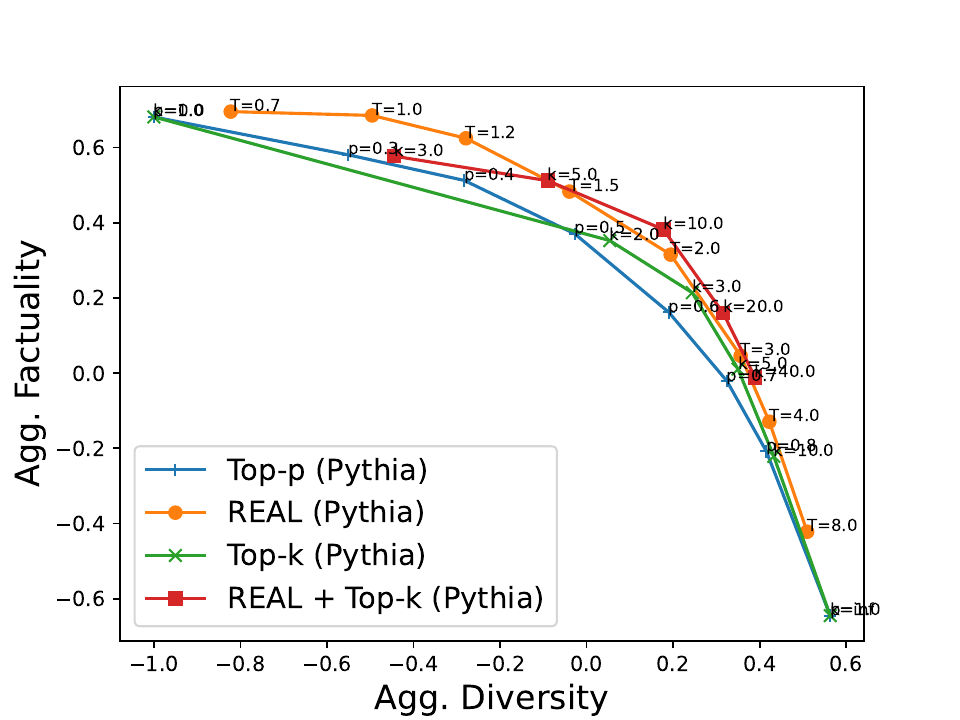}
  \caption{REAL vs Top-$k$ Sampling}
  \label{fig:topk_factual}
\end{subfigure}%
\begin{subfigure}{.5\textwidth}
  \centering
  \includegraphics[width=1\linewidth]{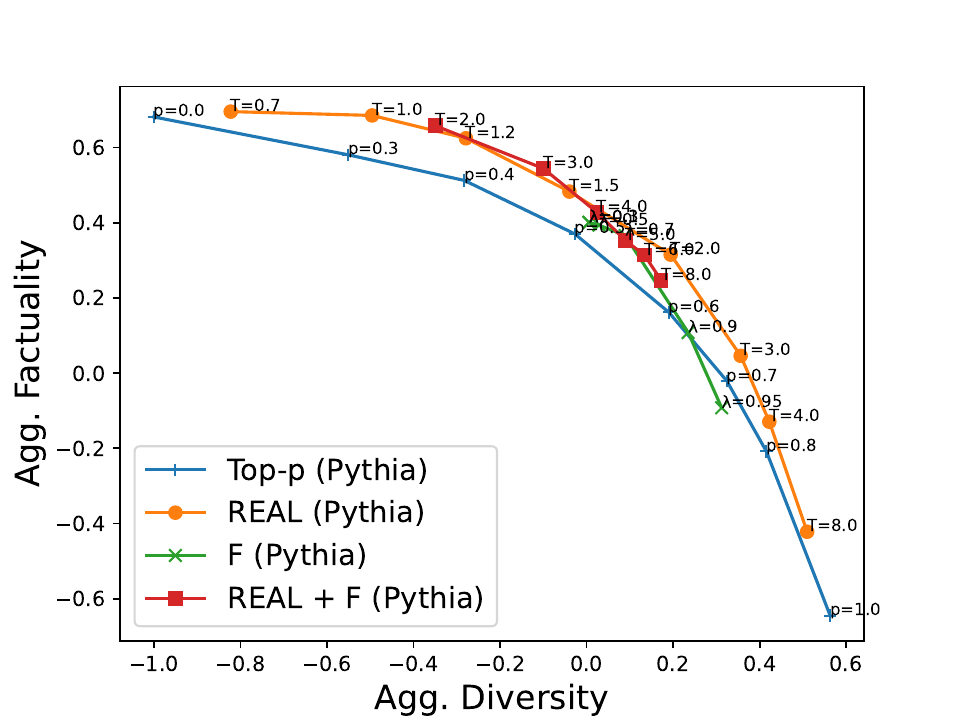}
  \caption{REAL vs Factual (F) Sampling}
  \label{fig:lee_factual}
\end{subfigure}
\begin{subfigure}{.5\textwidth}
  \centering
  \includegraphics[width=1\linewidth]{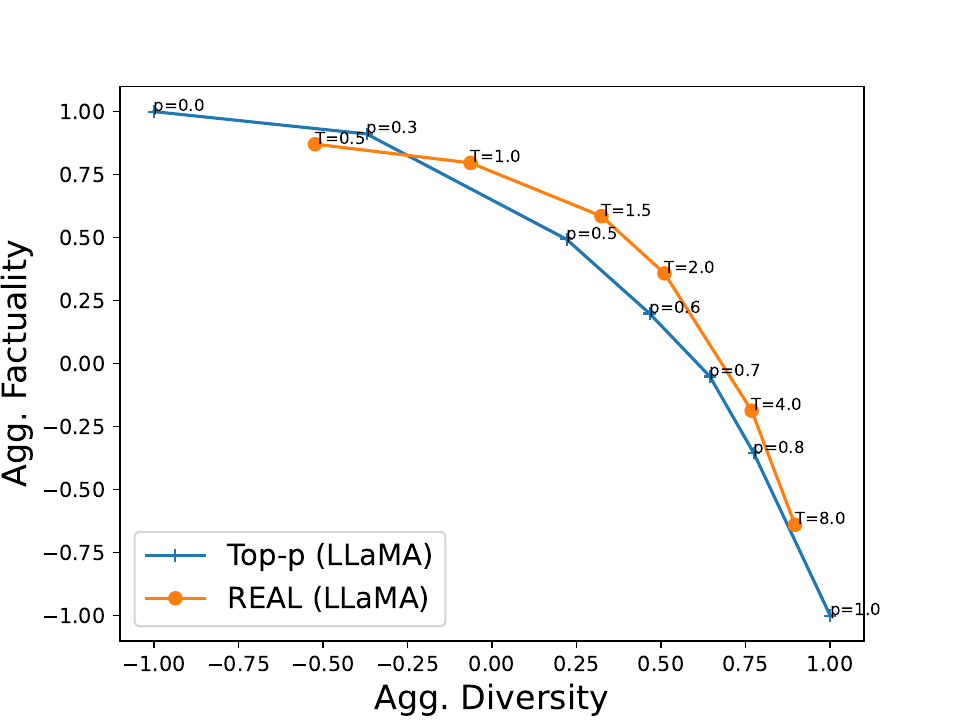}
  \caption{Using OpenLLaMA-7b as the generation LLM and the THF model trained on Pythia.}
  \label{fig:llama_factual}
\end{subfigure}
\caption{Comparison of open-ended text generation methods in \textsc{FactualityPrompts} including top-$k$~\citep{fan2018hierarchical} and factual (F)~\citep{lee2022factuality} sampling. We also conduct another out-of-domain evaluation for REAL sampling using OpenLLaMA-7b.}
\label{fig:comp_gen_more}
\end{figure*}

The evaluation code from \textsc{FactualityPrompts} only checks the factuality of the first generated sentence after the prompt. As the we generate more text, the hallucination problem is more likely to happen~\citep{zhang2023language}. To see if our conclusions still hold for longer continuations, we evaluate the three generated sentences and plot the results in \Cref{fig:comp_gen_s3}. We can see that the overall trend is similar to the first-sentence results in \Cref{fig:comp_gen} and \Cref{fig:comp_gen_more}, which suggests that our improvements are mostly not affected by the generation length.

%, they only evaluate the first sentence . 
%The trend is the same
%The results are more stable

%factual sampling(\textbf{F})~\citep{lee2022factuality}
%at the positions that are further away from the period

\subsection{More Comparisons}
\label{sec:more_baselines}

Due to the page limit and clarity of the figures, we move some comparisons to this section. In addition to top-$k$ sampling~\citep{fan2018hierarchical}, we also test the following generation methods:
\begin{itemize}[leftmargin=.1in,topsep=0pt]
\setlength\itemsep{-0.3em}
    \item \textbf{REAL + Top-$k$}: Using the THF model to dynamically adjust the threshold in Top-$k$ sampling as $t_c^k = t^k \cdot \exp(-\hat{d}_c^{RE})$, where $t^k$ is a constant hyperparameter.
    \item \textbf{F}: Factual sampling (\textbf{F})~\citep{lee2022factuality} exponentially reduces the $p$ value according to the distance to the last period. As suggested in the paper, we set the decay ratio $\lambda=0.9$ and fix the highest and the lowest sampling threshold to be the default values: $0.9$ and $0.3$, respectively. That is, $\hat{t}_c^p = \max(0.3, 0.9 \cdot 0.9^{x-1})$, where $x$ is the distance to the last period.
    \item \textbf{REAL + F}: Combining our methods with factual sampling using $\hat{t}_c^p = \max(0.3, 0.9^{x-1}) \cdot \exp(\frac{-\hat{d}_c^{RE}}{T})$. 
    \item \textbf{* (LLaMA)}: In the methods, we replace the Pythia 6.9B with OpenLLaMA2-7b~\citep{touvron2023llama,together2023redpajama,openlm2023openllama} as the generation LLM, respectively. Notice that the THF model is still trained using the Pythia family. 
\end{itemize}

\expsec{Results:} 
%\textbf{top-$k$} sampling uses a truncation strategy that is very different from \textbf{top-$p$} sampling, and 
\Cref{fig:comp_gen_more} shows that top-$k$ outperforms top-$p$ sampling at the high diversity side while the factual (\textbf{F}) sampling outperforms top-$p$ sampling at the low diversity side. Notice that factual sampling relies on the heuristic/assumption that hallucination is more likely to happen near the end of the sentence. The assumption might not work well in some languages or applications such as code generation. % for this dataset while .

REAL sampling could be easily combined with these approaches and boost their performance at \Cref{fig:topk_factual} and \Cref{fig:lee_factual}. In \Cref{fig:topk_factual_s3} and \Cref{fig:f_factual_s3}, the combinations are often significantly better than using REAL sampling alone.

Finally, similar to OPT 6.7B, our THF model trained on Pythia could still improve  OpenLLaMA2 in \Cref{fig:llama_factual}. Unlike OPT, we do not report the \textbf{CD} performance for OpenLLaMA2 because the smallest model in the family is too large (OpenLLaMA2-3b).
%The improvements for longer responses  are 
%We can see that 
%Our improvement after combining with top-$k$ sampling and factual sampling improves.

\begin{figure*}[t!]
\centering
\begin{adjustbox}{minipage=\linewidth, scale=0.8}
\begin{subfigure}{.33\textwidth}
  \centering
  \includegraphics[width=1\linewidth]{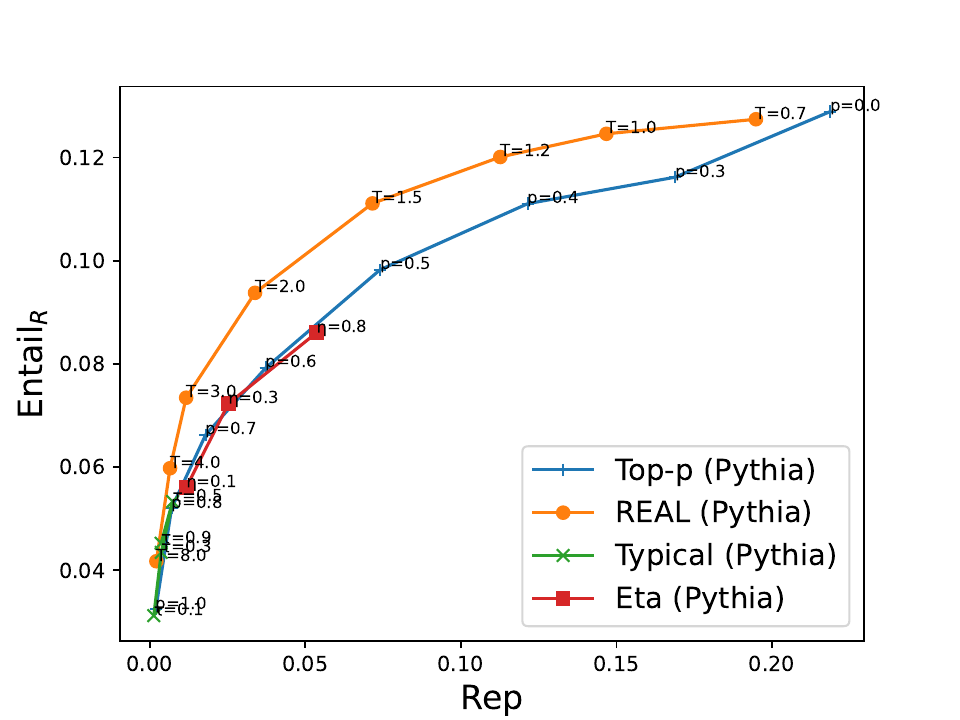}
  \captionsetup{justification=centering}
  \caption{REAL vs Thresholding Methods (Factual)}
  \label{fig:topp_factual_entail}
\end{subfigure}%
\begin{subfigure}{.33\textwidth}
  \centering
  \includegraphics[width=1\linewidth]{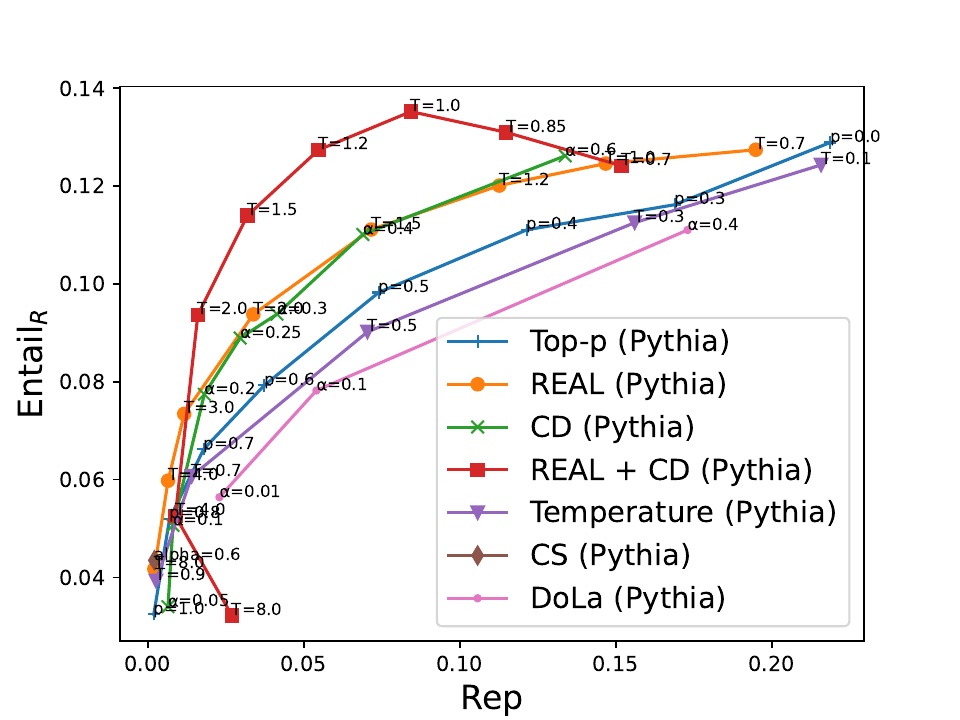}
  \captionsetup{justification=centering}
  \caption{REAL vs Distribution Modifications (Factual)}
  \label{fig:cd_factual_entail}
\end{subfigure}%
\begin{subfigure}{.33\textwidth}
  \centering
  \includegraphics[width=1\linewidth]{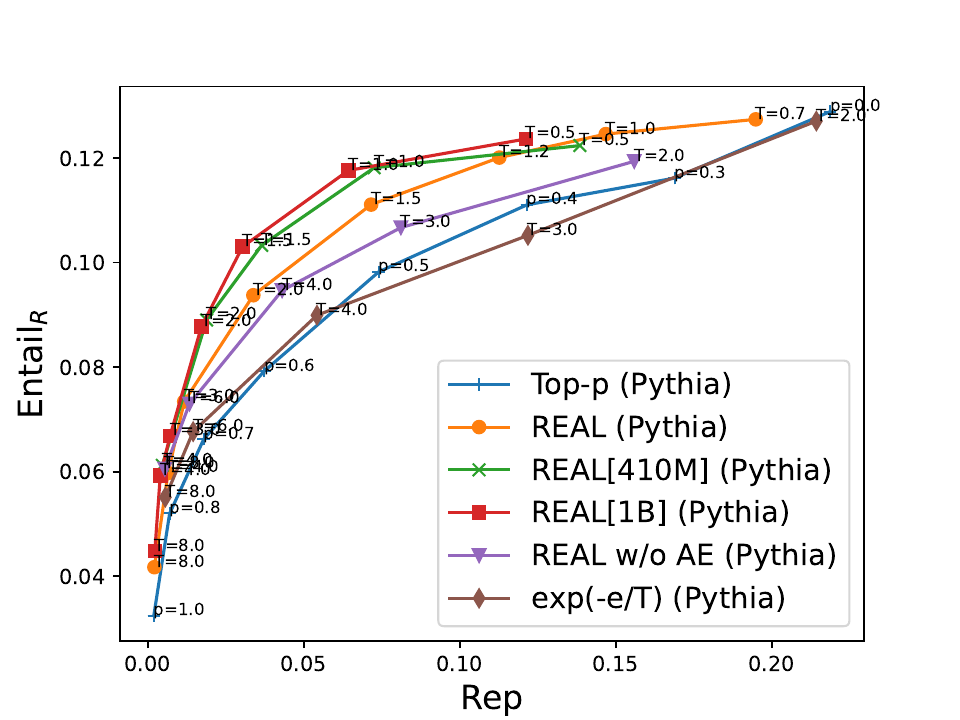}
  \caption{Ablation (Factual)}
  \label{fig:ablation_factual_entail}
\end{subfigure}
\begin{subfigure}{.33\textwidth}
  \centering
  \includegraphics[width=1\linewidth]{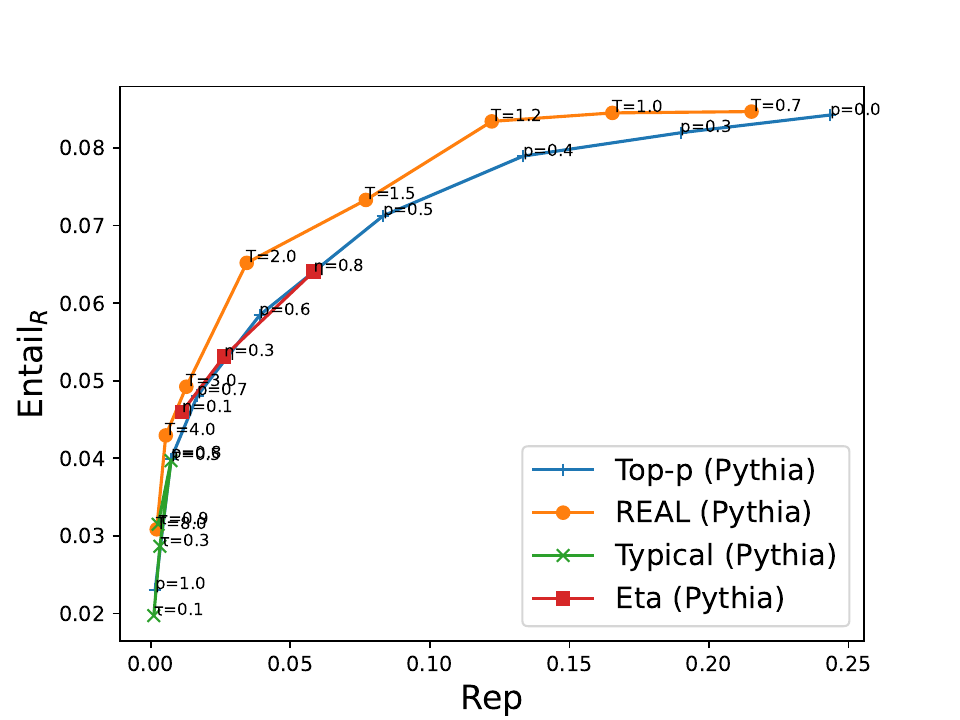}
  \captionsetup{justification=centering}
  \caption{REAL vs Thresholding Methods (Nonfactual)}
  \label{fig:topp_nonfactual_entail}
\end{subfigure}%
\begin{subfigure}{.33\textwidth}
  \centering
  \includegraphics[width=1\linewidth]{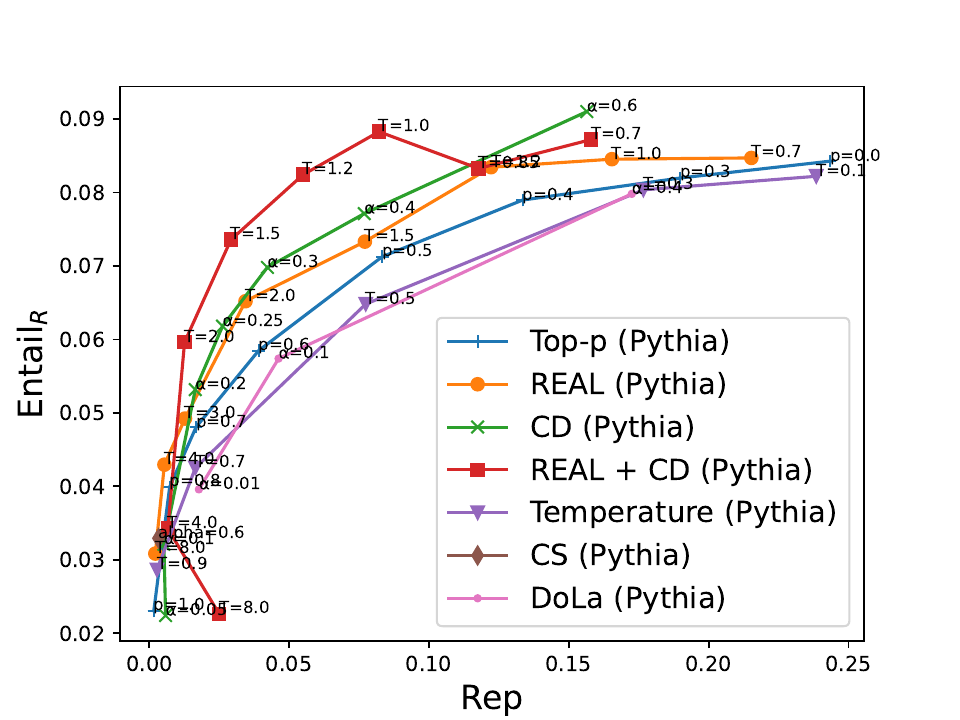}
  \captionsetup{justification=centering}
  \caption{REAL vs Distribution Modifications (Nonfactual)}
  \label{fig:cd_nonfactual_entail}
\end{subfigure}%
\begin{subfigure}{.33\textwidth}
  \centering
  \includegraphics[width=1\linewidth]{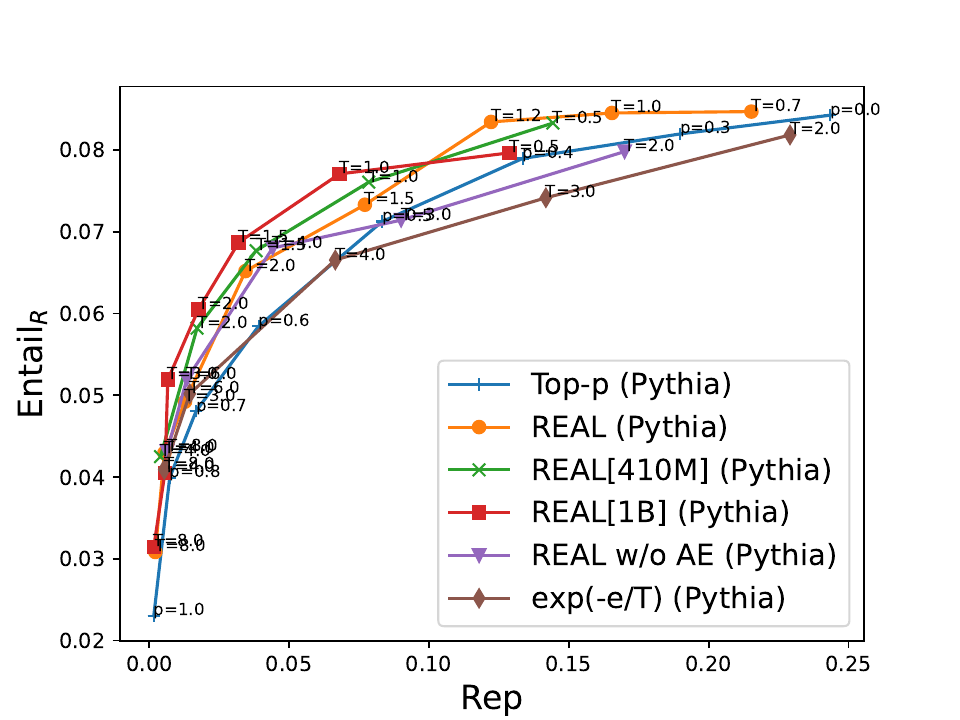}
  \caption{Ablation (Nonfactual)}
  \label{fig:ablation_nonfactual_entail}
\end{subfigure}
\begin{subfigure}{.33\textwidth}
  \centering
  \includegraphics[width=1\linewidth]{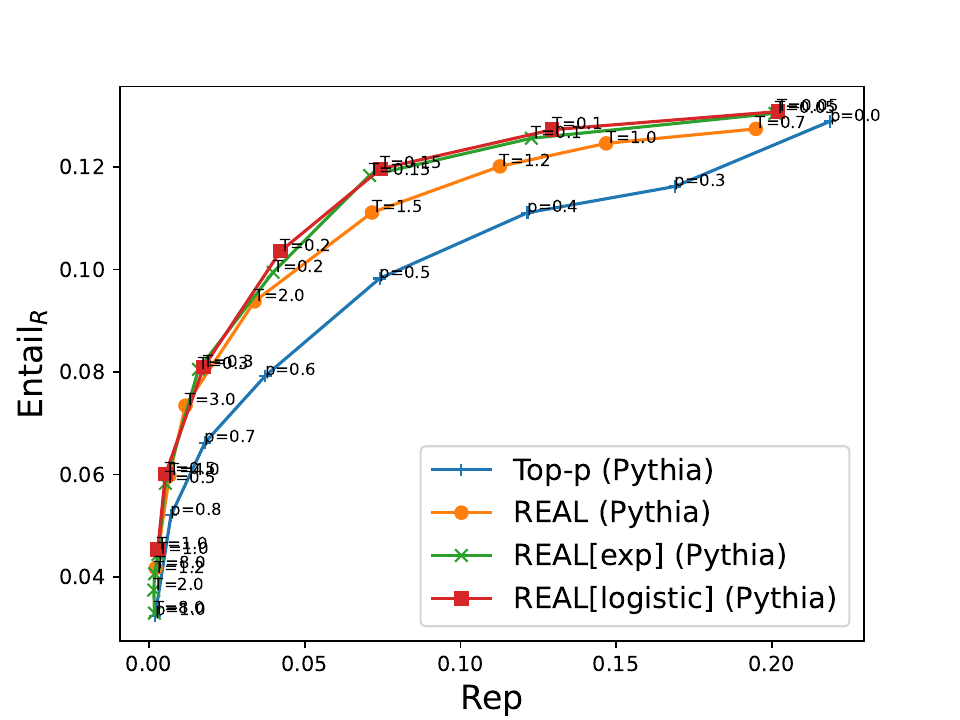}
  \captionsetup{justification=centering}
  \caption{Different Parameterizations \\ (Factual)}
  \label{fig:func_factual_entail}
\end{subfigure}%
\begin{subfigure}{.33\textwidth}
  \centering
  \includegraphics[width=1\linewidth]{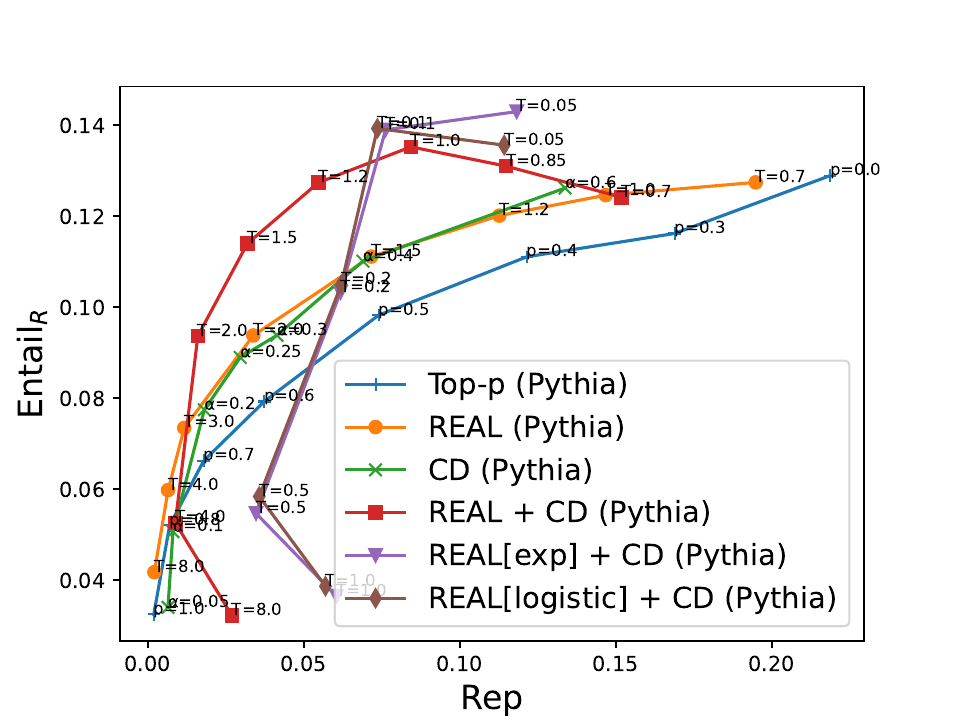}
  \captionsetup{justification=centering}
  \caption{Different Parameterizations \\ (Factual)}
  \label{fig:func_cd_factual_entail}
\end{subfigure}%
\begin{subfigure}{.33\textwidth}
  \centering
  \includegraphics[width=1\linewidth]{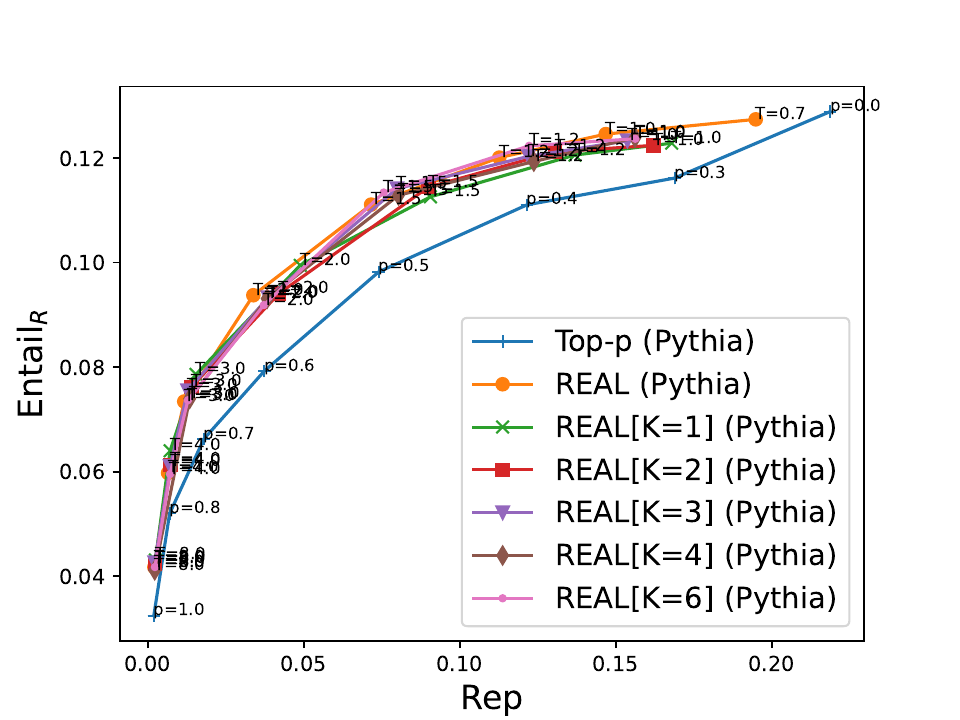}
  \captionsetup{justification=centering}
  \caption{Different Polynomial Degrees \\ (Factual)}
  \label{fig:K_factual_entail}
\end{subfigure}
\begin{subfigure}{.33\textwidth}
  \centering
  \includegraphics[width=1\linewidth]{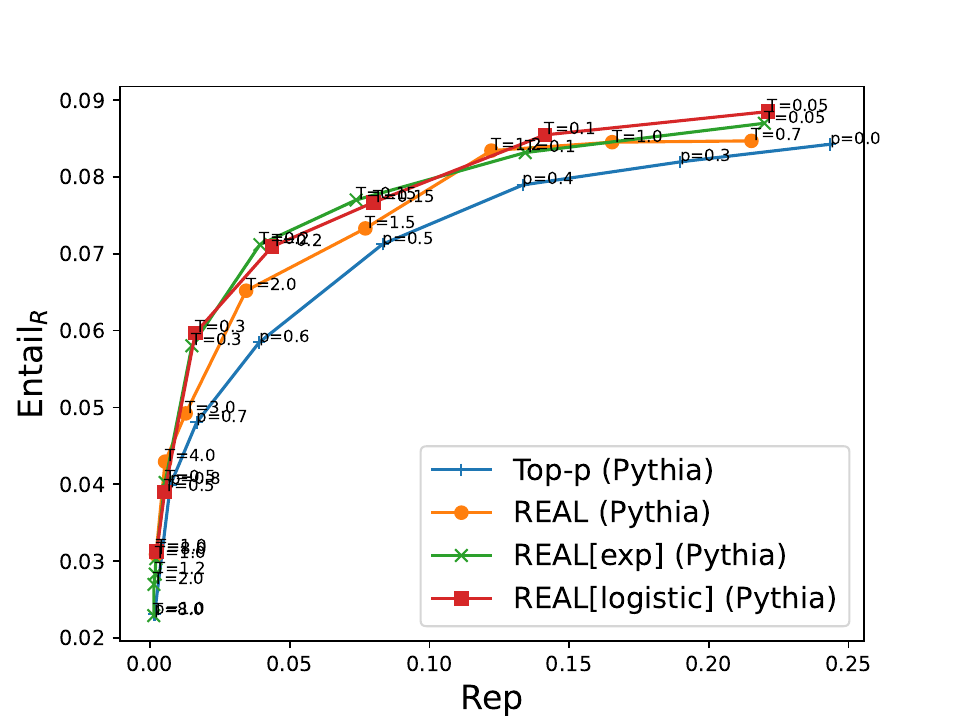}
  \captionsetup{justification=centering}
  \caption{Different Parameterizations \\ (Nonfactual)}
  \label{fig:func_nonfactual_entail}
\end{subfigure}%
\begin{subfigure}{.33\textwidth}
  \centering
  \includegraphics[width=1\linewidth]{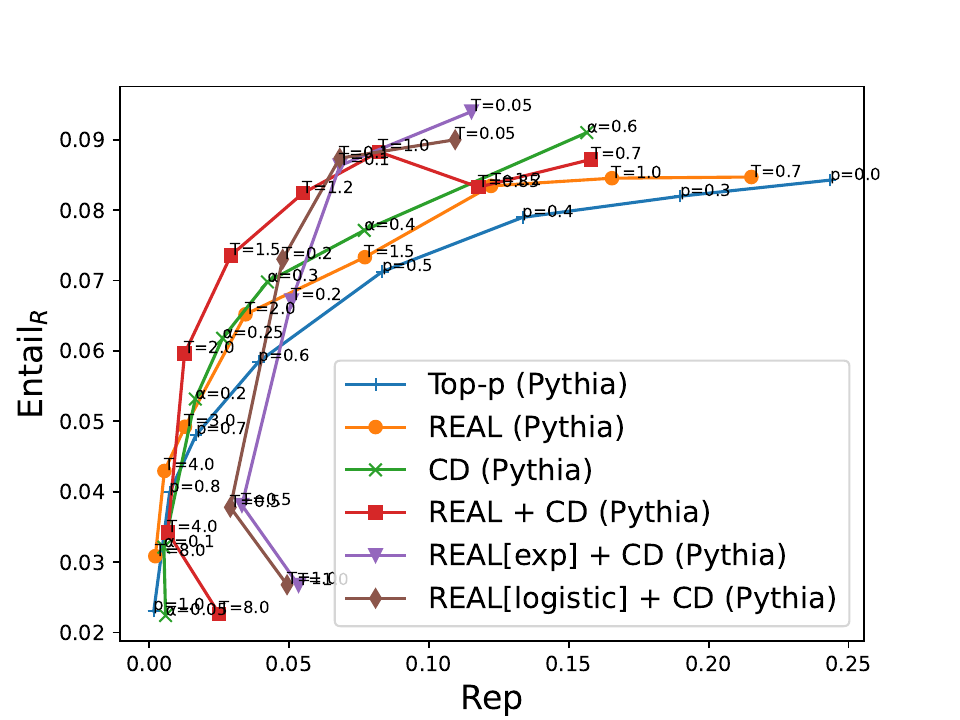}
  \captionsetup{justification=centering}
  \caption{Different Parameterizations \\ (Nonfactual)}
  \label{fig:func_cd_nonfactual_entail}
\end{subfigure}%
\begin{subfigure}{.33\textwidth}
  \centering
  \includegraphics[width=1\linewidth]{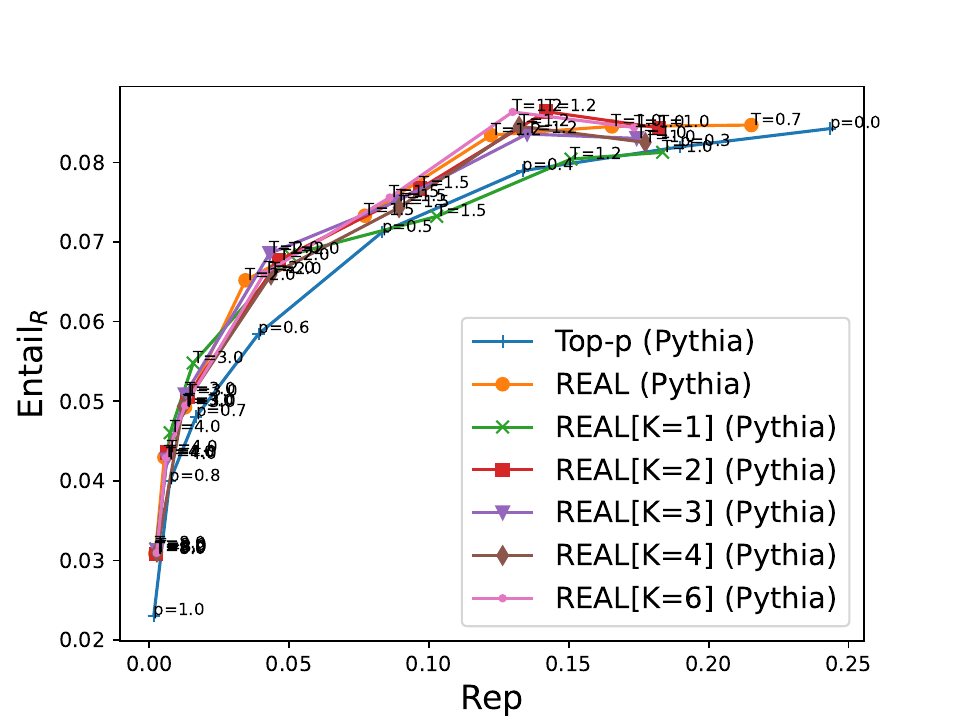}
  \captionsetup{justification=centering}
  \caption{Different Polynomial Degrees \\ (Nonfactual)}
  \label{fig:K_nonfactual_entail}
\end{subfigure}
\begin{subfigure}{.25\textwidth}
  \centering
  \includegraphics[width=1\linewidth]{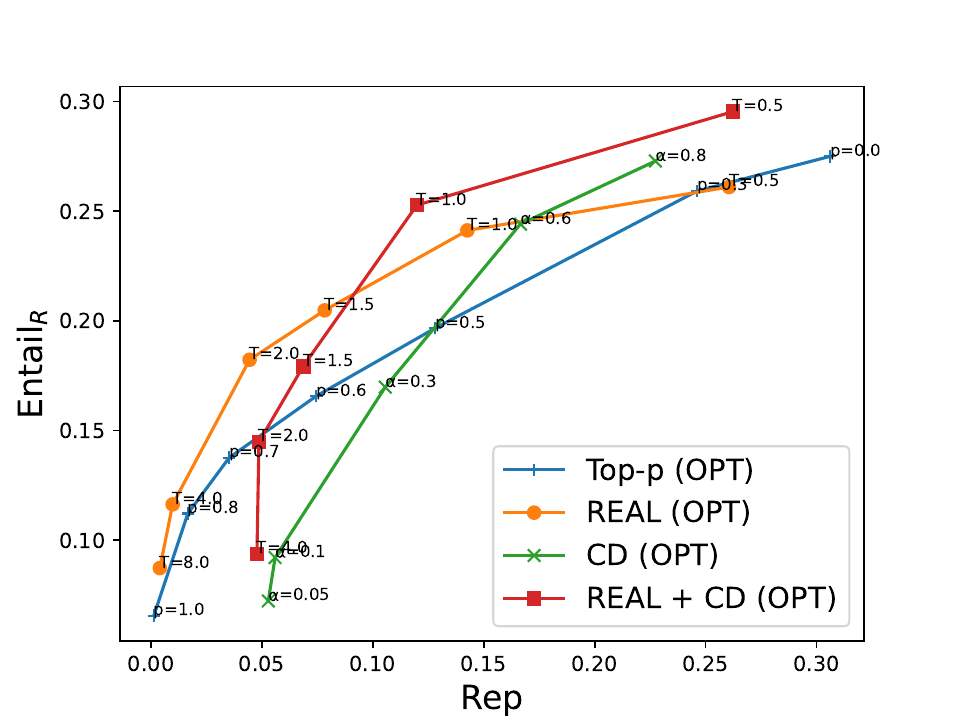}
  \caption{OPT-6.7b (Factual)}
  \label{fig:opt_factual_entail}
\end{subfigure}%
\begin{subfigure}{.25\textwidth}
  \centering
  \includegraphics[width=1\linewidth]{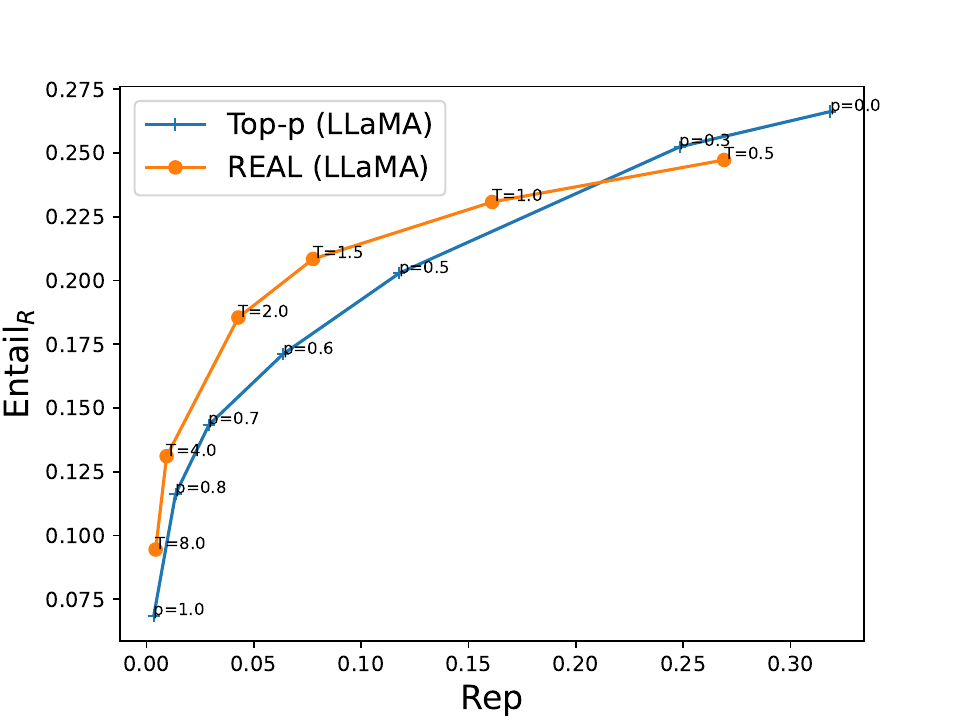}
  \captionsetup{justification=centering}
  \caption{OpenLLaMA-7b \\ (Factual)}
  \label{fig:llama_factual_entail}
\end{subfigure}  %
\begin{subfigure}{.25\textwidth}
  \centering
  \includegraphics[width=1\linewidth]{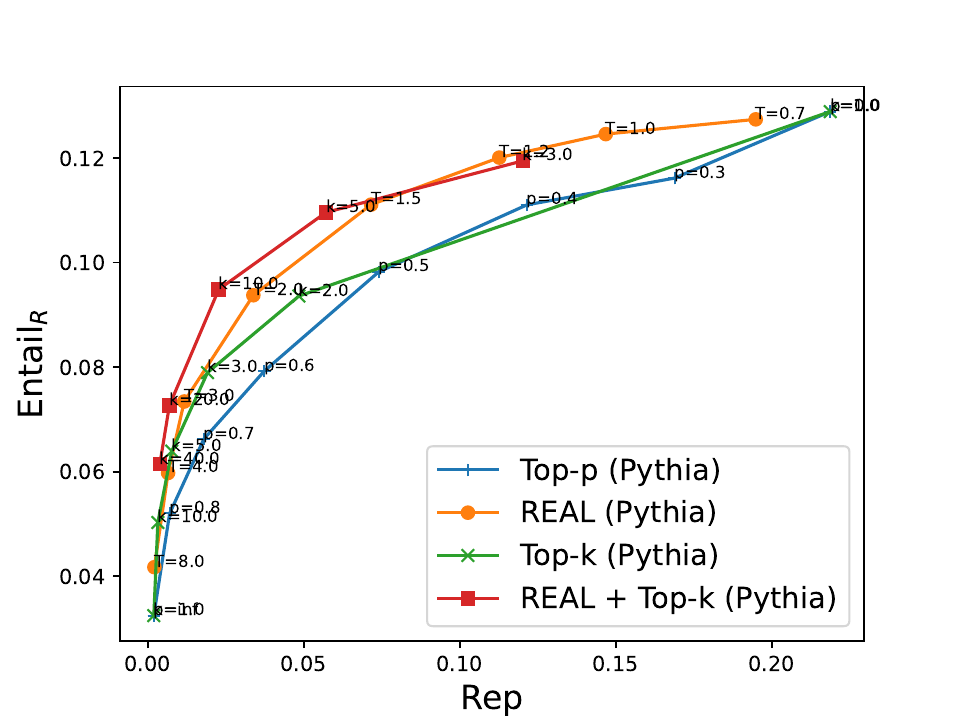}
  \captionsetup{justification=centering}
  \caption{REAL vs Top-k \\ (Factual)}
  \label{fig:topk_factual_entail}
\end{subfigure}%
\begin{subfigure}{.25\textwidth}
  \centering
  \includegraphics[width=1\linewidth]{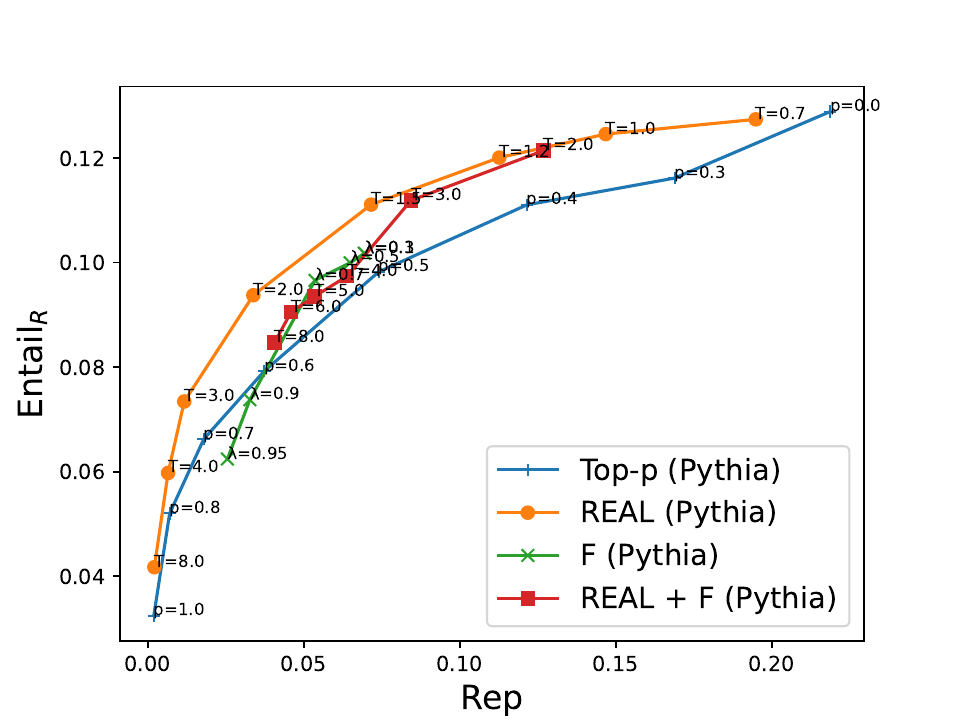}
  \caption{REAL vs F (Factual)}
  \label{fig:lee_factual_entail}
\end{subfigure}
\begin{subfigure}{.25\textwidth}
  \centering
  \includegraphics[width=1\linewidth]{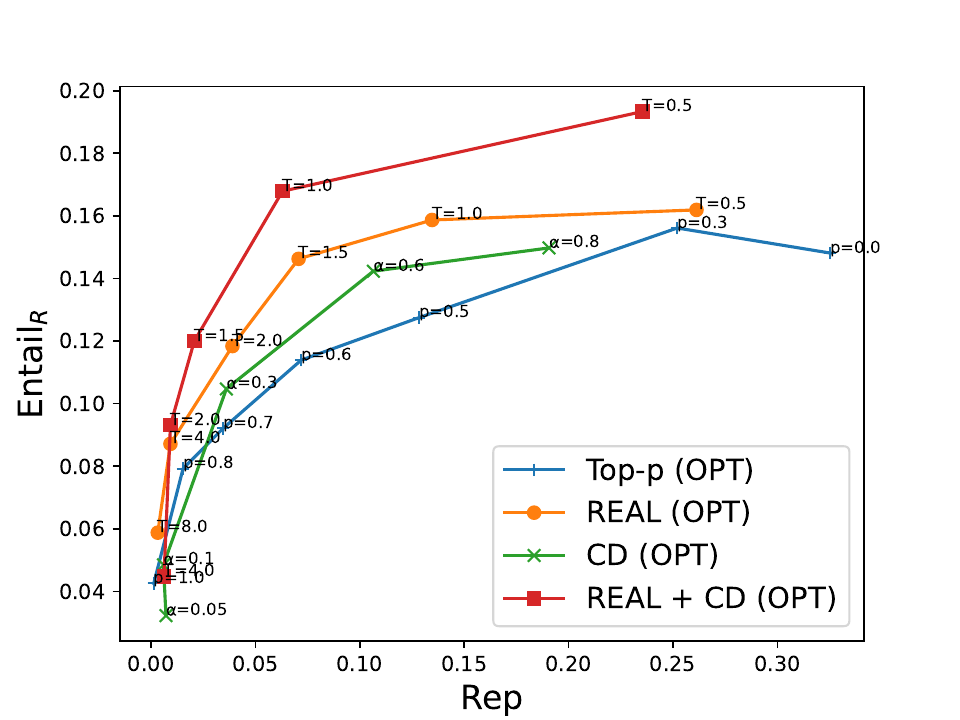}
  \caption{OPT-6.7b (Nonfactual)}
  \label{fig:opt_nonfactual_entail}
\end{subfigure}%
\begin{subfigure}{.25\textwidth}
  \centering
  \includegraphics[width=1\linewidth]{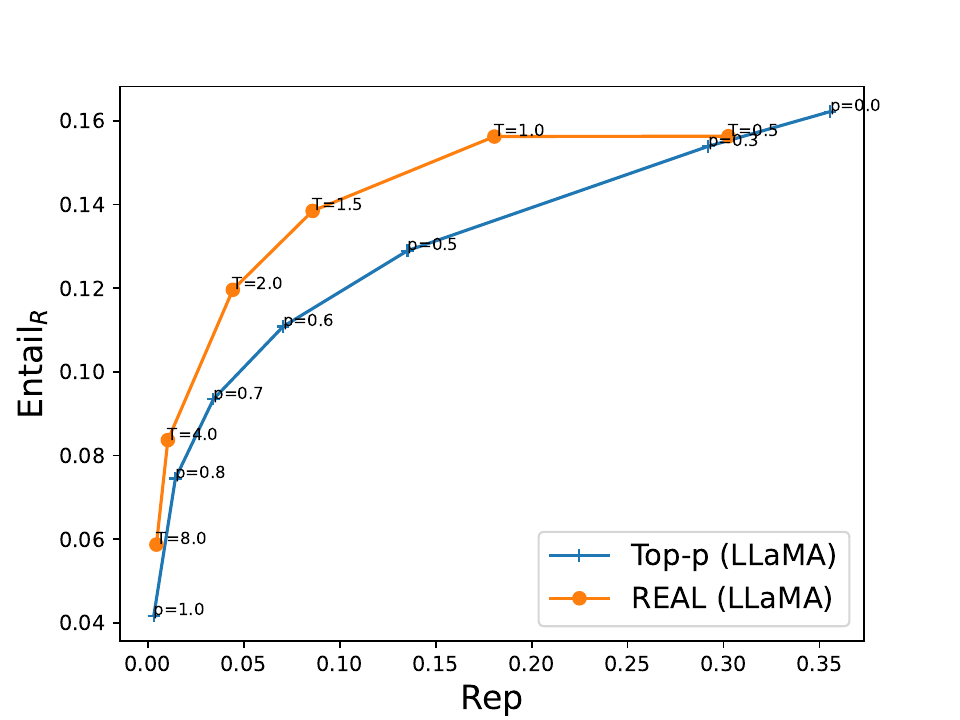}
  \captionsetup{justification=centering}
  \caption{OpenLLaMA-7b \\ (Nonfactual)}
  \label{fig:llama_nonfactual_entail}
\end{subfigure} %
\begin{subfigure}{.25\textwidth}
  \centering
  \includegraphics[width=1\linewidth]{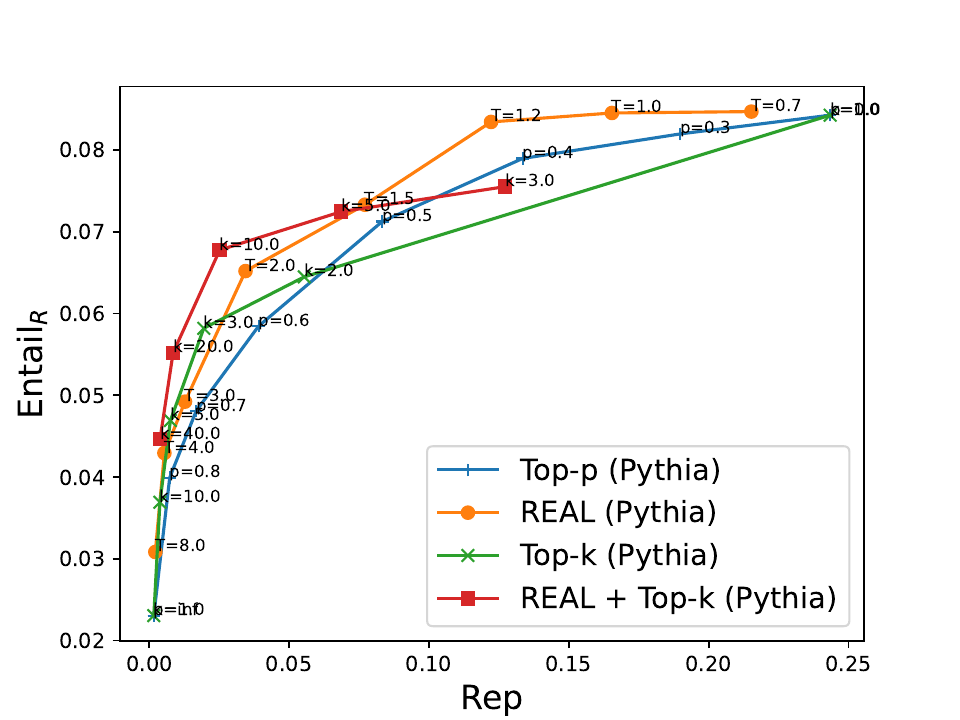}
  \captionsetup{justification=centering}
  \caption{REAL vs Top-k \\ (Nonfactual)}
  \label{fig:topk_nonfactual_entail}
\end{subfigure}%
\begin{subfigure}{.25\textwidth}
  \centering
  \includegraphics[width=1\linewidth]{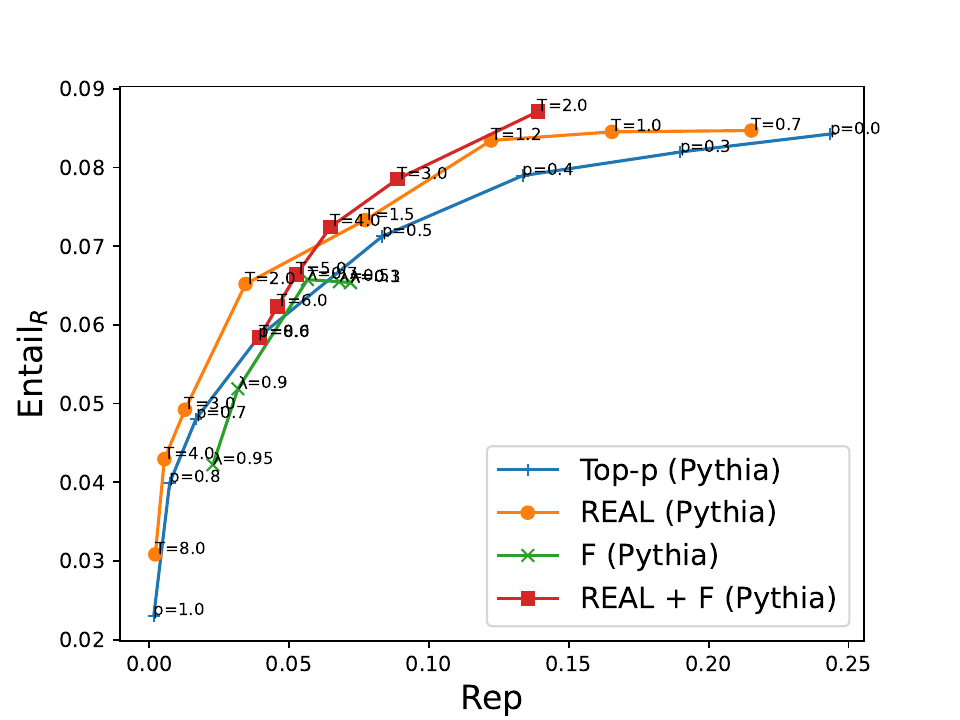}
  \caption{REAL vs F (Nonfactual)}
  \label{fig:lee_nonfactual_entail}
\end{subfigure}
\end{adjustbox}
\caption{The entailment ratio (Entail$_R$) versus repetition ratio (Rep). A lower repetition ratio is better, so the better methods are closer to the top-left corner. (Factual) in the captions means the prompt sentence is factual. The y-axis standard errors of every curve in this figure are 0.0015 on average and smaller than 0.005. } 
\label{fig:comp_gen_entail}
\end{figure*}

\begin{figure*}[t!]
\centering
\begin{adjustbox}{minipage=\linewidth, scale=0.8}
\begin{subfigure}{.33\textwidth}
  \centering
  \includegraphics[width=1\linewidth]{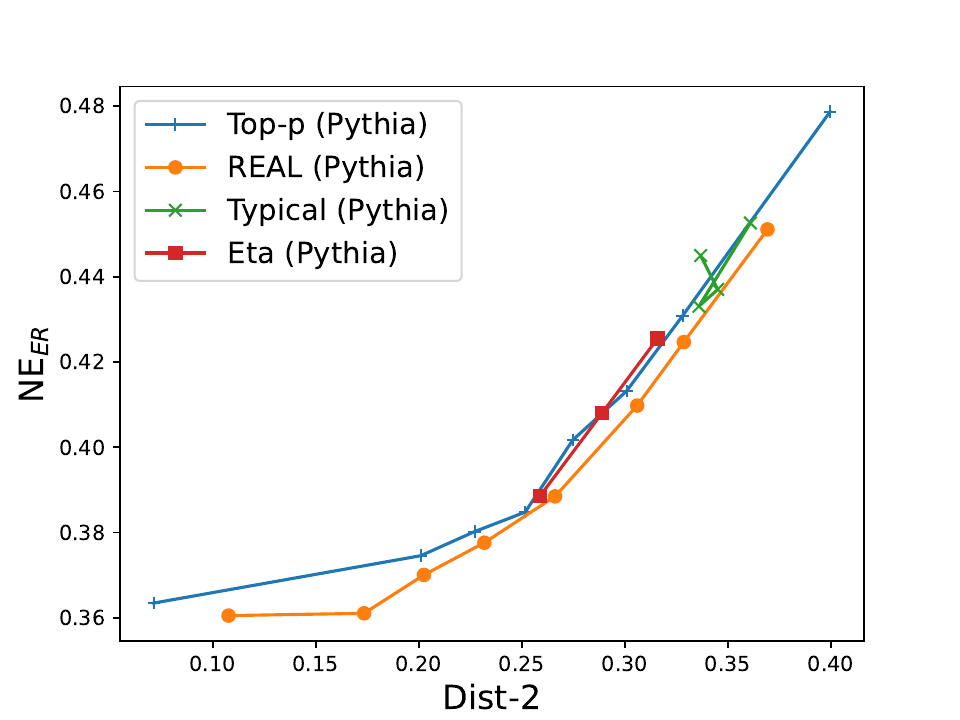}
  \captionsetup{justification=centering}
  \caption{REAL vs Thresholding Methods (Factual)}
  \label{fig:topp_factual_ne}
\end{subfigure}%
\begin{subfigure}{.33\textwidth}
  \centering
  \includegraphics[width=1\linewidth]{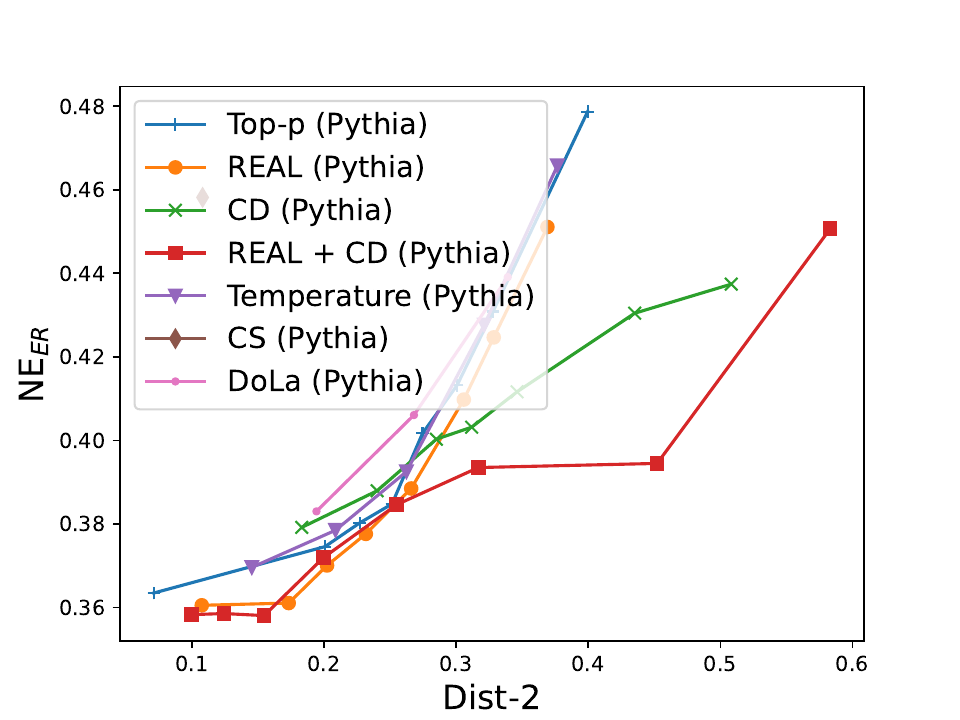}
  \captionsetup{justification=centering}
  \caption{REAL vs Distribution Modifications (Factual)}
  \label{fig:cd_factual_ne}
\end{subfigure}%
\begin{subfigure}{.33\textwidth}
  \centering
  \includegraphics[width=1\linewidth]{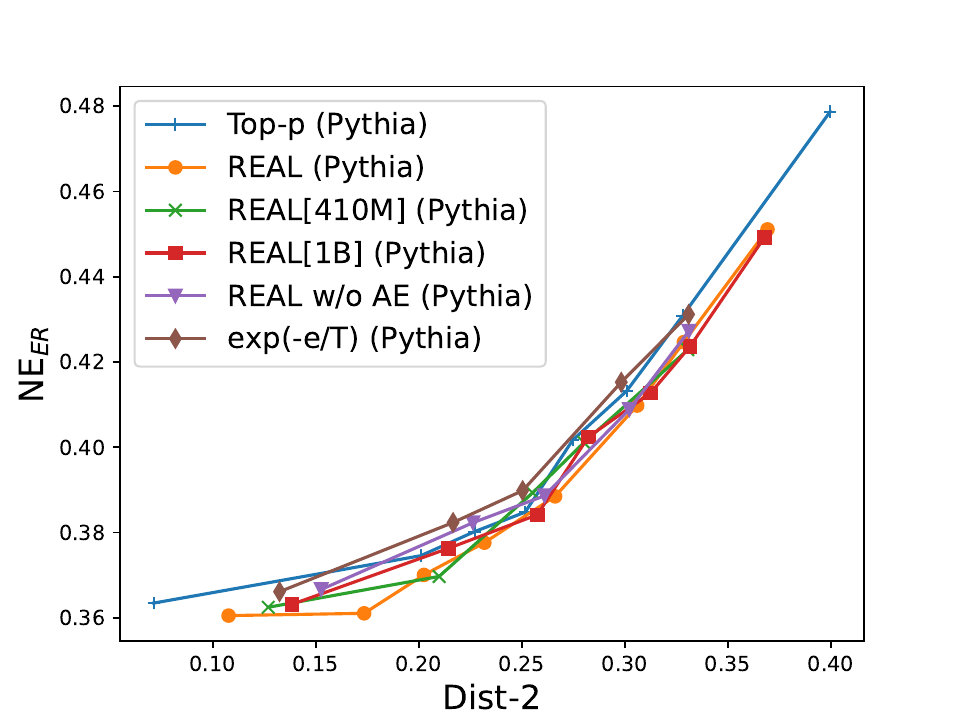}
  \caption{Ablation (Factual)}
  \label{fig:ablation_factual_ne}
\end{subfigure}
\begin{subfigure}{.33\textwidth}
  \centering
  \includegraphics[width=1\linewidth]{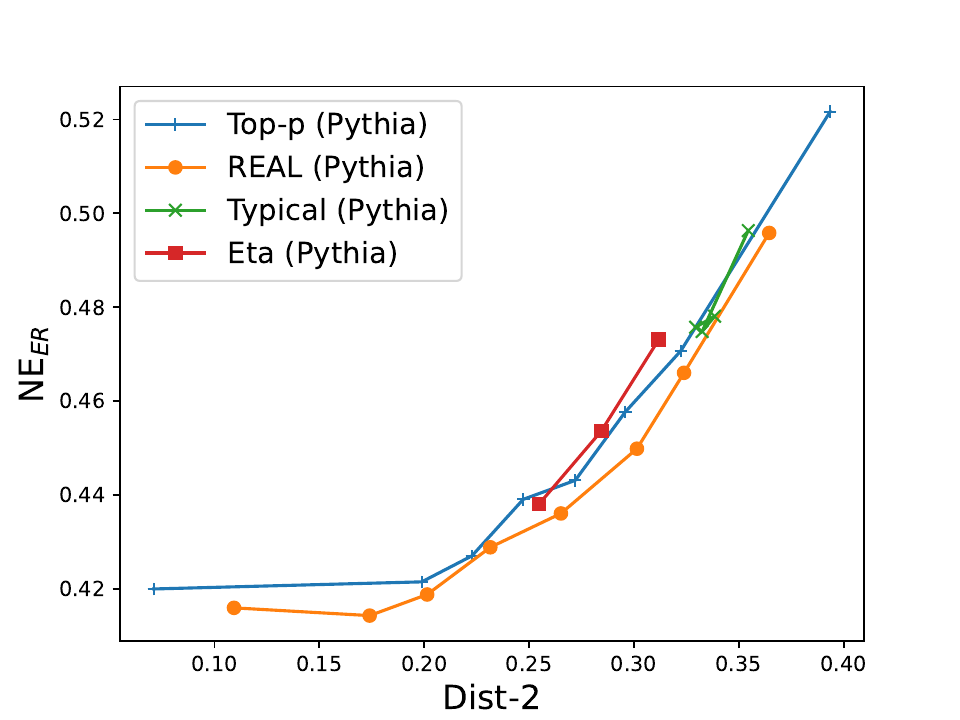}
  \captionsetup{justification=centering}
  \caption{REAL vs Thresholding Methods (Nonfactual)}
  \label{fig:topp_nonfactual_ne}
\end{subfigure}%
\begin{subfigure}{.33\textwidth}
  \centering
  \includegraphics[width=1\linewidth]{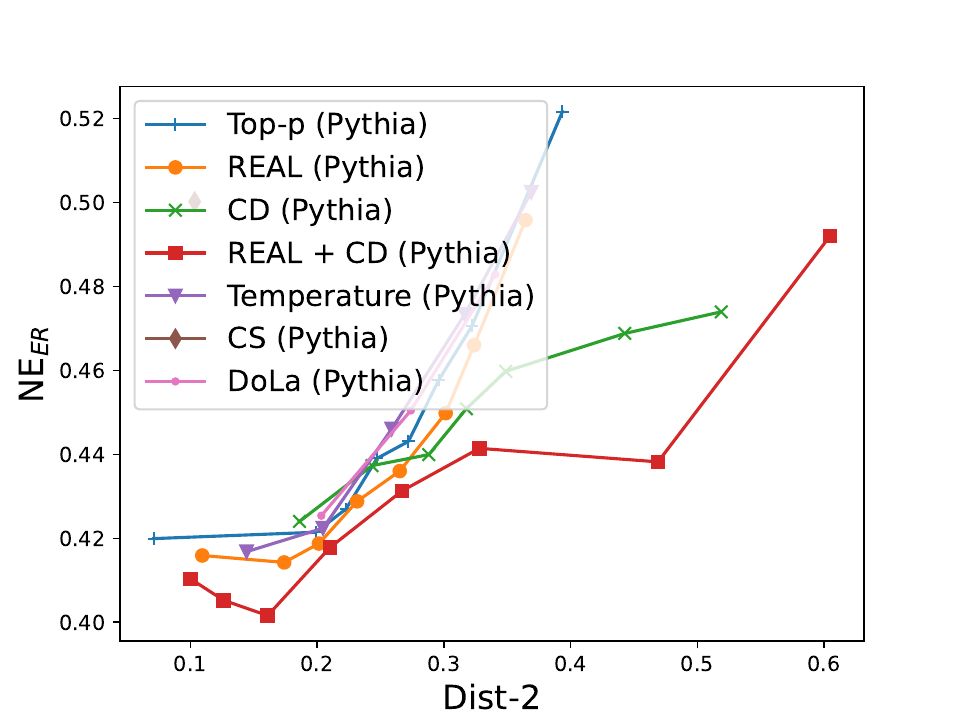}
  \captionsetup{justification=centering}
  \caption{REAL vs Distribution Modifications (Nonfactual)}
  \label{fig:cd_nonfactual_ne}
\end{subfigure}%
\begin{subfigure}{.33\textwidth}
  \centering
  \includegraphics[width=1\linewidth]{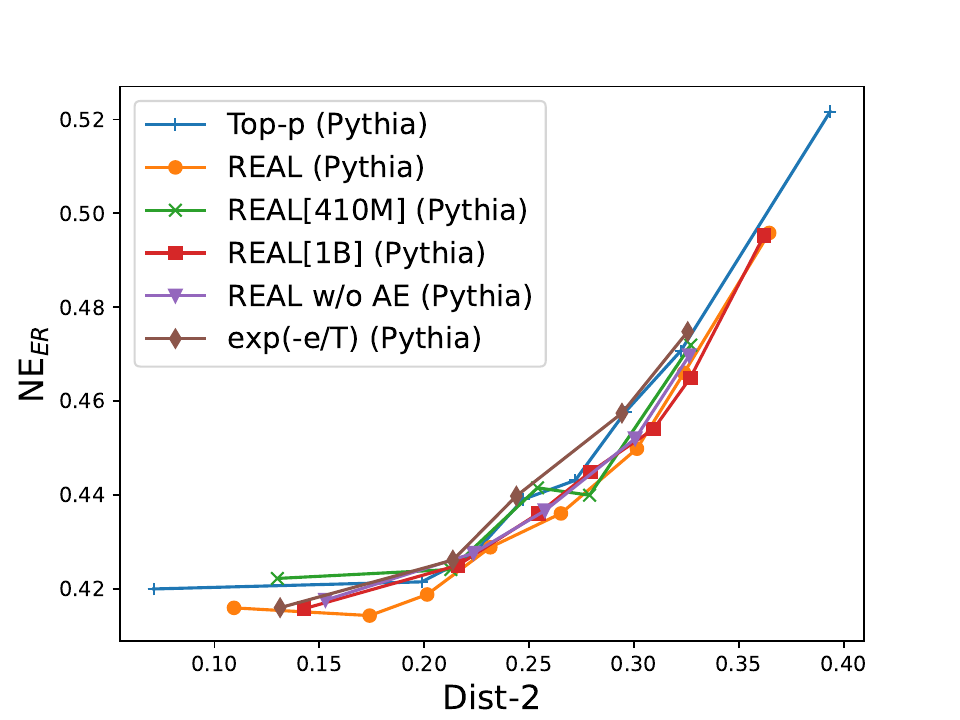}
  \caption{Ablation (Nonfactual)}
  \label{fig:ablation_nonfactual_ne}
\end{subfigure}
\begin{subfigure}{.33\textwidth}
  \centering
  \includegraphics[width=1\linewidth]{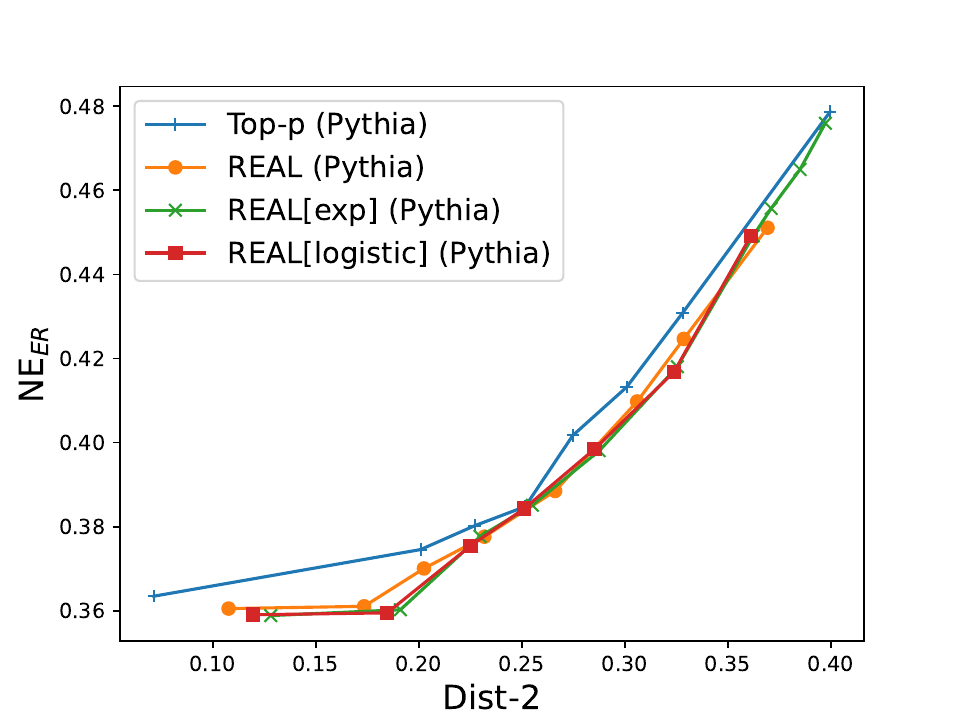}
  \captionsetup{justification=centering}
  \caption{Different Parameterizations \\ (Factual)}
  \label{fig:func_factual_ne}
\end{subfigure}%
\begin{subfigure}{.33\textwidth}
  \centering
  \includegraphics[width=1\linewidth]{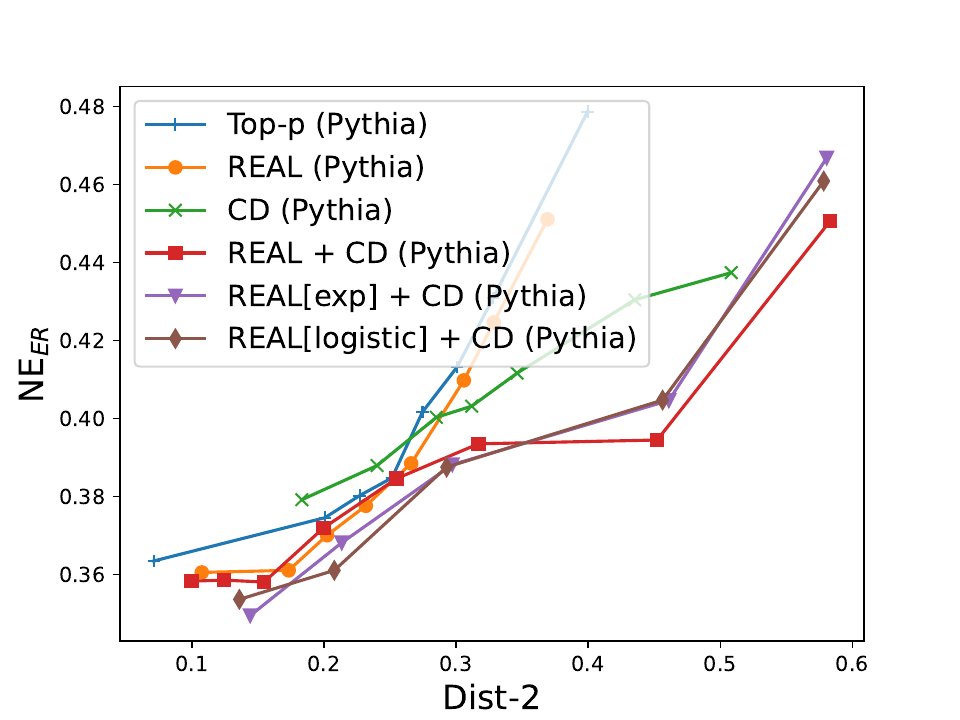}
  \captionsetup{justification=centering}
  \caption{Different Parameterizations \\ (Factual)}
  \label{fig:func_cd_factual_ne}
\end{subfigure}%
\begin{subfigure}{.33\textwidth}
  \centering
  \includegraphics[width=1\linewidth]{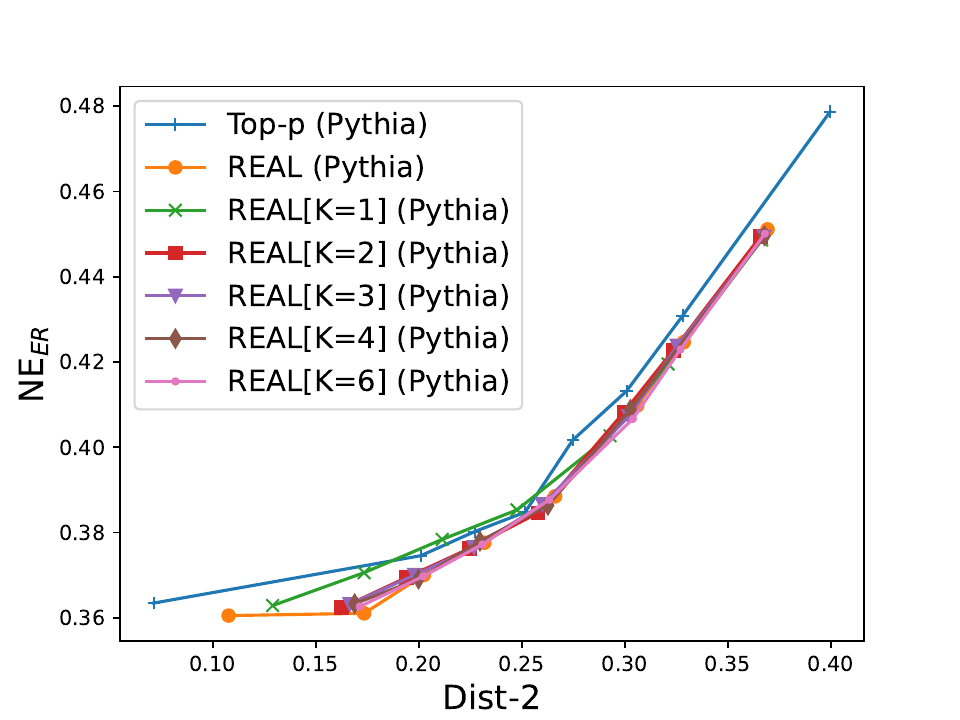}
  \captionsetup{justification=centering}
  \caption{Different Polynomial Degrees \\ (Factual)}
  \label{fig:K_factual_ne}
\end{subfigure}
\begin{subfigure}{.33\textwidth}
  \centering
  \includegraphics[width=1\linewidth]{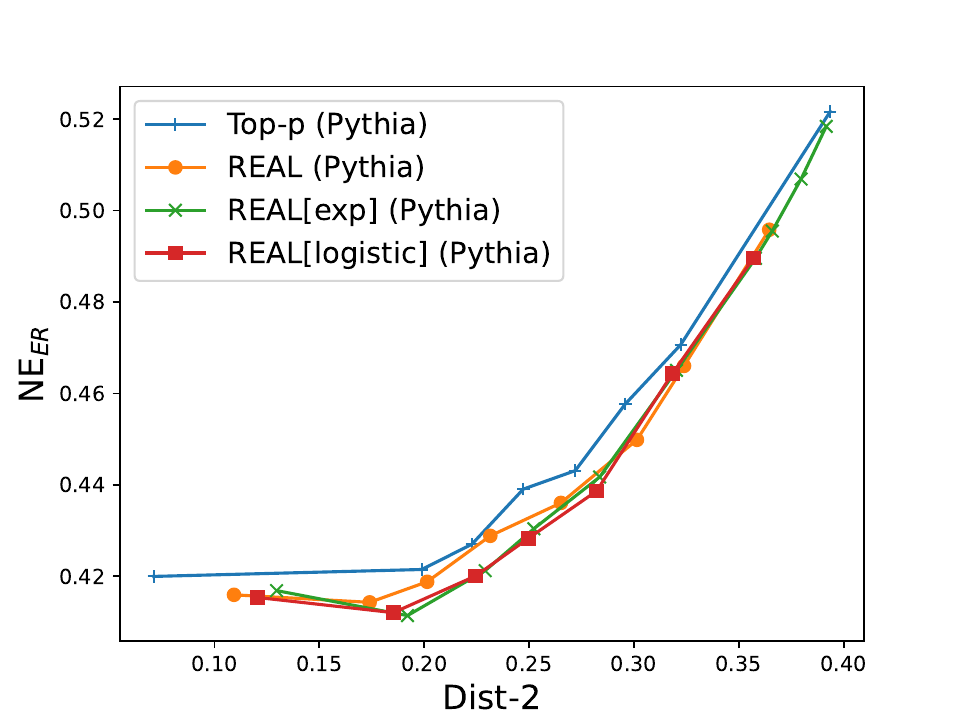}
  \captionsetup{justification=centering}
  \caption{Different Parameterizations \\ (Nonfactual)}
  \label{fig:func_nonfactual_ne}
\end{subfigure}%
\begin{subfigure}{.33\textwidth}
  \centering
  \includegraphics[width=1\linewidth]{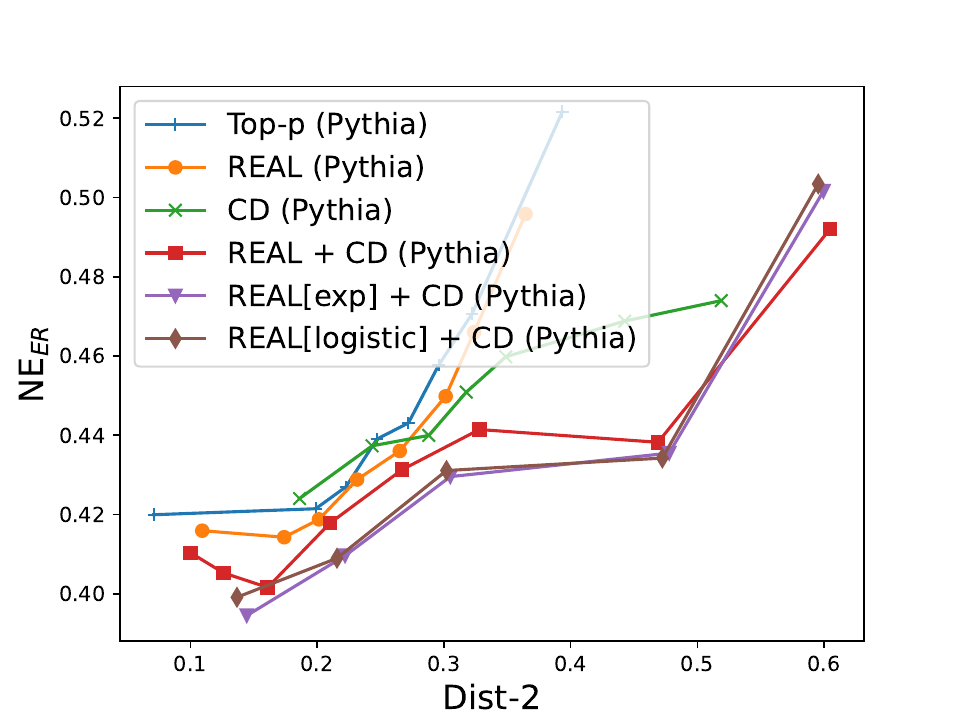}
  \captionsetup{justification=centering}
  \caption{Different Parameterizations \\ (Nonfactual)}
  \label{fig:func_cd_nonfactual_ne}
\end{subfigure}%
\begin{subfigure}{.33\textwidth}
  \centering
  \includegraphics[width=1\linewidth]{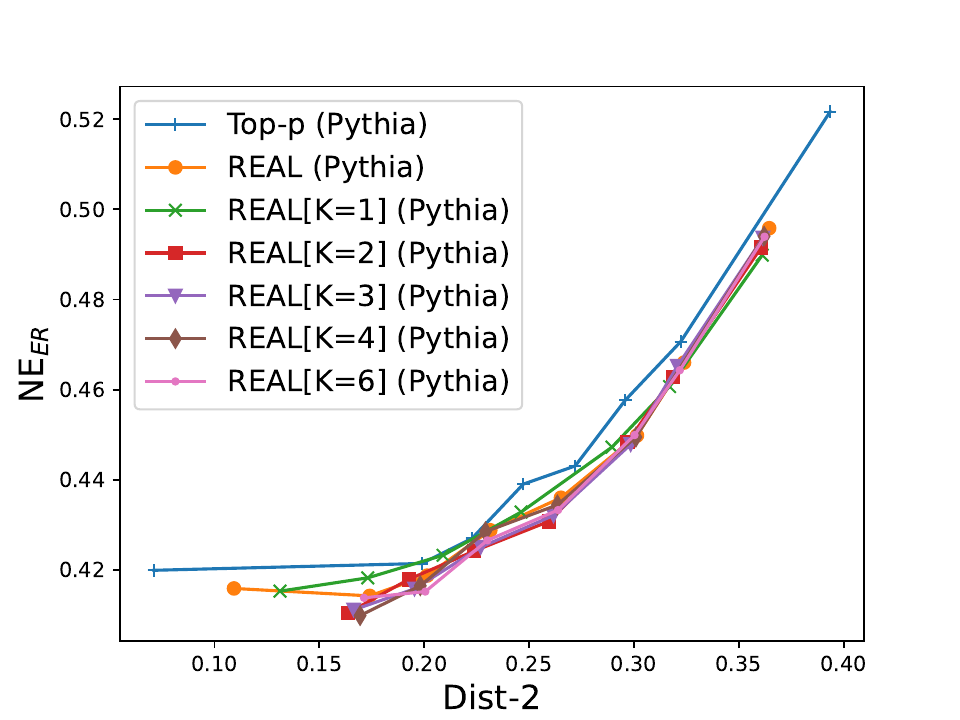}
  \captionsetup{justification=centering}
  \caption{Different Polynomial Degrees \\ (Nonfactual)}
  \label{fig:K_nonfactual_ne}
\end{subfigure}
\begin{subfigure}{.25\textwidth}
  \centering
  \includegraphics[width=1\linewidth]{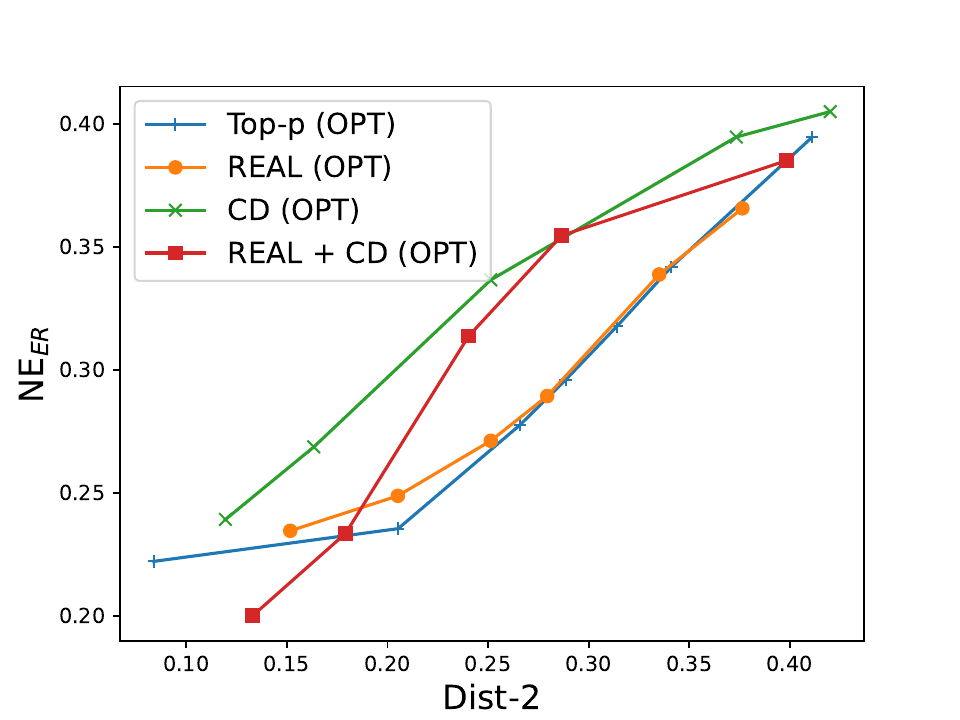}
  \caption{OPT-6.7b (Factual)}
  \label{fig:opt_factual_ne}
\end{subfigure}%
\begin{subfigure}{.25\textwidth}
  \centering
  \includegraphics[width=1\linewidth]{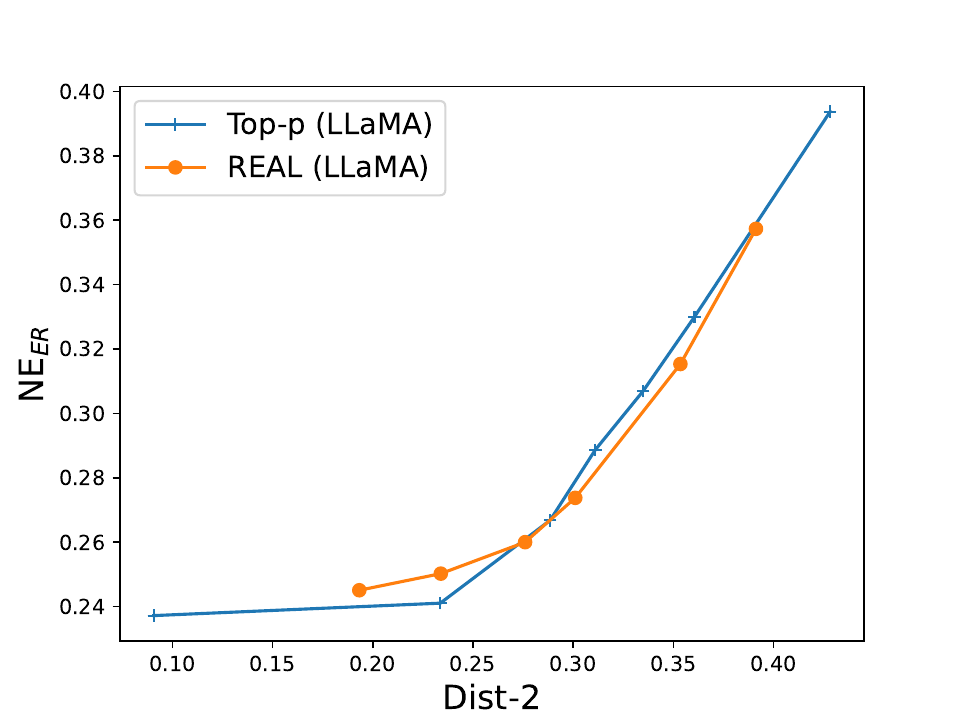}
  \captionsetup{justification=centering}
  \caption{OpenLLaMA-7b \\ (Factual)}
  \label{fig:llama_factual_ne}
\end{subfigure}%
\begin{subfigure}{.25\textwidth}
  \centering
  \includegraphics[width=1\linewidth]{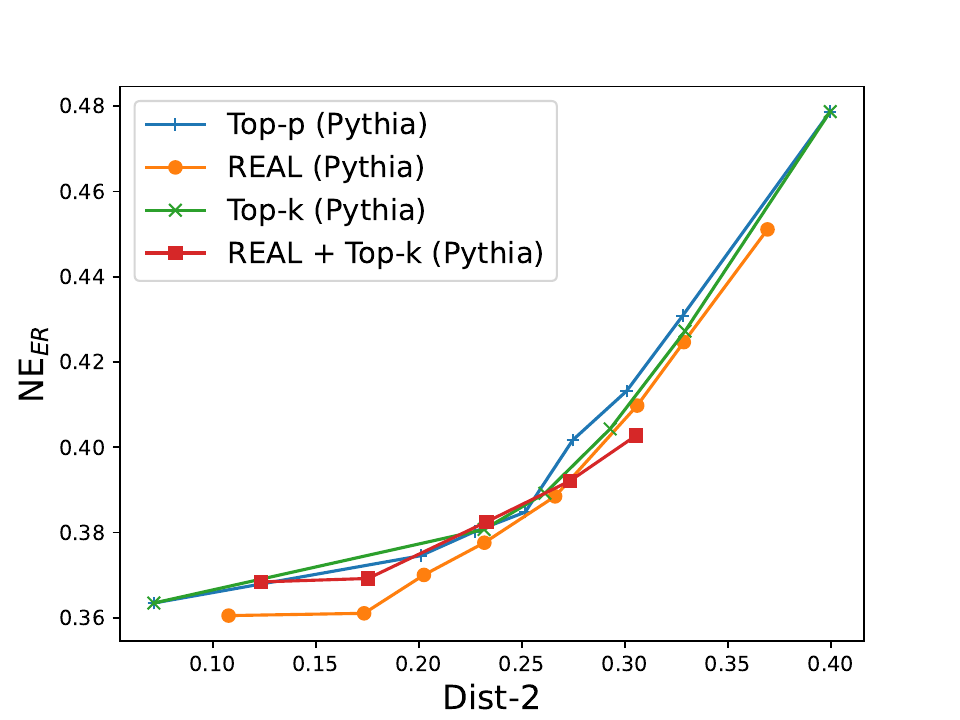}
  \captionsetup{justification=centering}
  \caption{REAL vs Top-k \\ (Factual)}
  \label{fig:topk_factual_ne}
\end{subfigure}%
\begin{subfigure}{.25\textwidth}
  \centering
  \includegraphics[width=1\linewidth]{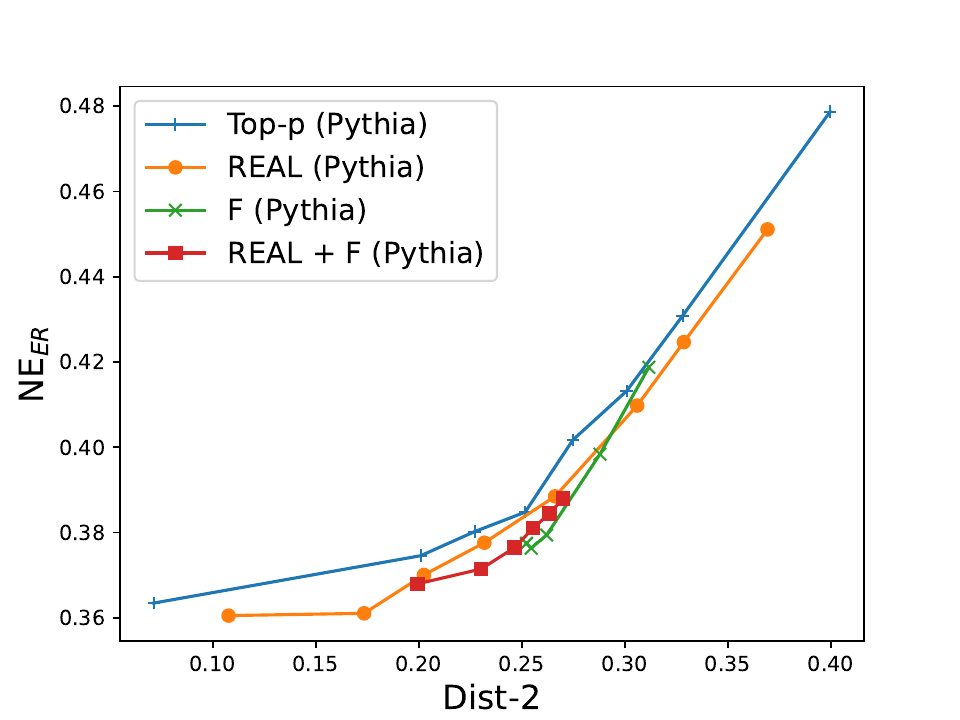}
  \caption{REAL vs F (Factual)}
  \label{fig:lee_factual_ne}
\end{subfigure}
\begin{subfigure}{.25\textwidth}
  \centering
  \includegraphics[width=1\linewidth]{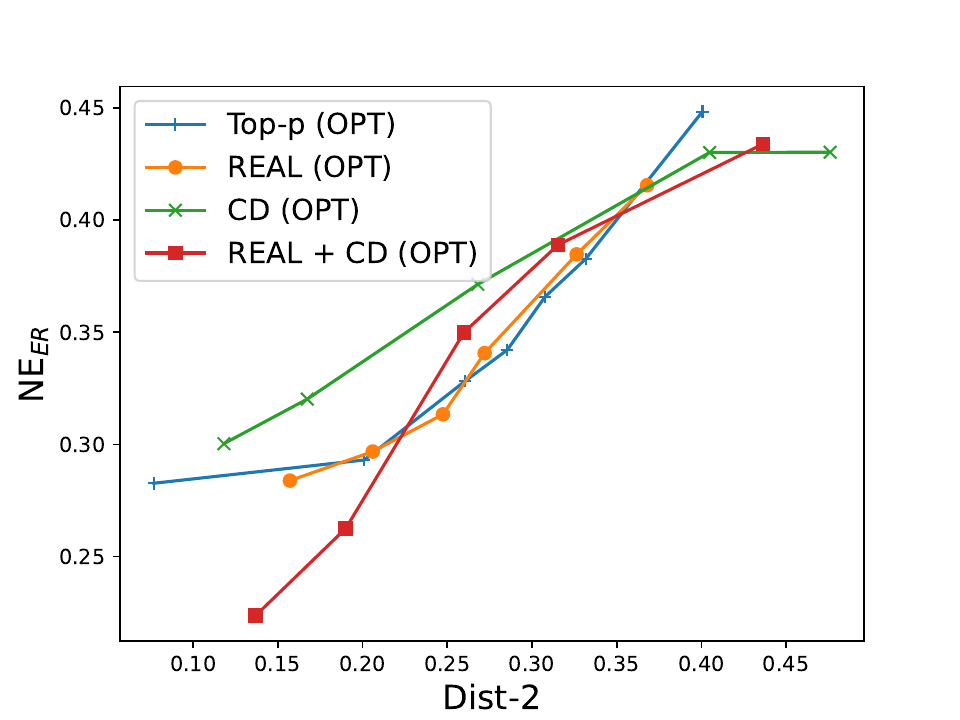}
  \caption{OPT-6.7b (Nonfactual)}
  \label{fig:opt_nonfactual_ne}
\end{subfigure}%
\begin{subfigure}{.25\textwidth}
  \centering
  \includegraphics[width=1\linewidth]{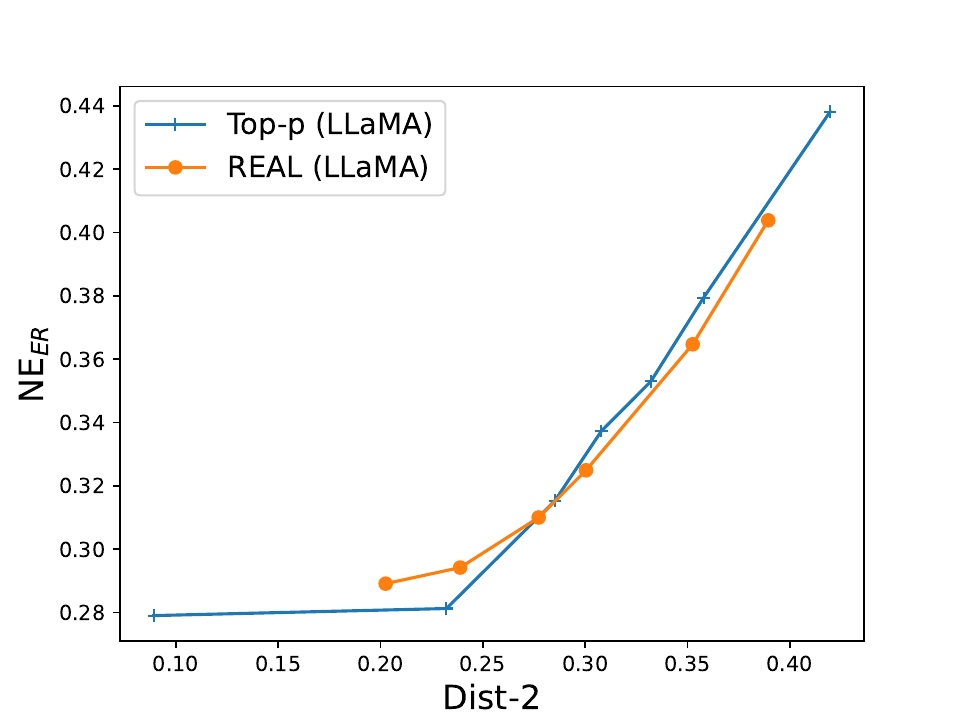}
  \captionsetup{justification=centering}
  \caption{OpenLLaMA-7b \\ (Nonfactual)}
  \label{fig:llama_nonfactual_ne}
\end{subfigure}%
\begin{subfigure}{.25\textwidth}
  \centering
  \includegraphics[width=1\linewidth]{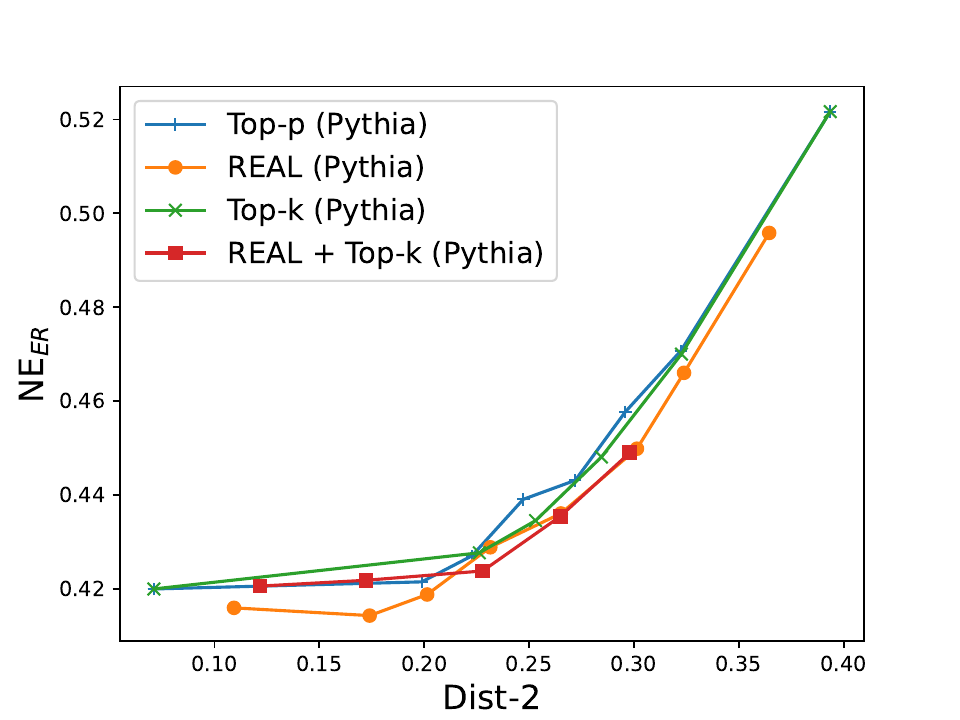}
  \captionsetup{justification=centering}
  \caption{REAL vs Top-k \\ (Nonfactual)}
  \label{fig:topk_nonfactual_ne}
\end{subfigure}%
\begin{subfigure}{.25\textwidth}
  \centering
  \includegraphics[width=1\linewidth]{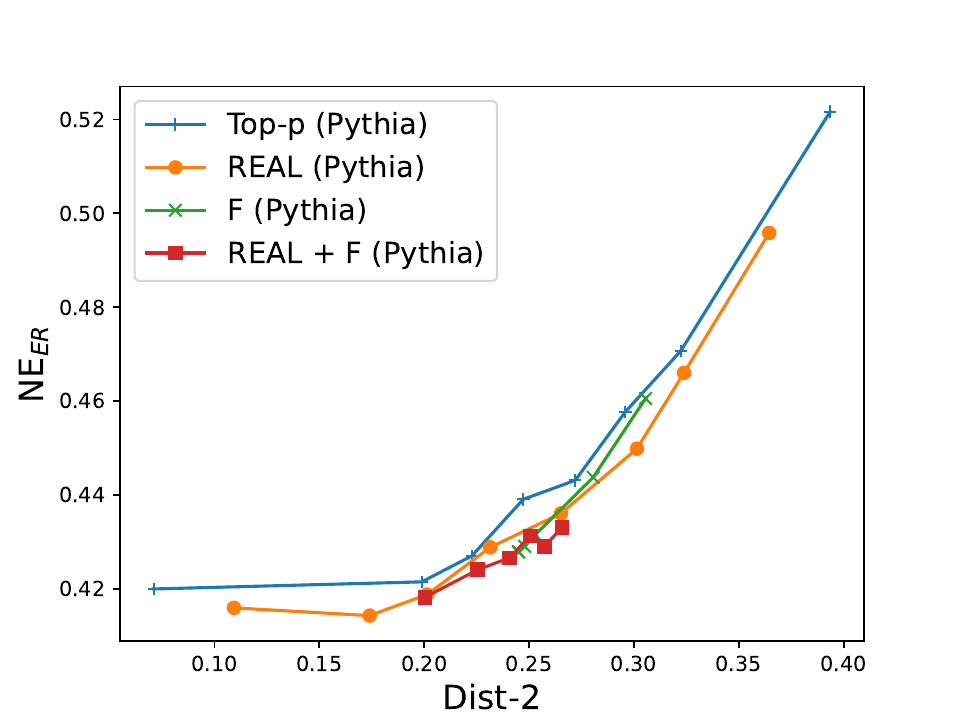}
  \caption{REAL vs F (Nonfactual)}
  \label{fig:lee_nonfactual_ne}
\end{subfigure}
\end{adjustbox}
\caption{The named entity error ratio (NE$_{ER}$) versus distinct bi-gram (Dist-2). Lower NE$_{ER}$ is better, so the better methods are closer to the bottom-right corner. (Factual) in the captions means the prompt sentence is factual. We hide the hyperparameter values in the figures to avoid blocking the curves. The y-axis standard errors of every curve in this figure are 0.002 on average and smaller than 0.006.} 
\label{fig:comp_gen_ne}
\end{figure*}

%\subsection{Comparison with top sampling}

%top k sampling
%\citep{fan2018hierarchical}

\subsection{Individual Metrics}
\label{sec:org_metrics}

We present the performances of entailment ratio (Entail$_R$) and repetition ratio (Rep) in \Cref{fig:comp_gen_entail}. We also present the performances of named entity error ratio (NE$_{ER}$) and distinct bi-gram (Dist-2) in \Cref{fig:comp_gen_ne}. In both figures, we separate the performances given factual prompts and nonfactual prompts. We can see that the overall trends are similar compared to \Cref{fig:comp_gen}. The better scores for the nonfactual prompts show that our methods are less likely to propagate factual errors when generating the text~\citep{zhang2023language}. Our improvements are larger in \Cref{fig:comp_gen_entail}. It suggests that our methods are especially effective in alleviating the repetition problem. Surprisingly, we observe that REAL sampling could be even slightly more factual than the greedy decoding (Top-$p$ with $t^p=0$), which is reported as the leftmost points in \Cref{fig:topp_factual_ne} and \Cref{fig:topp_nonfactual_ne}. This suggests greedy decoding might not always have less hallucination compared to the sampling.

\subsection{Speed Comparison}
\label{sec:speed}

Since the size of our 70M THF model is 100 times smaller than 7B LLM, the inference time of the THF model should be negligible if we parallelly run both the LLM and THF model at each decoding step. Without optimizing for speed, our current implementation simply runs the 70M THF model at every decoding step of LLM. In this naive implementation, the decoding time still only increases around 11\% (from 7.46 to 8.29 seconds) when the batch size is 8 and the maximal number of tokens is 128. 

%also predict the entropy of the last LLMs directly ($\theta_{s_N}$)

\begin{table}[t!]
\caption{Human experiment and automatic metrics for \textsc{FactualityPrompts}. }
\label{tb:human_abs_scores}
\scalebox{0.78}{
\begin{tabular}{c|cccc|cccc}
        & \multicolumn{4}{c|}{Human experiments (100  continuations)}        & \multicolumn{4}{c}{Automatic Metrics (28k continuations) (\%)}          \\
        & Overall        & Factuality     & Informativeness & Fluency        & Dist-2          & Rep ($\downarrow$)            & NE$_{ER}$ ($\downarrow$)          & Entail$_R$       \\ \hline
Top-$p$ ($p=0.6$)   & 2.55          & 2.15          & 3.46           & 4.01          & \textbf{27.466} & 3.736          & 40.171          & 7.925           \\
REAL ($T=2.0$)   & \textbf{2.78} & \textbf{2.32} & \textbf{3.53}  & \textbf{4.14} & 26.608          & \textbf{3.386} & \textbf{38.850} & \textbf{9.377}  \\ \hline
CD ($\alpha=0.3$)     & 2.84          & 2.47          & 3.61           & 4.08          & \textbf{28.511} & 4.143          & 40.031          & 9.380           \\
REAL+CD ($T=1.5$) & \textbf{2.86} & \textbf{2.48} & \textbf{3.64}  & \textbf{4.09} & 25.505          & \textbf{3.193} & \textbf{38.462} & \textbf{11.394}
\end{tabular}
}
\end{table}

\subsection{Human Experiment for \textsc{FactualityPrompts}}
\label{sec:human_stats}

In \Cref{tb:human_abs_scores}, we provide the average scores of each method. In the human experiment, we can see that the factuality improvements of \textbf{REAL} over \textbf{Top-$p$} are larger than the improvements of \textbf{REAL + CD} over \textbf{CD}, while the order is reversed in the automatic metrics. It might be due to the relatively small testing set in the human experiment, or some factors that can be measured by human evaluation. 

The average Pearson correlation between the two workers in every task is 23.5\% for overall, 37.3\% for factuality, 14.2\% for informativeness, and 12.3\% for fluency. Notice that we only change the truncation threshold in the sampling methods on top of the same generation LLM, so the generated next sentences are sometimes very similar. This makes workers sometimes hard to give different scores to different generations.
We observe that the agreements of informativeness and fluency are low while their average absolute scores are high. One possible reason is that all generations have similarly good fluency, so workers tend to disagree about which ones are slightly less fluent.

%\section{Analysis of only Predicting LLM's Entropy}
\section{More Result Analyses}
\label{sec:ent_baseline}

In \Cref{fig:cd_factual}, we observe that REAL sampling could double the improvements of contrastive decoding (CD). One possible reason is that CD might reduce the diversity when there are many correct next tokens and REAL sampling could alleviate the problem. For example, in \Cref{fig:asymptotic_entropy} (b), the distributions of small LM could be slightly flatter version of LLM's distribution. Then, CD could significantly reduce the probability of less likely options such as \textit{When} and thus lead to lower diversity. REAL sampling could fix the problem by dynamically increasing the $p$ threshold to include more options.
%We hypothesize that 

In \Cref{fig:ablation_factual}, we observe that \textbf{REAL w/o AE}, whose $\hat{t}_c^p = \exp(\frac{- e_c(s_N) }{T})$, performs better than \textbf{Top-$p$} and \textbf{exp(-e/T)} sampling, whose $\hat{t}_c^p = \exp(\frac{- e_c^{\theta_{s_N}}}{T})$. We want to know if the improvement comes from the extra entropy information from the Pythia LLM family, so we use the THF model to predict the $e_c^{\theta_{s_N}}$ directly. The preliminary result in our validation set shows that \textbf{REAL w/o AE} performs similarly if we replace the $e_c(s_N)$ with the direct prediction, so the improvement does not come from the entropy information of other smaller LLMs. 

This is an interesting finding. We hypothesize that the improvement of \textbf{REAL w/o AE} comes from smoothing the LLM's entropy $e_c^{\theta_{s_N}}$ by the prediction of the tiny model (e.g., the LLM might have very different entropies for the three similar context in \Cref{fig:curve_prediction}, while the tiny model tends to output the average of these entropies). That is, it is not a good strategy to select less lower probability tokens when the entropy of its distribution is high. Instead, it is better to ask the LLM to be more careful when the next token is generally hard to predict (e.g., being an entity name) regardless of LLM's certainty to the next token.
%if the LLM is uncertain about the next token or not.

%agreement between humans

%wiki
%Pearson correlation:
%All: 23.5
%Factuality: 37.3
%Informativeness: 14.2
%Fluency: 12.3
%Low because factuality is hard to check, the metric is subjective, and the generations are often very similar and hard to judge which one is better.

\begin{table}[t!]
\caption{Creative writing experiment in an out-of-domain setting. The generation model is Pythia 6.9B and we measure the winning rate against Top-p sampling with $p=0.5$ using GPT3.5. }
\label{tb:creative}
\vspace{2mm}
\centering
\begin{tabular}{c|cccc|cc}
                  & \multicolumn{4}{c|}{ Wining Rate using GPT3.5 (500 continuations)}        & \multicolumn{2}{c}{(8k continuations)} \\
                  & Fluency        & Coherence      & Likability & Overall        & \multicolumn{1}{c}{Dist-2}    & \multicolumn{1}{c}{Rep ($\downarrow$)}    \\ \hline
Top-$p$   ($p=0.5$) & 50 & 50          & 50  & 50          & \textbf{18.600}               & 7.463                      \\
REAL ($T=1.8$)    & \textbf{53}          & \textbf{53.4} & \textbf{52.6}           & \textbf{52.6} & 17.952                        & \textbf{4.563}             \\ \hline
            
\end{tabular}
\end{table}

\section{Human Experiment for Creative Writing in an Out-of-Domain Setting}
\label{sec:creative}

Creative writing is not the focus of this paper because the hallucination problem is usually not serious in the tasks. Nevertheless, we still evaluate our methods on a story-writing task. In the task, the prompt is composed of three stories from the ROC story dataset~\citep{mostafazadeh2016corpus} and the first two sentences from the fourth story. Then, we use different decoding methods to complete the fourth story. %We adjust the hyperparameters a little from the previous human experiment to make the diversity of \textbf{top-$p$} sampling and \textbf{REAL} similar. 

Evaluating the creative writing is a subjective task. To save cost, we use \textit{gpt-3.5-turbo-0125} to compare the continuations of 500 ROC stories from REAL sampling and from top-$p$ sampling. In \Cref{tb:creative}, we report the winning rate against the top-$p$ sampling.

%and make the diversity of \textbf{CD} and \textbf{REAL + CD} similar.

%Given the first two sentence
%ROC story

\expsec{Results:} Overall, \textbf{REAL} is similar \textbf{top-$p$} sampling even when our training data for the THF model (i.e., Wikipedia and OpenWebText) does not include too many short stories. This shows that REAL sampling could improve the factuality of \textbf{top-$p$} sampling without sacrificing its creative writing ability.
%This is probably because REAL sampling could reduce repetition and the chance of ``hallucination'' in a story (e.g., introducing new entities out of nowhere), which brings about higher coherence and fluency scores.

%The higher interestingness of \textbf{top-$p$} might also come from its incoherency in the story. 

%In all the metrics, \textbf{REAL + CD} is worse than contrastive decoding \textbf{CD}. One possible reason could be that the contrastive decoding changes the distribution of LLM very substantially, so our THF model that is trained on the original LLM's distribution also needs to be adjusted accordingly. How to better combine our method with contrastive decoding is a promising future research topic.

%The average Pearson correlation between the two workers in every task is 26.5\% for overall, 21.2\% for coherence, 40.3\% for informativeness, and 35.6\% for fluency. LLM often needs to output more than three sentences to complete the story, so the differences among the continuations are usually much larger compared to the previous human experiment. This explains why the agreements between the annotators are generally larger than the agreements in the other human experiment. 

\section{Impact Statements}
\label{sec:impact}

REAL sampling has strong open-ended text generation performances without supervision. For example, typical~\citep{meister2022typical} sampling, a state-of-the-art thresholding method, improves top-$p$ by 0.4\% (4.13->4.15) based on the human evaluation scores in their Table 1, while REAL sampling improves top-$p$ by 9\% (2.545->2.775) in \Cref{tb:human_abs_scores}. Furthermore, our method has a wide application on LLMs because we did not make any domain-specific assumptions and REAL sampling can be combined with other decoding methods to achieve further improvements. This suggests that our method can potentially improve most of the state-of-the-art LLMs such as GPT-4 in all the domains. 

%We believe that improving 5-10% factuality and/or diversity on top of the LLMs like GPT4 is a large contribution.
%In Table 1 of typical sampling (Meister et al., 2022), their human score improvement over top-p is only 0.4\% (4.13->4.15).

Low generation factuality/quality could incur negative societal implications. For example, LLM could generate a fake case for a lawyer to cite~\citep{lawyer_bad_cite} or a code with some subtle bugs for a software engineer to use.
Our hallucination forecasting model could be used to give users better hallucination warning by improving the current token highlighting based on its entropy or probability (similar to what we did in \Cref{fig:RE_vis}). 

Low generation diversity also prevents the users from knowing different perspectives on a question and might increase the polarization of society. Thus, increasing both factuality and diversity could bring positive social impacts. Our unsupervised training methods could also bring the above benefits to other languages more easily.
Nevertheless, increasing the factuality might also make it harder to distinct AI-generated text from human-written text and thus lead to some social problems. Discussing these potential social benefits and harms is out of the scope of this paper.

%working with a LLM, a lawyer might cite a fake case~\citep{lawyer_bad_cite} and a software engineer might submit a buggy code if they do not discover the mistakes made by LLMs. 

\section{Limitations and Future Work}
\label{sec:future}

%\hs{add limitation}

We assume the target LLM family has models with different sizes. Although it is not always the case, our experiments show that the THF models trained in Pythia could still help OPT and OpenLLaMA, which suggests the existence of an universal THF model that could be generalized to the other LLM families without models with different sizes.

To motivate/explain our method, we assume that there is an ideal next-token distribution and the infinitely large LLM that can output this ideal distribution. Hence, the entropy of the ideal next-token distribution $e^{AE}_c$ exists. Nevertheless, without the assumption, we still know that the entropy of a larger LLM should be a better estimation of the inherent uncertainty, so we extrapolate the existing entropy decay curves to estimate the entropy of a very large LLM. Therefore, the theoretical existence of $e^{AE}_c$ does not affect the empirical/practical value of our extrapolation and our derivation in Theorem 3.1.

%\hs{visualization could also be one of our applications.}

The unsupervised nature of REAL sampling makes it applicable to many domains and applications. For example, we could apply it to larger LLM. As shown in \Cref{fig:RE_vis}, the entropy from 6.9B LLM is still very far away from asymptotic entropy, so we should get much more accurate estimations when we use a even larger LLM. We could use it to improve the code generation~\citep{li2022competition} or combine REAL sampling with the prompting methods that require both diversity and factuality~\citep{wang2022self,bertsch2023s,yao2023tree,naik2023diversity} or the hallucination reduction methods for specific applications
\citep{van2022mutual,marfurt-etal-2022-corpus,chang2023kl,shi2023trusting}. We could also use the THF model to improve the factuality/effectiveness/efficiency of beam search
~\citep{wan2023faithfulness,tu2023unlocking}.

There are also several potential improvements for the THF model and REAL sampling. For example, our current ways of combining REAL sampling and other decoding methods are very simple. We could study how to better integrate these decoding methods. Moreover, we want to better support the off-the-shelf usage of the THF model. In \Cref{fig:ablation_factual}, we can see that the same temperature $T$ corresponds to two very different diversities for \textbf{REAL} and \textbf{REAL (410M)} and should be normalized based on the average RE ($\hat{d}_c^{RE}$). Besides, we can train a model that predicts the entropies of multiple LLM families to reduce the performance degradation in the out-of-domain transfer setting (e.g., training on Pythia and testing on OPT). 
%We can also try other ways to parameterize the entropy decay. 
In industry, running two models with very different sizes together might induce some engineering challenges or synchronization issues. Therefore, we can try to integrate the THF model with LLMs. One possible approach is to use the prediction of the THF model as a noisy signal to find the hallucination patterns of the hidden state as \citet{burns2022discovering, li2023inference,chuang2023dola} did.

%In the future, we can replace the training signal with residual entropy. 
%Then, we can use the hidden states of LLM to estimate residual entropy further increase the inference speed.

%Integrate REAL into the LLM itself
%\citep{chuang2023dola}

%the good temperature depends on the REAL LM model (e.g., the same temperature corresponding to different diversities for 410M model and for 70M model. )

%Our unsupervised training method 
%could potentially be used in many applications. 

%see if REAL sampling could improve the tasks or 

%prevent the hallucination using 

%if we can increase the downstream applications that require both diversity and factuality.

%Some language would put the object earlier

\section{Why does the Entropy Decay as the Model Size Increases?}
\label{sec:decrease_reason}

%\hs{The entropies are averaged across all tokens}

First, in \Cref{fig:ent_decay_vis}, we empirically observe that the average entropy across our Wikipedia validation set (around 9M tokens) steadily decreases as the model size increases. Furthermore, there are 90.2\% contexts given which the smallest Pythia LM (70M) has a larger next-token entropy compared to Pythia LLM (6.9B). We visualize some of the decay curves in \Cref{fig:RE_vis}.

%In our validation set (~9M tokens), %In \Cref{fig:ent_decay_vis}, we can see that the 
%we plot the average empirical next-token entropy curve and average predicted entropy curve in our validation set. 

%The theoretical reason for LLM's entropy decay is an interesting research topic, which is out of the scope of this paper. Nevertheless, we can provide one intuitive explanation and one formal explanation for n-gram LMs.

Intuitively speaking, a small language model is less likely to learn the ideal distribution, so it tends to put higher probabilities on more words so that it won’t receive a large penalty from the cross-entropy loss. Since its output distribution is closer to a uniform distribution, the entropy is higher.

We can also provide a more formal explanation by treating
a smaller LLM as a n-gram LM with a smaller n. To simplify our explanation, let’s just assume our vocabulary is {A,B,C} and we want to show the average entropy 1-gram LM is larger than the average entropy 2-gram LM, which predicts the next word just based on one context word. Let’s denote the probability of seeing the word x as P(x) and the probability of seeing the word y given the context x is P(y|x). Since the entropy function is a concave function, we know that entropy of 2-gram LM = $\sum_{x=A,B,C} P(x) Ent( P(y|x) ) \leq Ent( \sum_{x=A,B,C} P(y|x) P(x) )$ = the entropy of 1-gram LM. The intuitive explanation of this proof is that the probability distribution of 1-gram LM merges the 3 distributions of 2-gram LM, and merging distributions would lead to a higher entropy overall. We can easily generalize the above proof to show that the average entropy of n-gram LM is always larger than the average entropy of (n+1)-gram LM.

%Finally, since the entropy tends to decrease

\section{Method Details}
\label{sec:method_details}

In open-ended text generation, we empirically observe that the RE $\hat{d}_c^{RE}$ gradually decreases as the context length increases because the LLM tends to be more certain about the next token given a long context. To avoid the systematic shift of $\hat{t}_c^p$, we only input the last $40$ tokens into the THF model. This truncation also further reduces the computational cost for a long context input and stablize the estimation of the curve parameters by limiting the prediction power of the tiny THF model~\citep{li2022contrastive}.

%Initialization of parameter weights to prevent exp explode 

%\subsection{THF Model Training}
%\label{sec:REAL_training}

We train our models using Wikipedia 2021 and OpenWebText~\citep{radford2019language}. In the training corpus, we first compute the entropies of each word using the Pythia with sizes 70M, 160M, 410M, 1B, 1.4B, 2.8B, and 6.9B. When computing the Log(model size), we use the number of parameters after excluding the token embeddings. We set the highest degree of our fractional polynomial $K = 10$ by default and fine-tune the pretrained Pythia 70M for 3 epochs to predict their entropy decay curves. We set the learning rate as $5e-5$ and warm-up step as $100$. Furthermore, we initialize all values in the weight and bias of the linear layer before the final exponential layer with $0$ to prevent our exponential layer from causing too large gradients at the beginning of training. 
%\subsection{Training Details}

We download the Wikipedia from \url{http://medialab.di.unipi.it/wiki/
Wikipedia_Extractor} and download OpenWebText~\citep{radford2019language} from \url{https://github.com/jcpeterson/openwebtext} (GPL-3.0 license). We only use the first 5M lines of both datasets (around 5.6\% of text) to accelerate our training because our preliminary studies show that our performance is not sensitive to the training corpus.

During training, the maximal length of context is $1024$ to ensure that the THL model can handle the long context in hallucination detection. We set the batch size to be $128$ for 70M model and $32$ for 410M model based on the limit of our GPU memory. Our preliminary experiments show that the performances of text generation and hallucination detection are not sensitive to these hyperparameters.

%\hs{say we initialize the classifier weight and bias to be 0 (to make sure exp won’t make loss explode) (I wrote the code in the $train_entropy_prediction_model.py$)}

%\hs{Our model size does not include the embedding size}

%$K=10$ and batch sizes are determined by the validation set
%128 batch size for 70M
%32 batch size for 410M

%\subsection{Software and Hardware Resources}

\section{Experiment Details}
\label{sec:exp_details}

In our experiments, we always append a space at the beginning of the context for all generation LLMs due to the preference of LM tokenizer. All the training and experiments are done by 8 NVIDIA V100 32GB GPUs. Our code is built on Huggingface.

\subsection{Details for Open-Ended Text Generation}

%We use dist-2 instead of dist-4 because dist-4 sometimes saturate?
%The hyperparameter setting of \citet{lee2022factuality} and contrastive decoding.
For contrastive decoding (\textbf{CD})~\citep{li2022contrastive}, we fix the temperature for the amateur model to be $1$ and choose the smallest model in the LLM family as the amateur model (i.e., Pythia 70M in \textbf{CD} and OPT-125m in \textbf{CD (OPT)}). To make the comparison fair, we use sampling rather than beam search proposed in \citet{li2022contrastive}. For DoLa~\citep{chuang2023dola}, we try two layer subsets suggested in the paper: 0,2,4,6,8,10,12,14,32 and 16,18,20,22,24,26,28,30,32. We report the results of the former one because of its much better performance than the later one.
%, CD uses beam search. 

%\hs{The original CD use beam search, which requires more computational resources. we use sampling to make the comparison fair}

%\hs{explain the normalization details of Pythia and OPT}

%As in \citet{lee2022factuality}, we evaluate Entail$_R$, NE$_{ER}$, and Dist-2 using only the first generated sentence in the continuation by default to emphasize the difference between factual and nonfactual prompt. The results of first three 

The maximal length of the continuation is set as $128$. To allow the batch decoding during inference, we append $<$eos$>$ sequences before the input prompts.
When we compute Entail$_{Rn}$, NE$_{ERn}$, Dist-2$_n$, and Rep$_n$, we separate the max-min normalization for each LLM generation model and each prompt type (e.g., The decoding method for Pythia, OPT, or OpenLLaMA that achieve the highest Entail$_{R}$ given the factual prompts will all receive 1 in the Entail$_{Rn}$ metric for factual prompts).
%Across all decoding methods, the best Entail$_{Rn}$ score for Pythia/OPT/OpenLLaMA and is always 1). 

% we first normalize all metrics from a generation LLM using max-min normalization and average the scores from both factual prompts and nonfactual prompts as .

%we only look at the first sentence except in the repetition metrics
%repetition ratio
%max 128 token

%speed measurement details
%experts

\subsection{Details for Human Experiments}
\label{sec:human_details}

After having generated continuations from different methods in \textsc{FactualityPrompts}\footnote{\url{https://github.com/nayeon7lee/FactualityPrompt} Apache-2.0 license}, we first exclude the continuations that cause the difficulties in comparing the factuality, including the same continuations from different methods, the continuations that are less than 10 characters, and the continuations that mention ``External links''. For the story completion, we only keep the responses whose lengths are between 50-1000 characters. Then, we select the remaining top 100 testing factual prompts based on the original order of \textsc{FactualityPrompts} and randomly select 100 prompting stories.

We collaborate with a list of MTurk workers in multiple projects, so their annotation quality is much higher than the average MTurk workers. Then, we further manually filter MTurk workers based on the supporting URL and statements/reasons they provided. We control the hourly wage of these trusted MTurk workers to be around \$14 and provide \$2.2 reward for each task in \textsc{FactualityPrompts} and \$1.5 reward for each task in story generation.

\begin{figure*}[t!]
\centering
\begin{subfigure}{.5\textwidth}
  \centering
  \includegraphics[width=1\linewidth]{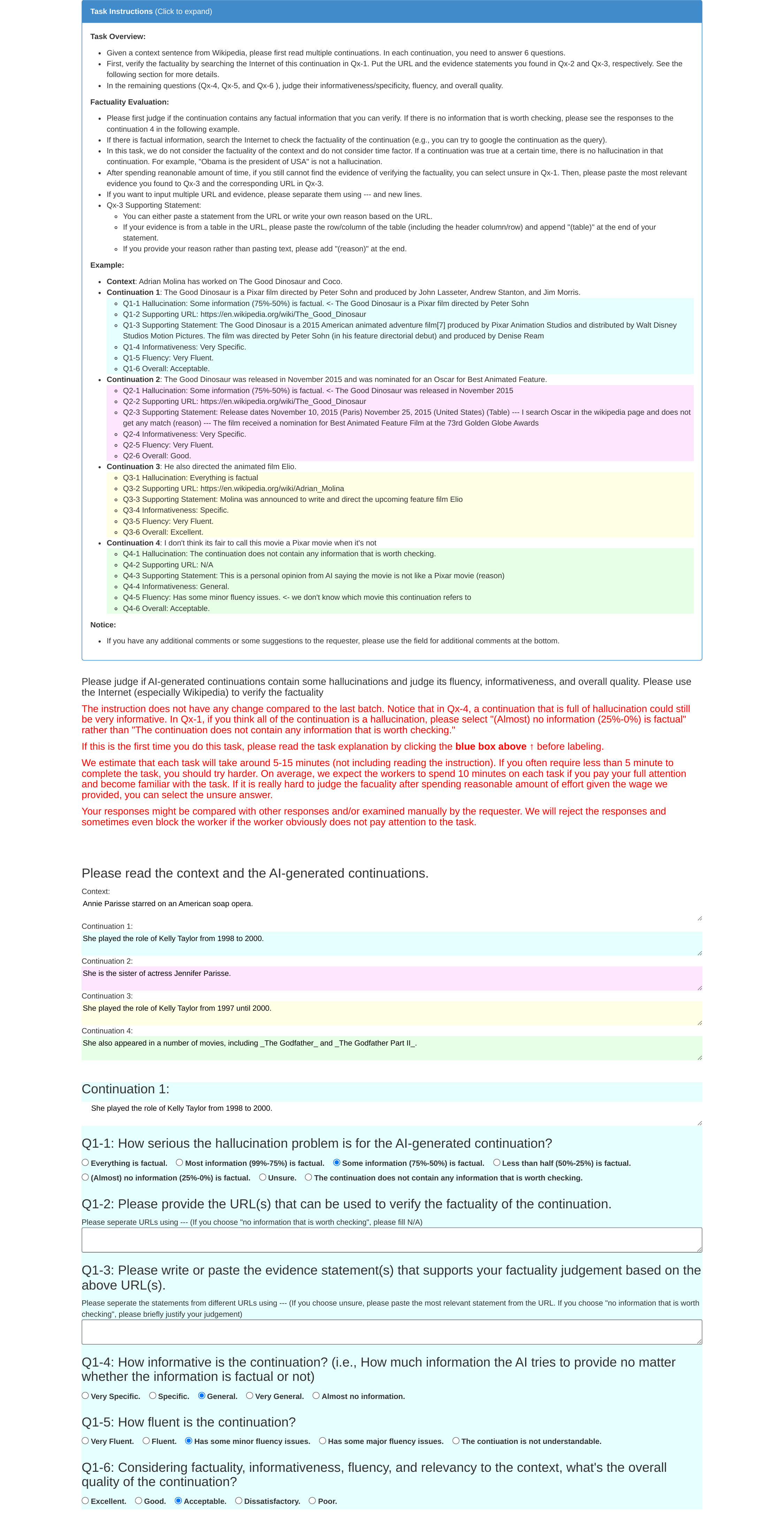}
  %\label{fig:function_abaltion}
\end{subfigure}%
\begin{subfigure}{.5\textwidth}
  \centering
  \includegraphics[width=1\linewidth]{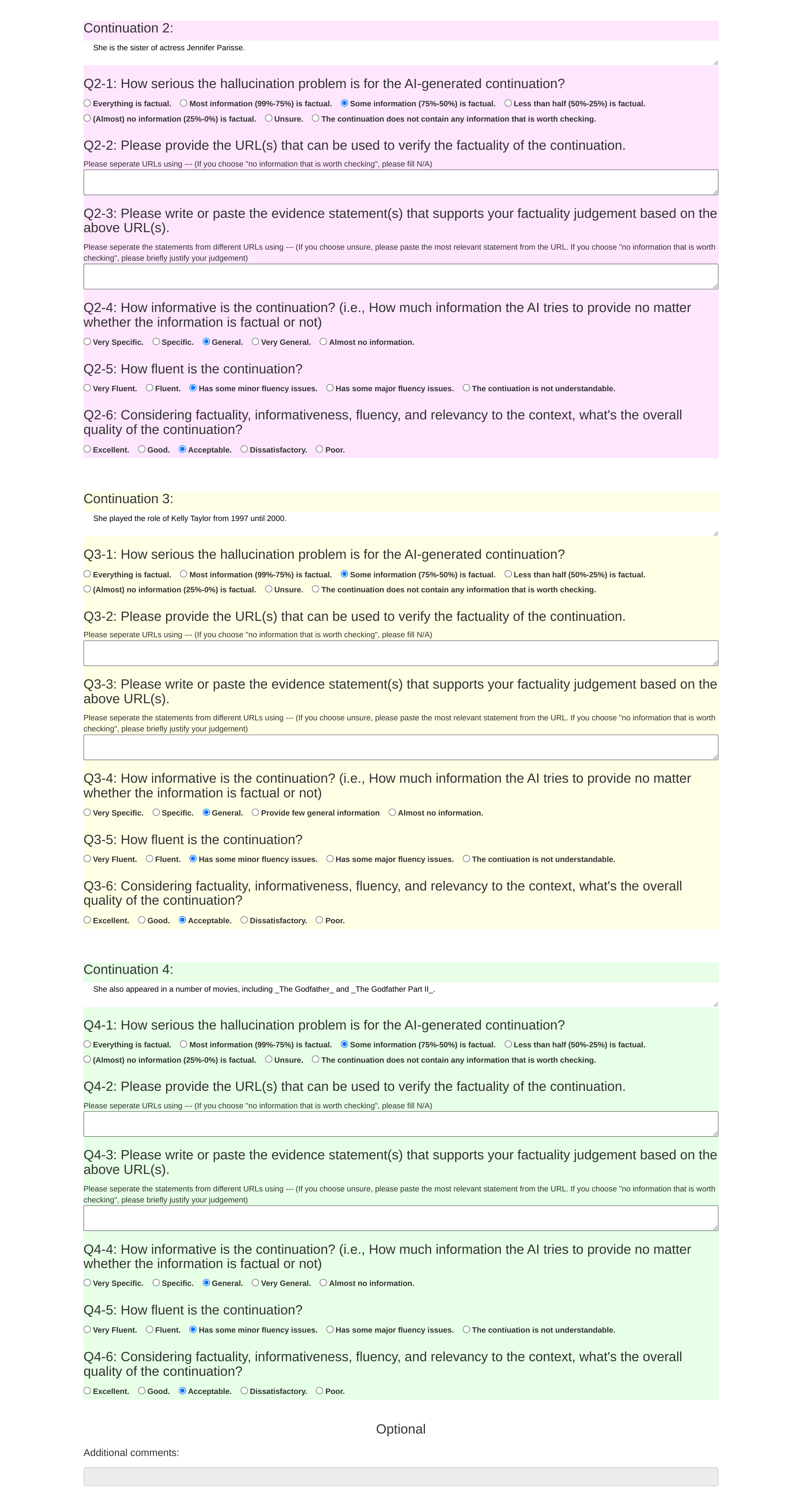}
  %\label{fig:fuction_abaltion_CD}
\end{subfigure}
\caption{The MTurk template for our human experiment.}
\label{fig:factuality_human}
\end{figure*}

In each task, the order of the text generated by all methods is randomized. In \textsc{FactualityPrompts}, the factuality score 5 means no hallucination, and the score 1 means less than 25\% of the continuation is factual. We allow the workers to select the ``unsure'' option if they really cannot find the relevant statement from the Internet and we also allow the workers to select ``no information that is worth checking'' option because the 7B LLM sometimes states their own opinions. We treat both options as score 1 in our evaluation. Please see \Cref{fig:factuality_human} for more details of our MTurk task.

\subsection{Details for Hallucination Detection}

Factor creates nonfactual sentences by revising the factual sentences given a context using ChatGPT, HaDes provides human labels on the phrases infilled by BERT, and TF ext mostly uses templates and tables in different topics to create the factual and nonfactual sentences. The hallucination datasets are created using very different methods and none of the input text is not from Pythia. Our improvements highlight the generalization capability of our model and its potential to improve the applications that leverage the perplexity and entropy measurement (e.g., hallucination detection, AI-generated text detection, and creativity measurement).

Notice that the main purpose of this experiment is to demonstrate the potential of using the THF model to improve unsupervised hallucination detection in general rather than achieving state-of-the-art performances. \citet{muhlgay2023generating} have shown that larger LLMs can better detect the hallucination. In this experiment, all the features are derived from Pythia LLMs with a size smaller than 7B without any additional training or supervision (e.g., supervised training, instruction finetuning, RLHF~\citep{ouyang2022training}, distillation, etc), so it might not work better than directly fine-tuning a model for a particular hallucination detection dataset or a more advanced/larger LLM. 

We use the training and testing split in HaDes~\citep{liu2022token}. For Factor and TF ext, we split each subset into equally large training set and testing set. The maximal depth of the random forest is set as $5$. For Hades, we use only the perplexity and the entropy of the first token in the input phrase as our features, which works better than averaging the perplexities and entropies of all the tokens in the input phrase. In the last two rows of \Cref{tb:hallucination_detection}, we use the code of HaDes~\citep{liu2022token} to perform exhaustive feature selection based on the testing scores, so we can view the results as validation scores. In Hades and TF ext, we choose the best feature set based on AUC and in Factor, we select features using 1-4 ACC.

\subsection{Details for Creative Writing Experiments}

\citet{chiang2023closer} suggest that asking ChatGPT to rate first and give explanation next could increase the quality of the scores. Following the suggestion, we design our prompt and report it in \Cref{prompt:gpt3.5}. To avoid the position bias in the evaluation, we alternatively assign the generation from REAL sampling and from top-$p$ sampling to be story continuation A.

%\hs{Explain the experiment setting of creative writing experiments}
%randomize the order

%percentage 
%We treat None and unsure as hallucination

%excluding the sentences including "External links" because its hard to judge if it is factual or not. 
%exclude the cases where two different methods output the same result
%exclude responses that are less than 10 characters
%top 100 example in the testing set

%\expsec{Metrics:}
%ask workers to provide the url and 

%\todo{showing the crowdsourcing template}

%\expsec{Methods:}
%Top-$p$ ($p=0.6$)
%REAL ($T=2.0$)
%CD ($\alpha=0.3$)
%REAL + CD ($T=1.5$)

%\expsec{Results:}

%Halu Eval
%summarization data truncation to 1024
\newpage
\footnotesize
\begin{myprop}
You are an English writing expert and you can compare and evaluate two continuations on these metrics with the following definitions -

    1. Fluency: Which continuation has better writing and grammar comparitively?
    
    2. Coherence: Which continuation has a better logical flow and the writing fits together with respect to the plot?
    
    3. Likability: Which continuation is more interesting and enjoyable to read?

You will be given two continuations - continuation A and continuation B.

Specify which continuation you prefer for each metric by responding with just the letter “A” or “B” followed by a hyphen and two line justifications for your preference.

Assign an overall winner continuation as the letter “A” or “B” based on the category wins and provide two line justifications.

IMPORTANT - DO NOT GIVE ANY OTHER TEXT APART FROM THE METRICS, PREFERENCE, AND JUSTIFICATIONO.

EXAMPLE OUTPUT 1:

Fluency: B

A: A has some complex sentences that are difficult to follow, with occasional grammatical errors.

B: B is well-written with minor grammatical mistakes and clear sentence structures.

Coherence: B

A: The plot of A is somewhat confusing and disjointed, especially with the sudden introduction of an old sage.

B: B maintains a coherent narrative, with each event logically building on the previous one, enhancing the continuation’s flow.

Likability: B

A: A is heartfelt but its erratic narrative structure detracts from its overall appeal.

B: B is compelling and maintains consistent character development, making it more enjoyable and engaging.

Overall Winner: B

A: A is moderately fluent, coherent, and interesting.

B: B is perfect except for some minor grammar issues.

EXAMPLE OUTPUT 2:

Fluency: A

A: A has a few minor grammatical issues, but overall, it demonstrates strong control of language.

B: B is well-written but has slightly more noticeable issues in grammar and sentence structure.

Coherence: A

A: B has a strong coherence, effectively conveying the progression of events.

B: A maintains a consistent and engaging narrative flow, though some parts are a bit abstract.

Likability: A

A: B’s realistic and emotional narrative is likely to resonate more with a wide range of readers.

B: A is imaginative and intriguing, but its abstract nature might not appeal to all readers.

Overall Winner: A

A: A is very good and it would be better if it can be more interesting.

B: B is too abstract to be interesting.

Context: \textbf{\{Context\}}

Continuation A: \textbf{\{Context\}} \textbf{\{Story Continuation A\}}

Continuation B: \textbf{\{Context\}} \textbf{\{Story Continuation B\}}
\label{prompt:gpt3.5}
\end{myprop}
\normalsize

% \begin{figure}[t!]
%     \centering
%     \includegraphics[width=0.5\linewidth]{figs/factuality_wiki_idx00.pdf}
%     \caption{The MTurk template for our human experiment.}
%     \label{fig:factuality_human}
% \end{figure}

%%%%%%%%%%%%%%%%%%%%%%%%%%%%%%%%%%%%%%%%%%%%%%%%%%%%%%%%%%%%

\newpage
\section*{NeurIPS Paper Checklist}

\begin{enumerate}

\item {\bf Claims}
    \item[] Question: Do the main claims made in the abstract and introduction accurately reflect the paper's contributions and scope?
    \item[] Answer: \answerYes{} % Replace by \answerYes{}, \answerNo{}, or \answerNA{}.
    \item[] Justification: We claim that the proposed REAL sampling improves both factuality and diversity. The claim is supported by various experiments and analyses.

\item {\bf Limitations}
    \item[] Question: Does the paper discuss the limitations of the work performed by the authors?
    \item[] Answer: \answerYes{} % Replace by \answerYes{}, \answerNo{}, or \answerNA{}.
    \item[] Justification: Please see \Cref{sec:future}.

\item {\bf Theory Assumptions and Proofs}
    \item[] Question: For each theoretical result, does the paper provide the full set of assumptions and a complete (and correct) proof?
    \item[] Answer: \answerYes{} % Replace by \answerYes{}, \answerNo{}, or \answerNA{}.
    \item[] Justification: Please see \Cref{sec:proof}

    \item {\bf Experimental Result Reproducibility}
    \item[] Question: Does the paper fully disclose all the information needed to reproduce the main experimental results of the paper to the extent that it affects the main claims and/or conclusions of the paper (regardless of whether the code and data are provided or not)?
    \item[] Answer: \answerYes{} % Replace by \answerYes{}, \answerNo{}, or \answerNA{}.
    \item[] Justification: Please see \Cref{sec:method_details} and \Cref{sec:exp_details}.

\item {\bf Open access to data and code}
    \item[] Question: Does the paper provide open access to the data and code, with sufficient instructions to faithfully reproduce the main experimental results, as described in supplemental material?
    \item[] Answer: \answerNo{} % Replace by \answerYes{}, \answerNo{}, or \answerNA{}.
    \item[] Justification: We plan to publicly release all our code after the paper is accepted.

\item {\bf Experimental Setting/Details}
    \item[] Question: Does the paper specify all the training and test details (e.g., data splits, hyperparameters, how they were chosen, type of optimizer, etc.) necessary to understand the results?
    \item[] Answer: \answerYes{} % Replace by \answerYes{}, \answerNo{}, or \answerNA{}.
    \item[] Justification: Please see \Cref{sec:lee_eval} and \Cref{sec:exp_details}.

\item {\bf Experiment Statistical Significance}
    \item[] Question: Does the paper report error bars suitably and correctly defined or other appropriate information about the statistical significance of the experiments?
    \item[] Answer: \answerYes{} % Replace by \answerYes{}, \answerNo{}, or \answerNA{}.
    \item[] Justification: Please see the captions of \Cref{tb:human_exp}, \Cref{fig:lee_nonfactual_entail}, and \Cref{fig:lee_nonfactual_ne}. We put them into the caption for the clarity of the figures and table. %adding the error bars would make the figures and table too messy.

\item {\bf Experiments Compute Resources}
    \item[] Question: For each experiment, does the paper provide sufficient information on the computer resources (type of compute workers, memory, time of execution) needed to reproduce the experiments?
    \item[] Answer: \answerYes{} % Replace by \answerYes{}, \answerNo{}, or \answerNA{}.
    \item[] Justification: Please see \Cref{sec:exp_details}.
    
\item {\bf Code Of Ethics}
    \item[] Question: Does the research conducted in the paper conform, in every respect, with the NeurIPS Code of Ethics \url{https://neurips.cc/public/EthicsGuidelines}?
    \item[] Answer: \answerYes{} % Replace by \answerYes{}, \answerNo{}, or \answerNA{}.
    \item[] Justification: Please see \Cref{sec:impact}.

\item {\bf Broader Impacts}
    \item[] Question: Does the paper discuss both potential positive societal impacts and negative societal impacts of the work performed?
    \item[] Answer: \answerYes{} % Replace by \answerYes{}, \answerNo{}, or \answerNA{}.
    \item[] Justification: Please see \Cref{sec:impact}.
    
\item {\bf Safeguards}
    \item[] Question: Does the paper describe safeguards that have been put in place for responsible release of data or models that have a high risk for misuse (e.g., pretrained language models, image generators, or scraped datasets)?
    \item[] Answer: \answerNA{} % Replace by \answerYes{}, \answerNo{}, or \answerNA{}.
    \item[] Justification: We do not create a new dataset. Our THF model is an unsupervised model for detecting hallucination, so it should not increase the existing risks of LLM.

\item {\bf Licenses for existing assets}
    \item[] Question: Are the creators or original owners of assets (e.g., code, data, models), used in the paper, properly credited and are the license and terms of use explicitly mentioned and properly respected?
    \item[] Answer: \answerYes{} % Replace by \answerYes{}, \answerNo{}, or \answerNA{}.
    \item[] Justification: We provide the url and the corresponding license for each dataset we download.

\item {\bf New Assets}
    \item[] Question: Are new assets introduced in the paper well documented and is the documentation provided alongside the assets?
    \item[] Answer: \answerNA{} % Replace by \answerYes{}, \answerNo{}, or \answerNA{}.
    \item[] Justification: We do not create a new dataset.

\item {\bf Crowdsourcing and Research with Human Subjects}
    \item[] Question: For crowdsourcing experiments and research with human subjects, does the paper include the full text of instructions given to participants and screenshots, if applicable, as well as details about compensation (if any)? 
    \item[] Answer: \answerYes{} % Replace by \answerYes{}, \answerNo{}, or \answerNA{}.
    \item[] Justification: Please see \Cref{sec:human_details}.

\item {\bf Institutional Review Board (IRB) Approvals or Equivalent for Research with Human Subjects}
    \item[] Question: Does the paper describe potential risks incurred by study participants, whether such risks were disclosed to the subjects, and whether Institutional Review Board (IRB) approvals (or an equivalent approval/review based on the requirements of your country or institution) were obtained?
    \item[] Answer: \answerNo{} % Replace by \answerYes{}, \answerNo{}, or \answerNA{}.
    \item[] Justification: We ask crowd workers to evaluate the texts generated by Pythia, which should not pose risks to the workers.

\end{enumerate}

\end{document}